\documentclass[11pt]{article}
\usepackage{amssymb}
\usepackage[preprint]{acl}

\usepackage{times}
\usepackage{latexsym}
\usepackage{booktabs}
\usepackage{amsmath}

\usepackage[T1]{fontenc}

\usepackage[utf8]{inputenc}

\usepackage{microtype}
\usepackage{longtable}
\usepackage{inconsolata}

\usepackage{graphicx}
\usepackage{pifont}
\usepackage{enumitem}
\DeclareMathOperator*{\argmin}{arg\,min}
\title{Putting HUMANS first: Efficient LAM Evaluation with \\Human Preference Alignment}

\author{
  \textbf{Woody Haosheng Gan\textsuperscript{1*}},
  \textbf{William Held\textsuperscript{2,3}},
  \textbf{Diyi Yang\textsuperscript{2}}
\\
\\
  \textsuperscript{1}University of Southern California,
  \textsuperscript{2}Stanford University,
  \textsuperscript{3}OpenAthena
\\
  \small{\texttt{woodygan@usc.edu, held@stanford.edu, diyiy@stanford.edu}}
}
\begin{document}
\maketitle
\def\thefootnote{*}\footnotetext{Work done while visiting Stanford University.}

\begin{abstract}

The rapid proliferation of large audio models (LAMs) demands efficient approaches for model comparison, yet comprehensive benchmarks are costly. To fill this gap, we investigate whether minimal subsets can reliably evaluate LAMs while reducing costs and data redundancy. Analyzing 10 subset selection methods with 18 audio models across 40 tasks covering major LAM evaluation dimensions, we show that subsets of just 50 examples (0.3\% of data) can achieve over 0.93 Pearson correlation with full benchmark scores. To understand how well these scores align with what practitioners ultimately care about--user satisfaction--we collect 776 human preference ratings from realistic voice assistant conversations, finding that both subsets and full benchmark achieve only 0.85 correlation with human. To better predict preferences, we trained regression models on these selected subsets, achieving 0.98 correlation—outperforming regression models trained on both random subsets and the full benchmark. This demonstrates that in regression modeling, well-curated subsets outpredict the full benchmark, showing quality over quantity. We open-source these regression-weighted subsets as the HUMANS benchmark, an efficient proxy for LAM evaluation that captures both benchmark performance and user preferences.
\end{abstract}

\noindent\textbf{HUMANS Benchmark:} \url{https://huggingface.co/datasets/woodygan/humans-benchmark} 

\section{Introduction}
\begin{figure}[t]
\centering
\includegraphics[width=\columnwidth]{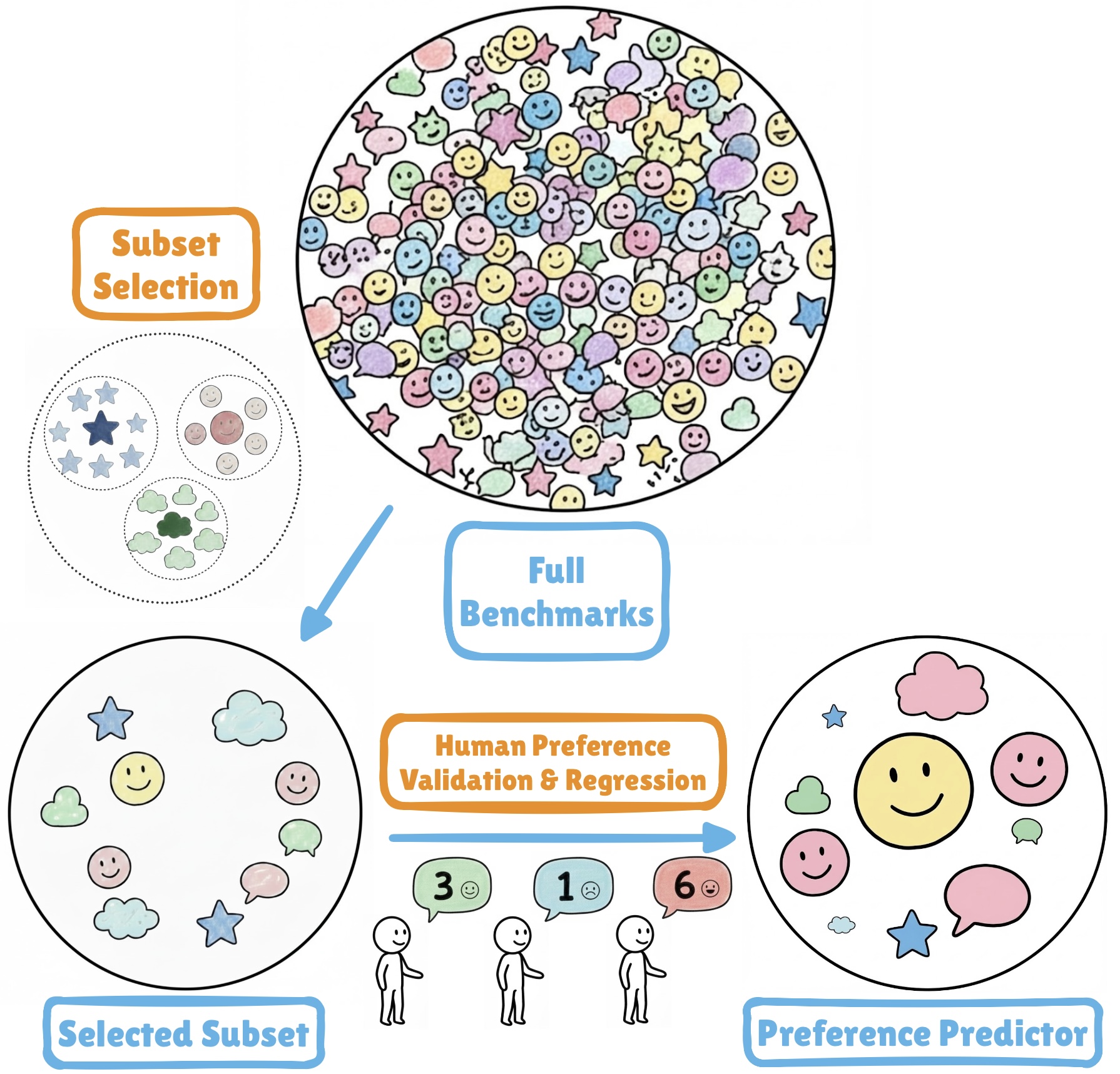}
\caption{\textbf{Overview.} We select minimal subsets from full benchmark pools, validate alignment with human preferences through interactive evaluations, and train regression models to efficiently predict user satisfaction.}
\label{fig:overview}
\end{figure}
The landscape of large audio models (LAMs) has expanded rapidly, with families like Gemini~\citep{team2023gemini}, GPT-audio~\citep{openai2024gpt4o-audio}, Qwen-Omni~\citep{xu2025qwen2}, and Ultravox~\citep{fixie2024ultravox} demonstrating diverse capabilities. This proliferation creates a practical challenge: how to quickly compare and select models without exhaustive evaluation. Existing LAM benchmarks containing thousands of examples create substantial computational barriers—audio evaluation requires 10--100$\times$ more tokens than text, making single-model evaluation cost hundreds of GPU-hours and dollars. This makes it impractical to quickly compare candidate models, evaluate checkpoints, or A/B test configurations. More critically, static benchmarks may poorly align with human preferences~\citep{li2025mind, schaeffer2025correlating}, failing to reflect what practitioners care about when conversational quality and user experience are paramount. This raises two critical questions: \textit{Can we reliably rank LAMs using small benchmark subsets? How to capture what users actually care about?}

To answer these questions, we first conduct a comprehensive analysis of benchmark subset selection for LAMs. Evaluating 18 audio models across 40 tasks covering major LAM conversation scenarios (${\sim}16{,}000$ datapoints from 5 benchmarks), we systematically compare 10 subset selection methods, showing that carefully selected subsets of just 50 examples (0.3\% of data) achieve 0.934 Pearson correlation with full benchmark scores.

To understand whether benchmark scores reflect user satisfaction in real-world deployment, we collect 776 human preference ratings from 10-minute interactive conversations with 7 representative models across realistic scenarios spanning tool calling, task-oriented dialogue, and open chat. Our analysis reveals that both selected subsets and the full benchmark plateau at 0.85 correlation, indicating a substantial gap between static evaluation metrics and real-world user experience. Beyond quantitative ratings, we analyze qualitative feedback, revealing failure modes such as excessive verbosity and robotic speech not emphasized by benchmarks.

To further improve human preference prediction, we train regression models on selected subsets, creating HUMANS (HUman-aligned Minimal Audio evaluatioN Subsets) benchmark, which achieves 0.978 correlation with user satisfaction, outperforming regression models on random sampling and full benchmark. Our contributions are:

\begin{enumerate}[leftmargin=*,noitemsep]
    \item Systematic evaluation of 10 subset selection methods for audio benchmarks, demonstrating that small subsets enable reliable model ranking while dramatically reducing costs
    \item A human preference dataset of 776 ratings from realistic voice assistant interactions, understanding benchmark-human alignment, providing qualitative analysis on user feedback, and enabling meta-evaluation of future benchmarks
    \item Demonstration that regression models trained on benchmark subsets predict human preferences well, providing an efficient proxy that captures benchmark performance and user preference

\end{enumerate}

\section{Related Work}

\subsection{Large Audio Models}

The landscape of large audio models (LAMs) has rapidly evolved from specialized architectures like Whisper~\citep{radford2023robust} for speech recognition and VALL-E~\citep{wang2023neural} for synthesis into versatile, general-purpose models. Contemporary LAMs include audio-in text-out models that process speech for text generation (e.g., Gemini~\citep{team2023gemini}, Ultravox~\citep{fixie2024ultravox}, Voxtral~\citep{liu2025voxtral}, Gemma 3n~\citep{gemma_3n_2025}, Phi-4-multimodal~\citep{abouelenin2025phi}) and end-to-end omni-modal systems that natively handle audio input and output (e.g., GPT-realtime~\citep{openai2025gpt-realtime}, Qwen-Omni~\citep{xu2025qwen2}, GLM-4-Voice~\citep{zeng2024glm}, and MiniCPM-o~\citep{yao2024minicpm}). This diversity makes systematically comparing models complex.
\subsection{LAM Evaluation Benchmarks}
The LAM evaluation landscape includes specialized benchmarks targeting specific aspects (SpeakBench~\citep{manakul2025audiojudge} for paralinguistics, MMAU~\citep{sakshi2024mmau} for reasoning, ADU-Bench~\citep{gao2024benchmarking} for dialogue), application-oriented benchmarks focusing on voice assistant scenarios (WildSpeech-Bench~\citep{zhang2025wildspeech}, VoiceBench~\citep{chen2024voicebench}), general audio understanding benchmarks (AudioBench~\citep{wang2024audiobench}, AIR-Bench~\citep{yang2024air}), and comprehensive frameworks (Dynamic-SUPERB~\citep{huang2024dynamic}, UltraEval-Audio~\citep{he2024ultraeval}, CAVA~\citep{cava2025}). This fragmentation and extensive scope creates substantial computational burden, motivating efficient subset selection.

\subsection{Benchmark Subset Selection Methods}
Sample-efficient benchmarking has roots in psychometrics, particularly Item Response Theory (IRT) for selecting discriminative items~\citep{lalor2016building, martinez2019item}, with foundational extensions to diversity-based clustering~\citep{misir2021benchmark}, training dynamics~\citep{swayamdipta2020dataset}, and gradient-based active learning~\citep{coleman2020selection}. Modern adaptations on LLMs include Anchor Points~\citep{vivek2023anchor}, Efficient Benchmarking~\citep{perlitz2024efficient}, TinyBenchmarks~\citep{polo2024tinybenchmarks}, and SUBLIME~\citep{saranathan2025sublime} that achieve high correlation with full rankings using minimal subsets. However, these techniques remain largely unexplored for LAM evaluation.

\subsection{Human Preference and Meta-Analysis}
Traditional human evaluation assessed perceptual quality using metrics like MOS~\citep{ITU-T-P800-1996}. In the LLM era, Chatbot Arena~\citep{chiang2024chatbot} introduced large-scale pairwise preference collection, with the LMSYS dataset~\citep{zheng2023lmsys} becoming a gold standard for meta-evaluating benchmarks. Human preferences are leveraged to predict satisfaction on unseen models~\citep{schaeffer2025correlating, ryan2025autometrics}. Recent work extended this to audio models: TalkArena~\citep{li2025mind} collected preferences on audio-in text-out systems in single-turn interactions, revealing significant misalignment between benchmark scores and human preferences. In our human evaluation, we further captures more realistic deployment scenarios of LAMs: real-time voice assistants handling multi-turn conversations and tool interactions.

\section{Subset Selection}
\label{sec:subset_selection}

In this section, we systematically evaluate methods for selecting minimal yet informative benchmark subsets that preserve model rankings, across 40 tasks covering major LAM conversation evaluation dimensions (from 5 benchmarks, ${\sim}16{,}000$ datapoints). Through cross-validation on 18 diverse audio models, we identify the most effective subset selection methods and construct final minimal benchmark subsets for practical use.
\subsection{Experimental Setup}

\subsubsection{Audio Models}
\label{sec:audio_models}
We evaluate 18 publicly available audio models with diverse characteristics to ensure our findings generalize across the LAM landscape. Our selection spans multiple architectural paradigms: end-to-end omni-modal systems that natively process and generate speech (e.g., GPT-4o-audio, Qwen2.5-Omni), speech-to-text models that encode audio for text-based reasoning (e.g., Gemini 2.5, Ultravox, Voxtral), and pipeline systems combining separate components (e.g., Llama-3.2 with external STT/TTS). Models range from 1B parameters to large proprietary systems. For consistency in evaluation, all models without native audio output use GPT-4o-mini-tts~\citep{openai2025gpt4o-mini-tts} for speech synthesis, while text-only models in pipeline 
configurations use GPT-4o-transcribe~\citep{openai2025gpt4o-transcribe} for audio input transcription. Complete model specifications, architectural details, and processing configurations 
are provided in Appendix~\ref{appendix:models}.

\subsubsection{Benchmarks}

We construct our evaluation suite from 5 established audio benchmarks: Dynamic-SUPERB Phase 2~\citep{huang2024dynamic2}, CAVA~\citep{cava2025}, UltraEval-Audio~\citep{he2024ultraeval}, SpeakBench~\citep{manakul2025audiojudge}, and WildSpeech-Bench~\citep{zhang2025wildspeech}, selecting tasks focused on LAM capabilities on human conversation and speech. These benchmarks are selected to be complementary, collectively providing tasks that represent the majority of evaluation scenarios in recent LAM literature. This yields 40 distinct evaluation tasks evaluating different dimensions of audio model capabilities, including speech recognition, dialogue understanding, instruction following, multiturn function calling, and more. To ensure comparability across diverse metrics, we unify scales of all metrics to [0,1] where 1 represents best performance. Complete task descriptions and metric unification procedures are detailed in Appendix~\ref{appendix:benchmarks}.

Our full evaluation across 18 models and ${\sim}16{,}000$ examples required approximately 1520 GPU-hours on NVIDIA A6000 GPUs for open-source models and \$2{,}400 in API costs for proprietary models, motivating our investigation of selecting minimal and informative subsets.

\subsubsection{Full Benchmark Reference Scores}

To establish reference scores, we compute task-averaged scores where each task contributes equally regardless of item count:

\begin{equation}
\text{Score}(m) = \frac{1}{T} \sum_{t=1}^{T} \bar{s}_{m,t}
\end{equation}

where $T$ is the number of tasks and $\bar{s}_{m,t}$ is model $m$'s average score across all items in task $t$. Since our 40 tasks are selected to cover major LAM conversation evaluation dimensions, equal task weighting ensures each dimension contributes proportionally, preventing tasks with more examples from dominating rankings. These reference scores serve as our gold standard for evaluating whether selected subsets preserve model rankings.

\subsection{Subset Selection Methods}

\subsubsection{Random Sampling Methods}

\paragraph{}\textbf{Task-Balanced Random Sampling.} As our baseline, we employ task-balanced random sampling where each datapoint in task $t$ (containing $x_t$ items) has sampling probability $p_i = 1/(T \cdot x_t)$, where $T$ is the number of tasks, ensuring each task contributes equally in expectation: $\mathbb{E}[\text{samples from task } t] = n/T$.

\textbf{Random-Sampling-Learn.} Building on the baseline subset, it uses Ridge regression to predict full benchmark scores from subset scores~\citep{zhang2025benchmark}. we learn $g$ by minimizing regularized loss over source models $\mathcal{M}$, then predict target scores as $h(f) = g[s(f, C)]$ (see Appendix~\ref{appendix:random_sampling_learn}).

\textbf{Random-Search-Learn.} This extends Random-Sampling-Learn by performing $N=1000$ random sampling iterations, training Ridge regression on each using 75\% of source models for training and 25\% for validation, selecting the subset with lowest validation error, then retraining on all models.

\subsubsection{Intrinsic Item Property Methods}

\paragraph{}\textbf{Variance-Based Selection.} We select items with highest discriminative power. For each item $i$, we compute variance $\sigma_i^2 = \frac{1}{K-1} \sum_{k=1}^{K} (s_{i,k} - \bar{s}_i)^2$ across model performances where $s_{i,k}$ is model $k$'s score on item $i$, and select the top $n$ highest-variance items globally.

\textbf{Difficulty-Based Selection.} We employ stratified sampling to span the full difficulty spectrum~\citep{saranathan2025sublime}. We define difficulty as $D_i = 1 - \frac{1}{K}\sum_{k=1}^{K} s_{i,k}$, partition items into $B=10$ bins, and allocate equal samples per bin using task-balanced probabilities.

\subsubsection{Embedding-Based Clustering Methods}

\paragraph{}\textbf{IRT-Based Performance Prediction.} Inspired by tinyBenchmarks~\citep{polo2024tinybenchmarks}, we train a 5-dimensional two-parameter IRT model on source model responses to estimate latent item parameters (discrimination $\alpha_i \in \mathbb{R}^5$ and difficulty $\beta_i \in \mathbb{R}$). We construct 6-dimensional item embeddings $E_i = [\alpha_i; \beta_i]$ that encode each item's latent characteristics, which are more robust to distribution shift than raw correctness patterns. We use these embeddings for task-aware weighted K-Means clustering to select $n$ anchor points. For target model prediction, we estimate ability parameters $\hat{\theta}_m$ from its anchor responses, then compute task-averaged scores using actual responses for observed items and IRT-predicted probabilities $\hat{p}_{im} = \sigma(\hat{\alpha}_i^\top \hat{\theta}_m - \hat{\beta}_i)$ for unseen items. Details are in Appendix~\ref{appendix:irt_details}.

\textbf{Anchor-Based Selection.} We adapt the anchor points framework~\citep{vivek2023anchor} with task-aware weighting. We apply weighted K-Means clustering on item embeddings using Euclidean distance, with each item weighted by $1/(T \cdot |T_t|)$ where $T$ is the number of tasks and $|T_t|$ is the task size, resulting in $n$ clusters. We select the datapoint nearest to each centroid as an anchor point.

For target model $m$, the Anchor Point Weighted (APW) score is:
\begin{equation}
\text{APW}(m) = \sum_{i=1}^{n} w_i \cdot s_{m,a_i}
\end{equation}
where $a_i$ is anchor point $i$, $s_{m,a_i}$ is model $m$'s normalized score on $a_i$, and $w_i = \sum_{j \in C_i} b_j$ is the cluster weight (sum of task-normalized weights $b_j$ for items in cluster $C_i$), ensuring equal task contribution. Implementation details are in Appendix~\ref{appendix:anchor_details}.

\paragraph{Embedding Choices for Clustering.} We explore four embedding spaces for the anchor-based clustering step, each producing a method variant:

\textbf{Anchor Points (APW):} Clusters directly on source model score vectors (the original method).

\textbf{Semantic Embedding:} Prompts encoded using OpenAI's text-embedding-3-large~\citep{openai2024embeddings}, PCA-reduced from 3072 to 50 dimensions.

\textbf{Acoustic Embedding:} 1024-dim acoustic embeddings extracted using WavLM-Large~\citep{chen2022wavlm}, reduced to 50 dimensions via PCA.

\textbf{Combined Embedding:} Combined representations concatenating: (1) acoustic embeddings using WavLM-Large~\citep{chen2022wavlm}, (2) semantic embeddings using OpenAI's text-embedding-3-large~\citep{openai2024embeddings}, (3) source model performance scores, and (4) binary metadata indicating whether audio input/output is required. Acoustic and semantic embeddings are PCA-reduced to match source model count (e.g., 18 dims if 18 models in (3)) and MinMax-scaled to [0,1] to match the range of performance scores and metadata. Equal dimensionality for the first three components ensures balanced contribution to clustering.

\subsection{Subset Selection Performance}
\label{ssec:performance}
\begin{table*}[t]
\centering
\small
\resizebox{\textwidth}{!}{%
\begin{tabular}{lcccccccc}
\toprule
\textbf{Method} & \multicolumn{6}{c}{\textbf{Pearson Correlation by Subset Size}} & \textbf{AUCC} & $N_{90}$ / $N_{95}$ \\
\cmidrule(lr){2-7} \cmidrule(lr){8-8} \cmidrule(lr){9-9}
& $n=10$ & $n=20$ & $n=30$ & $n=50$ & $n=100$ & $n=200$ & $[10,200]$ & \\
\midrule
Random Sampling & 0.559$_{\pm\text{0.021}}$ & 0.718$_{\pm\text{0.014}}$ & 0.784$_{\pm\text{0.011}}$ & 0.856$_{\pm\text{0.008}}$ & 0.916$_{\pm\text{0.005}}$ & 0.959$_{\pm\text{0.003}}$ & 0.891 & 83 / 164 \\
Random-Sampling-Learn & 0.544$_{\pm\text{0.022}}$ & 0.656$_{\pm\text{0.017}}$ & 0.719$_{\pm\text{0.014}}$ & 0.791$_{\pm\text{0.012}}$ & 0.887$_{\pm\text{0.006}}$ & 0.940$_{\pm\text{0.003}}$ & 0.854 & 119 / 300 \\
Random-Search-Learn & 0.619$_{\pm\text{0.018}}$ & 0.676$_{\pm\text{0.017}}$ & 0.734$_{\pm\text{0.015}}$ & 0.803$_{\pm\text{0.010}}$ & 0.894$_{\pm\text{0.005}}$ & 0.937$_{\pm\text{0.004}}$ & 0.866 & 99 / 300 \\
Variance-based & 0.525$_{\pm\text{0.020}}$ & 0.628$_{\pm\text{0.016}}$ & 0.676$_{\pm\text{0.014}}$ & 0.716$_{\pm\text{0.013}}$ & 0.756$_{\pm\text{0.012}}$ & 0.804$_{\pm\text{0.011}}$ & 0.742 & -- / -- \\
Difficulty-based & 0.608$_{\pm\text{0.020}}$ & 0.761$_{\pm\text{0.012}}$ & 0.811$_{\pm\text{0.009}}$ & 0.863$_{\pm\text{0.007}}$ & 0.924$_{\pm\text{0.004}}$ & 0.964$_{\pm\text{0.002}}$ & 0.902 & 71 / 157 \\
IRT-based & 0.486$_{\pm\text{0.020}}$ & 0.698$_{\pm\text{0.014}}$ & 0.778$_{\pm\text{0.013}}$ & 0.864$_{\pm\text{0.007}}$ & 0.919$_{\pm\text{0.004}}$ & \underline{0.960$_{\pm\text{0.002}}$} & 0.892 & 81 / 156 \\
Anchor Points & \textbf{0.797$_{\pm\text{0.011}}$} & \textbf{0.856$_{\pm\text{0.007}}$} & \textbf{0.884$_{\pm\text{0.006}}$} & \underline{0.907$_{\pm\text{0.005}}$} & \underline{0.940$_{\pm\text{0.004}}$} & 0.952$_{\pm\text{0.003}}$ & \underline{0.927} & \underline{40} / 155 \\
Semantic Embedding & 0.466$_{\pm\text{0.023}}$ & 0.627$_{\pm\text{0.016}}$ & 0.781$_{\pm\text{0.012}}$ & 0.877$_{\pm\text{0.007}}$ & 0.921$_{\pm\text{0.005}}$ & 0.936$_{\pm\text{0.003}}$ & 0.856 & 60 / 350 \\
Acoustic Embedding$^\dagger$ & \underline{0.736$_{\pm\text{0.013}}$} & 0.445$_{\pm\text{0.019}}$ & 0.672$_{\pm\text{0.016}}$ & 0.870$_{\pm\text{0.008}}$ & 0.904$_{\pm\text{0.006}}$ & 0.943$_{\pm\text{0.003}}$ & 0.850 & 92 / 250 \\
Combined Embedding$^\dagger$ & 0.651$_{\pm\text{0.019}}$ & \underline{0.831$_{\pm\text{0.010}}$} & \underline{0.878$_{\pm\text{0.007}}$} & \textbf{0.934$_{\pm\text{0.004}}$} & \textbf{0.963$_{\pm\text{0.002}}$} & \textbf{0.977$_{\pm\text{0.001}}$} & \textbf{0.943} & \textbf{32} / \textbf{67} \\
\bottomrule
\end{tabular}
}
\caption{\textbf{Subset selection performance across methods and sizes.} Pearson correlation between subset and full benchmark scores (mean $\pm$ SEM over 300 evaluations). AUCC computed over $n \in [10,200]$. $N_{90}$/$N_{95}$ show minimum sizes achieving $r \geq 0.90/0.95$. ``--'' indicates threshold not achieved within $n=1000$. \textbf{Bold} indicates best, \underline{underline} second-best. $^\dagger$Audio-specific methods unique to this work (leverage acoustic features).\vspace{-1em}}
\label{tab:subset_performance}
\end{table*}

\begin{figure}[t]
\centering
\includegraphics[width=\columnwidth]{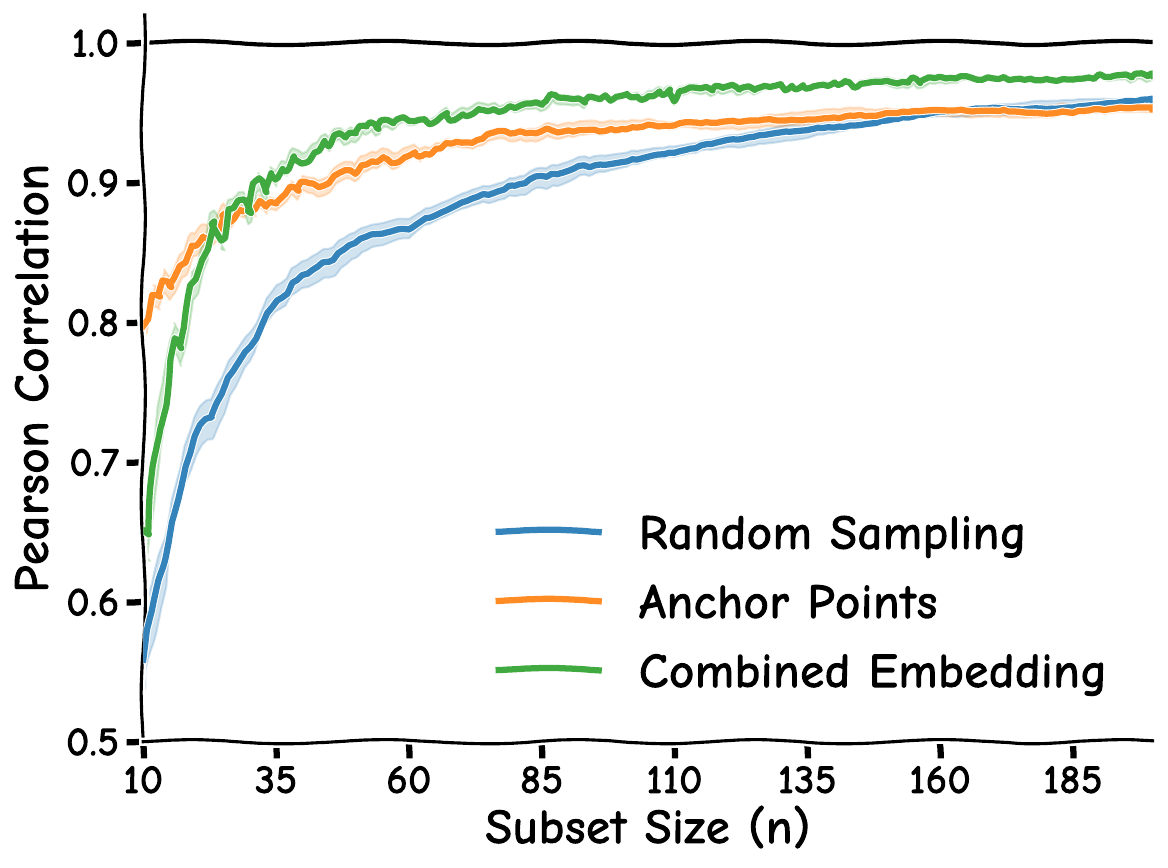}
\caption{\textbf{Subset selection performance.} Pearson correlation with full benchmark scores. Combined Embedding achieves best overall performance (AUCC=0.943), while Anchor Points excel at small sizes ($n \leq 30$). For clarity, only the top-performing and baseline methods are shown here; complete correlation curves for all 10 methods are provided in Appendix~\ref{appendix:complete_results}.}
\label{fig:correlation_curves}
\end{figure}

We evaluate each subset selection method across varying subset sizes from $n=10$ to $n=1000$ using 3-fold cross-validation with 100 random repeats (300 total evaluations per size). In each fold, we use 12 models for subset selection and evaluate alignment with full benchmark scores on the remaining 6 held-out models via Pearson correlation. Table~\ref{tab:subset_performance} reports correlation at key subset sizes ($n \in \{10, 20, 30, 50, 100, 200\}$) and Area Under the Correlation Curve (AUCC) over $n \in [10, 200]$, representing the average correlation achieved across this range. We also report $N_{90}$ and $N_{95}$—the minimum subset sizes achieving Pearson correlation $r \geq 0.90$ and $r \geq 0.95$ with full benchmark scores, respectively. Figure~\ref{fig:correlation_curves} shows the correlation curves for the two top-performing methods and random sampling baseline across subset sizes. Complete correlation curves for all methods and results for alternative correlation metrics are provided in Appendix~\ref{appendix:complete_results}.

Key findings from our evaluation:

\begin{itemize}[noitemsep]

    \item \textbf{Combined Embedding achieves best overall performance}: Highest AUCC (0.943) and superior correlation for $n \geq 50$, reaching $r=0.977$ at $n=200$.

    \item \textbf{Anchor Points excel at small subset sizes}: Best performance for $n \leq 30$ (e.g., $r=0.797$ at $n=10$), demonstrating superior sample efficiency in minimal-evaluation scenarios.

    \item \textbf{Random Sampling provides surprisingly strong baseline}: With 0.891 AUCC and $r=0.959$ at $n=200$, random sampling outperforms variance-based, learning-based, and single-modality embedding approaches.

    \item \textbf{Learning-based methods substantially underperform}: Random-Sampling-Learn and IRT-based approaches achieve relatively lower AUCC (0.854, 0.892), potentially due to overfitting when generalizing learned patterns from limited source models to unseen models even with regularization.

    \item \textbf{Small subsets strongly correlate with full benchmarks}: Combined Embedding reaches $r=0.934$ with only 50 samples ($\sim$0.3\% of full benchmark), enabling reliable model ranking with minimal evaluation.

\end{itemize}
Based on these results, we select Anchor Points for $n \leq 30$ and Combined Embedding for $n \geq 50$ as our best-performing methods. For each subset size, we apply the corresponding method using all 18 models as source models to construct the final benchmark subsets. These selected subsets provide practitioners with reliable, minimal evaluation sets that align with full benchmark scores while dramatically reducing costs. We use them as our "best" subsets for the analysis in Sections~\ref{sec:subset_validation} and~\ref{sec:preference_prediction}. Further analysis of task composition in these subsets reveals that these clustering-based methods naturally prioritize foundational capabilities like ASR and speaker diarization in smaller subsets, progressively incorporating more refined tasks as subset size increases (detailed in Appendix~\ref{appendix:task_distribution}).


\section{Human Evaluation}

Beyond benchmark performance, practitioners ultimately care about real-world user experience. To obtain this gold standard for model performance, we collect human preference ratings in realistic voice assistant scenarios. This enables us to examine whether selected subsets and full benchmark scores align with user satisfaction. 

\subsection{Experimental Design}

\subsubsection{Conversational Agent Framework}
Recognizing that large audio models are predominantly deployed as real-time conversational agents, we develop a model-agnostic voice agent framework adapted from LiveKit~\citep{livekit2024} to enable real-time conversational evaluation. The framework supports end-to-end audio models, audio-in text-out models, and text-only models. For models without native audio input or output capabilities, we use the same STT and TTS components as described in Section~\ref{sec:audio_models} to ensure consistent audio processing across all evaluations. To further ensure fair comparison, we implement consistent system prompts, conversation management, and interaction protocols across all models. Technical details are provided in Appendix~\ref{appendix:agent_framework}.

\subsubsection{Model Selection}
We evaluate 7 representative models from our pool of 18, spanning diverse architectures and sizes: GPT-4o-audio-preview, GPT-4o-mini-audio-preview, Gemini-2.5-Flash, Qwen3-Omni-30B-A3B-Instruct-thinker, Ultravox-v0.4-ToolACE-8B, Voxtral-Small-24B-2507, and GPT-4o-mini.

\subsubsection{Participant Recruitment}
We recruited native English speakers from the United States via Prolific. This research was approved by the Institutional Review Board (IRB) at the authors' institution. Full human evaluation details are in Appendix~\ref{appendix:human_subjects}.

\subsubsection{Conversation Protocol}
Each participant engaged in a single 10-minute conversation\footnote{We chose 10 minutes based on pilot testing - it's long enough to capture multi-turn dialogue patterns and task completion (15-25 turns typical) while avoiding participant fatigue.} with a randomly assigned model and scenario. To capture realistic deployment conditions and ensure human evaluations broadly cover audio models' capabilities, we designed three scenario categories based on common voice assistant use cases~\cite{bentley2018understanding}, emphasizing structured, evaluable tasks while maintaining representation of free-form conversation. Scenario generation details in Appendix~\ref{appendix:scenario_design}:

\begin{itemize}[noitemsep]
    \item \textbf{Open Chat (20\%)}: Free-form conversations without specific goals or instructions, allowing natural interaction patterns to emerge.
    \item \textbf{Goal-Oriented Dialogue (40\%)}: Structured conversations with defined objectives, using real interaction patterns sampled from LMSYS~\citep{zheng2023lmsys} and WildChat~\citep{zhao2024wildchat} datasets.
    \item \textbf{Tool Calling Tasks (40\%)}: Objective-driven interactions requiring specific actions (shopping, messaging, calendar management, flight booking) where task completion can be measured and displayed to the participant.
\end{itemize}

\subsubsection{Rating Collection}
Following each conversation, participants provided ratings on a 6-point Likert scale across five dimensions as well as open-ended feedback justifying their ratings:

\begin{itemize}[noitemsep]
    \item \textbf{Overall Satisfaction}: Holistic assessment of the interaction experience
    \item \textbf{Speech Understanding}: How well the assistant understood speech, intent, and paralinguistic cues
    \item \textbf{Naturalness}: How natural, conversational, and appropriately concise the interaction felt
    \item \textbf{Response Quality}: Accuracy, safety, relevance, and helpfulness of responses
    \item \textbf{Task Effectiveness}: Success and efficiency in helping achieve goals
\end{itemize}

\subsection{Human Evaluation Results}
\label{sec:human_results}
\begin{table*}[t]
\centering
\resizebox{\textwidth}{!}{
\begin{tabular}{lccccccc}
\toprule
& \multicolumn{5}{c}{\textbf{Human Evaluation}} & & \textbf{Benchmark} \\
\cmidrule(lr){2-6}
\textbf{Model} & \textbf{Overall} & \textbf{Understanding} & \textbf{Naturalness} & \textbf{Quality} & \textbf{Effectiveness} & \textbf{N} & \textbf{Score} \\
\midrule
GPT-4o-audio-preview & $\textbf{4.982} \pm {\scriptstyle 0.091}$ & $\textbf{5.368} \pm {\scriptstyle 0.080}$ & $\textbf{4.368} \pm {\scriptstyle 0.117}$ & $\textbf{5.123} \pm {\scriptstyle 0.086}$ & $\textbf{4.947} \pm {\scriptstyle 0.105}$ & 114 & $\underline{0.575}$ \\
Gemini-2.5-Flash+TTS & $\underline{4.664} \pm {\scriptstyle 0.111}$ & $\underline{5.191} \pm {\scriptstyle 0.100}$ & $\underline{4.218} \pm {\scriptstyle 0.123}$ & $\underline{4.936} \pm {\scriptstyle 0.090}$ & $4.673 \pm {\scriptstyle 0.113}$ & 110 & $\textbf{0.589}$ \\
GPT-4o-mini+STT+TTS & $4.509 \pm {\scriptstyle 0.122}$ & $5.158 \pm {\scriptstyle 0.093}$ & $4.132 \pm {\scriptstyle 0.109}$ & $4.772 \pm {\scriptstyle 0.100}$ & $\underline{4.754} \pm {\scriptstyle 0.110}$ & 114 & $0.498$ \\
Qwen3-Omni-30B+TTS & $4.211 \pm {\scriptstyle 0.140}$ & $4.872 \pm {\scriptstyle 0.122}$ & $4.000 \pm {\scriptstyle 0.129}$ & $4.578 \pm {\scriptstyle 0.122}$ & $4.385 \pm {\scriptstyle 0.143}$ & 109 & $\underline{0.575}$ \\
Voxtral-Small-24B+TTS & $3.982 \pm {\scriptstyle 0.135}$ & $4.618 \pm {\scriptstyle 0.128}$ & $3.845 \pm {\scriptstyle 0.135}$ & $4.264 \pm {\scriptstyle 0.128}$ & $4.036 \pm {\scriptstyle 0.140}$ & 110 & $0.507$ \\
GPT-4o-mini-audio-preview & $3.685 \pm {\scriptstyle 0.147}$ & $4.741 \pm {\scriptstyle 0.116}$ & $3.546 \pm {\scriptstyle 0.129}$ & $4.296 \pm {\scriptstyle 0.134}$ & $4.176 \pm {\scriptstyle 0.143}$ & 108 & $0.466$ \\
Ultravox-v0.4-ToolACE-8B+TTS & $3.342 \pm {\scriptstyle 0.156}$ & $4.036 \pm {\scriptstyle 0.159}$ & $3.063 \pm {\scriptstyle 0.139}$ & $3.721 \pm {\scriptstyle 0.155}$ & $3.550 \pm {\scriptstyle 0.161}$ & 111 & $0.384$ \\
\bottomrule
\end{tabular}
}
\caption{\textbf{Human preference evaluation results.} Mean ratings $\pm$ standard error of the mean across five dimensions on a 6-point Likert scale (higher is better). Benchmark shows task-averaged scores from the full benchmark suite. N indicates the number of 10-minute conversations collected per model. Models with +TTS use GPT-4o-mini-tts for speech synthesis, while +STT+TTS indicates a full pipeline system. Models are ordered by Overall Satisfaction.}
\label{tab:human_ratings}
\end{table*}
We collected 776 total evaluations across the 7 models (approximately 111 conversations per model on average). Table~\ref{tab:human_ratings} presents the average human ratings on each dimension.
\begin{itemize}[leftmargin=*]

    \item \textbf{Dimension-Specific Insights}: Understanding consistently exceeds overall satisfaction across models, indicating speech comprehension is not a limiting factor. In contrast, Naturalness scores fall below overall satisfaction, revealing conversational flow as the primary bottleneck. Dimension correlation analysis in Appendix~\ref{appendix:dimension_correlations} reveals Response Quality (r=0.773) and Task Effectiveness (r=0.781) drive satisfaction most strongly, while Naturalness shows the weakest correlation (r=0.626)—suggesting users prioritize functional capabilities in their evaluation.
    
    \item \textbf{Qualitative Failure Mode Analysis}: Open-ended feedback reveals conversational quality issues dominate complaints: robotic speech style (42.8\%), stilted flow (18.8\%), and excessive verbosity (17.2\%) appear in 56.7\% of dissatisfied cases. Poor speech recognition accounts for only 8.7\%, confirming ASR is largely solved. This reveals a potential mismatch: static benchmarks focus on correctness while users prioritize conversational experience.

     \item \textbf{Model-Specific Patterns}: Pipeline systems show elevated robotic complaints (GPT-4o-mini+STT+TTS: 50.6\%), while open-source models struggle with verbosity (Qwen3-Omni: 27.1\%, Voxtral: 23.3\% vs. 17.2\% average). GPT-4o-audio achieves highest satisfaction (4.98) despite highest latency complaints (38.6\%), indicating users tolerate delays for quality.

\end{itemize}

More comprehensive qualitative analysis of human evaluations is provided in Appendix~\ref{appendix:qualitative}. These human preference rankings provide our gold standard for validating benchmark subset selection in the following sections.

\subsection{Alignment with Human Preferences}
\label{sec:subset_validation}
\begin{figure}[t]
\centering
\includegraphics[width=\columnwidth]{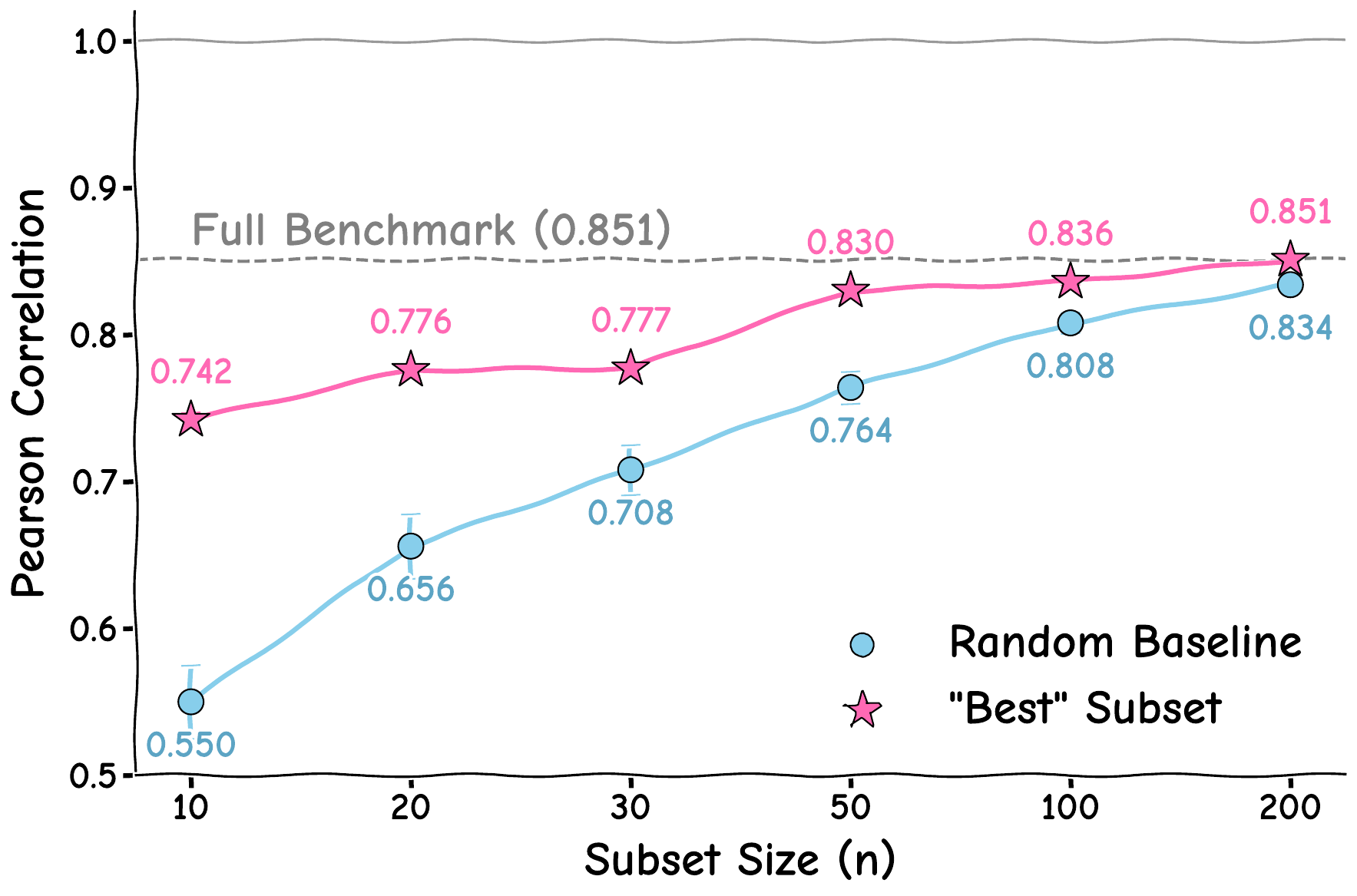}
\caption{\textbf{Benchmark alignment with human preferences.} Pearson correlation between subset scores (averaged over 100 random initializations) and human overall ratings. "Best" Subset: Anchor Points for $n \leq 30$ and Combined Embedding for $n \geq 50$.}
\label{fig:human_correlation}
\end{figure}

To examine alignment between benchmark performance and human preferences, we evaluate the full benchmark, task-balanced random sampling baseline, and the "best" subsets of sizes $n \in \{10, 20, 30, 50, 100, 200\}$ constructed in Section~\ref{sec:subset_selection} (Anchor Points for $n \leq 30$, Combined Embedding for $n \geq 50$) on the 7 models with human preference data, computing Pearson correlations between each method's model scores and human overall satisfaction ratings.

Figure~\ref{fig:human_correlation} presents the results. The full benchmark achieves moderate correlation with human preferences ($r=0.851$). Our "best" subsets approach this ceiling efficiently—the 200-item subset ($\sim$1.3\% of data) matches full correlation—and consistently outperform random sampling, confirming that principled selection which excelled at benchmark score prediction also preserves human preference alignment. Yet this 0.85 ceiling possibly reflects the mismatch from our qualitative analysis of user feedback in Section~\ref{sec:human_results}: benchmarks and users may prioritize different quality dimensions.

\subsection{Predicting Human Preferences}
\label{sec:preference_prediction}
Given the gap between static benchmark scores and human preferences, can we improve human preference prediction with benchmark items so that practitioners could estimate models' likely human reception without costly user studies.

\paragraph{Motivation} Our hypothesis is that human preferences emerge as a composite function of model performance across diverse benchmark dimensions~\cite{schaeffer2025correlating}. If the full benchmark contains these key dimensions, and our selected subsets also capture them—as evidenced by their correlations with human preferences (Section~\ref{sec:subset_validation})—then we can learn to weight benchmark items to better predict overall human satisfaction.

\subsubsection{Prediction Framework}
We employ Ridge regression on both the full benchmark and selected subsets from Section~\ref{sec:subset_selection} to learn the relationship between benchmark performance and human preferences. For each model $m$, let $\mathbf{x}_m \in \mathbb{R}^n$ denote its item-level score vector on the $n$ items in the benchmark or subset, where each score $s_{m,i} \in [0,1]$ is normalized. Let $y_m \in [0,1]$ denote the model's human overall satisfaction rating (linearly rescaled from the original 6-point Likert scale). We learn a linear predictor:
\begin{equation}
\hat{y}_m = \mathbf{w}^\top \mathbf{x}_m + b
\end{equation}
where weights $\mathbf{w} \in \mathbb{R}^n$ and bias $b \in \mathbb{R}$ are learned via Ridge regression with $L_2$ regularization.

\subsubsection{Evaluation Protocol}
We evaluate the effectiveness of regressions on different subsets with leave-one-model-out (LOMO):
\begin{enumerate}[leftmargin=*,noitemsep]
    \item For each held-out model $m_{\text{test}}$:
    \begin{itemize}[noitemsep]
        \item Train Ridge regression on the remaining 6 models' subset scores and human ratings: $\{(\mathbf{x}_{m_i}, y_{m_i})\}_{i \neq \text{test}}$
        \item Select regularization strength $\alpha \in \{10^{-4}, 10^{-3}, \ldots, 10^4\}$ via nested leave-one-out CV with 5 training models, and retrain on all 6 models with the selected $\alpha$
        \item Compute \textbf{Pearson correlation} between predicted scores $\{\hat{y}_{m_i}\}_{i=1}^7$ and actual human ratings $\{y_{m_i}\}_{i=1}^7$ across all models
    \end{itemize}
    \item Average Pearson correlations across all 7 LOMO folds to obtain the final metric
\end{enumerate}

Given our limited pool of 7 models, this protocol maximizes regression training informativeness while ensuring fair comparison: all regression models use identical LOMO evaluation, making relative comparisons fair despite inflated absolute values. While not directly comparable to original benchmark correlations (which involve no training data), this metric reliably ranks regression models' effectively. We further validate in Appendix~\ref{appendix:fair_comparison} that regression on best subsets outperforms original unweighted scores on held-out models for $n>10$.

\subsubsection{Results}
\begin{figure}[t]
\centering
\includegraphics[width=\columnwidth]{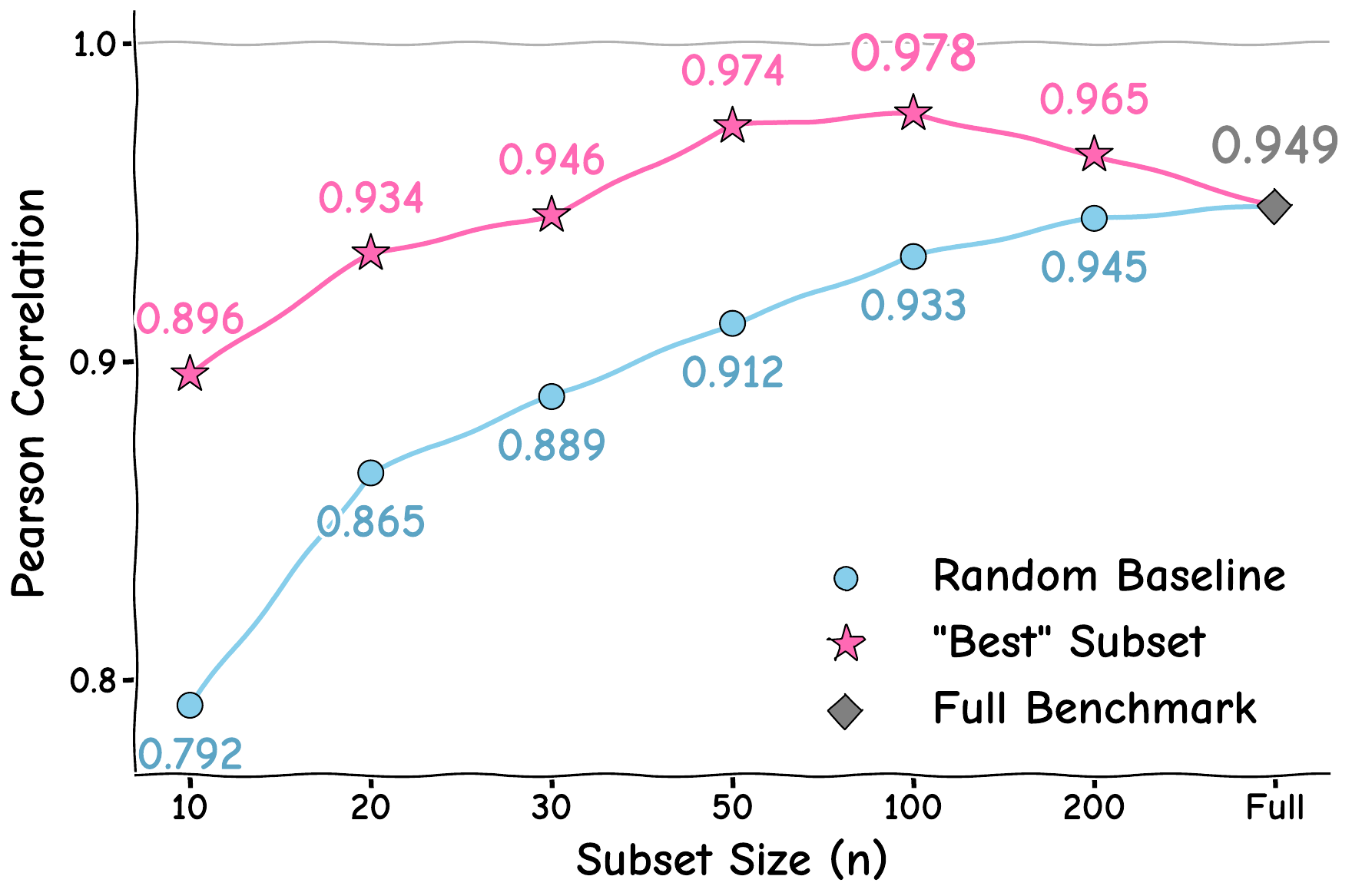}
\caption{\textbf{Human preference prediction via Ridge regression.} Pearson correlation between predictions and actual satisfaction using LOMO CV. See Appendix~\ref{appendix:fair_comparison} for fair comparison excluding in-sample predictions.}
\label{fig:human_prediction}
\end{figure}

Figure~\ref{fig:human_prediction} presents the results:
\begin{itemize}[leftmargin=*,noitemsep]
    \item \textbf{``Best'' subsets perform better}: Principled selection methods consistently outperform random sampling at different subset sizes, demonstrating that they capture more discriminative and informative benchmark items, while random sampling includes redundant or low-signal items.

    \item \textbf{Quality over quantity}: Performance peaks at $n=100$ ($r=0.978$) for ``best'' subset before dropping to $r=0.965$ at $n=200$ and $r=0.949$ for the full benchmark. This non-monotonic trend suggests additional items can introduce lower-informative examples that perturb regression weights learning and degrades generalization. Effective human preference prediction requires high-quality, diverse item selection rather than maximizing evaluation coverage—a well-curated subset of 100 items outpredicts the full 15,964-item benchmark. 
       
\end{itemize}

\section{Open Benchmarks for Practitioners}

To provide practitioners with efficient and ready-to-use evaluation tools, we present HUMANS (HUman-aligned Minimal Audio evaluatioN Subsets). For each subset size, we select the best-performing subset based on highest average Pearson correlation across 100-seed cross-validation, and train final Ridge regression models on all 7 human-evaluated models to predict human preferences. Each subset provides two evaluation modes: (1) \textbf{regression scores} using learned Ridge weights to predict human preference (overall satisfaction, as well as finer-grained dimensions including understanding, naturalness, response quality, and task effectiveness, each with its own regression model), and (2) \textbf{benchmark scores} using the subset selection method's original weights to efficiently approximate full benchmark.

HUMANS benchmark subsets of different sizes with selected items and weights are available at \url{https://huggingface.co/datasets/woodygan/humans-benchmark}.
\paragraph{Recalibration Guide.} HUMANS supports incremental updates as the LAM landscape evolves. For new benchmarks or tasks, subsets can be reselected and regression weights retrained without additional human preference collection---existing ratings remain valid. For substantially new model architectures, the regression can be retrained with incremental human evaluations on the new models. In both cases, the framework supports targeted, low-cost updates rather than full re-evaluation.

\section{Conclusion}

This work addresses the computational challenge of evaluating large audio models through systematic subset selection. Our analysis demonstrates that principled selection methods identify minimal subsets that preserve both benchmark rankings and alignment with human preferences. Qualitative analysis of user feedback reveals critical gaps between what benchmarks measure and what users value, with conversational quality issues dominating complaints. Regression models trained on selected subsets outperform full benchmarks in predicting user satisfaction, showing that quality trumps quantity in evaluation. We release our subsets and human preference ratings to support efficient model comparison and meta-evaluation.

\section*{Limitations}

Our work has several limitations: (1) Our human evaluation focuses on native English speakers from the United States, 
which may not represent the full spectrum of global users, particularly non-native speakers or speakers of other languages. The released HUMANS benchmark is therefore designed for practitioners evaluating LAMs deployed for English-speaking users. While the core framework---that curated subsets efficiently predict both benchmark 
scores and human preferences---should generalize across languages, the specific task design, subset composition, and regression weights would require recalibration for other languages. We recommend future work extending HUMANS to multilingual evaluation by incorporating language-specific tasks into the task pool. (2) Due to budget constraints and the need for statistical power, we evaluated only 7 models with human preferences, limiting the informativeness of correlation analysis and the robustness of our regression models when generalizing to new architectures. This small sample necessitated a LOMO evaluation protocol that may overestimate absolute generalization performance, though relative comparisons between methods remain valid. A larger pool of human-evaluated models would enable more rigorous held-out evaluation and stronger generalization claims. (3) Our benchmark subsets are optimized for conversational scenarios and may not generalize to other audio domains such as music understanding or generation. (4) Our subset selection methods are trained on current LAMs and may face extrapolation challenges when evaluating substantially more capable future models with different capability profiles. (5) While our methods aim at predicting model-level rankings for rapid comparison, they do not predict item-level scores for individual benchmark examples---though methods like IRT and anchor-based selection do support item-level analysis using selected items, as discussed in their original works. Future work could extend our approach to multilingual evaluation, more comprehensive human evaluations with larger model pools, adaptive subset selection for scenarios beyond conversational use cases such as creative audio generation, specialized domain applications, or emerging model capabilities, and item-level diagnostic evaluation for audio models. Additionally, human-aligned subsets and preference scores could also inform training data selection and model fine-tuning.

\section*{Ethical Considerations}
\label{sec:ethics}

We identify potential risks of our work. Our publicly released benchmark subsets could enable models to overfit to specific evaluation items, artificially inflating performance scores without improving real-world capabilities such as privacy protection, fairness across demographic groups, or robustness to production edge cases. Practitioners should not rely solely on our benchmarks for deployment decisions, particularly for applications affecting vulnerable populations. Regarding our human evaluation study, we collected audio recordings and feedback from 776 participants under approval from our institution's Institutional Review Board (IRB). All participants were recruited through Prolific and provided informed consent before recording their voices. Our current analysis focuses exclusively on participant ratings and text feedback. We apply automated filtering to remove personal information from feedback text before analysis. The raw audio recordings are currently stored securely with restricted access. Before any potential data sharing, we will apply noise-masking techniques to reduce voice recognizability and prevent individual identification. Processed audio data will only be made available upon request for research purposes under controlled distribution agreements.

\section*{Acknowledgements}
We appreciate the feedback provided by SALT members. We are thankful for computing support provided by the Stanford HAI-GCP Cloud Credit Grants and OpenAI. This work is funded in part by ONR Grant N000142412532, Sloan Foundation, and NSF grant IIS-2247357.

\bibliography{custom}

\begin{thebibliography}{63}
\providecommand{\natexlab}[1]{#1}

\bibitem[{Abouelenin et~al.(2025)Abouelenin, Ashfaq, Atkinson, Awadalla, Bach, Bao, Benhaim, Cai, Chaudhary, Chen et~al.}]{abouelenin2025phi}
Abdelrahman Abouelenin, Atabak Ashfaq, Adam Atkinson, Hany Awadalla, Nguyen Bach, Jianmin Bao, Alon Benhaim, Martin Cai, Vishrav Chaudhary, Congcong Chen, and 1 others. 2025.
\newblock Phi-4-mini technical report: Compact yet powerful multimodal language models via mixture-of-loras.
\newblock \emph{arXiv preprint arXiv:2503.01743}.

\bibitem[{Bentley et~al.(2018)Bentley, Luvogt, Silverman, Wirasinghe, White, and Lottridge}]{bentley2018understanding}
Frank Bentley, Chris Luvogt, Max Silverman, Rushani Wirasinghe, Brooke White, and Danielle Lottridge. 2018.
\newblock \href {https://doi.org/10.1145/3264901} {Understanding the long-term use of smart speaker assistants}.
\newblock \emph{Proceedings of the ACM on Interactive, Mobile, Wearable and Ubiquitous Technologies}, 2(3):1--24.

\bibitem[{Chen et~al.(2022)Chen, Wang, Chen, Wu, Liu, Chen, Li, Kanda, Yoshioka, Xiao et~al.}]{chen2022wavlm}
Sanyuan Chen, Chengyi Wang, Zhengyang Chen, Yu~Wu, Shujie Liu, Zhuo Chen, Jinyu Li, Naoyuki Kanda, Takuya Yoshioka, Xiong Xiao, and 1 others. 2022.
\newblock Wavlm: Large-scale self-supervised pre-training for full stack speech processing.
\newblock \emph{IEEE Journal of Selected Topics in Signal Processing}, 16(6):1505--1518.

\bibitem[{Chen et~al.(2024)Chen, Yue, Zhang, Gao, Tan, and Li}]{chen2024voicebench}
Yiming Chen, Xianghu Yue, Chen Zhang, Xiaoxue Gao, Robby~T Tan, and Haizhou Li. 2024.
\newblock Voicebench: Benchmarking llm-based voice assistants.
\newblock \emph{arXiv preprint arXiv:2410.17196}.

\bibitem[{Chiang et~al.(2024)Chiang, Zheng, Sheng, Angelopoulos, Li, Li, Zhu, Zhang, Jordan, Gonzalez et~al.}]{chiang2024chatbot}
Wei-Lin Chiang, Lianmin Zheng, Ying Sheng, Anastasios~Nikolas Angelopoulos, Tianle Li, Dacheng Li, Banghua Zhu, Hao Zhang, Michael Jordan, Joseph~E Gonzalez, and 1 others. 2024.
\newblock Chatbot arena: An open platform for evaluating llms by human preference.
\newblock In \emph{Forty-first International Conference on Machine Learning}.

\bibitem[{Coleman et~al.(2020)Coleman, Yeh, Mussmann, Mirzasoleiman, Bailis, Liang, Leskovec, and Zaharia}]{coleman2020selection}
Cody Coleman, Christopher Yeh, Stephen Mussmann, Baharan Mirzasoleiman, Peter Bailis, Percy Liang, Jure Leskovec, and Matei Zaharia. 2020.
\newblock Selection via proxy: Efficient data selection for deep learning.
\newblock In \emph{International Conference on Learning Representations (ICLR)}.

\bibitem[{Dubey et~al.(2024)Dubey, Jauhri, Pandey, Kadian, Al-Dahle, Letman, Mathur, Schelten, Yang, Fan et~al.}]{dubey2024llama}
Abhimanyu Dubey, Abhinav Jauhri, Abhinav Pandey, Abhishek Kadian, Ahmad Al-Dahle, Aiesha Letman, Akhil Mathur, Alan Schelten, Amy Yang, Angela Fan, and 1 others. 2024.
\newblock The llama 3 herd of models.
\newblock \emph{arXiv e-prints}, pages arXiv--2407.

\bibitem[{{DuckDuckGo}(2008)}]{duckduckgo}
{DuckDuckGo}. 2008.
\newblock Duckduckgo search engine.
\newblock \url{https://duckduckgo.com}.
\newblock Accessed: 2025-11-08.

\bibitem[{{Fixie AI}(2024)}]{fixie2024ultravox}
{Fixie AI}. 2024.
\newblock \href {https://github.com/fixie-ai/ultravox} {Ultravox: A fast multimodal llm for real-time voice}.
\newblock Open source multimodal speech model.

\bibitem[{Gao et~al.(2024)Gao, Xia, Xu, Torr, and Gu}]{gao2024benchmarking}
Kuofeng Gao, Shu-Tao Xia, Ke~Xu, Philip Torr, and Jindong Gu. 2024.
\newblock Benchmarking open-ended audio dialogue understanding for large audio-language models.
\newblock \emph{arXiv preprint arXiv:2412.05167}.

\bibitem[{{Gemini Team} et~al.(2023){Gemini Team}, Anil, Borgeaud, Alayrac, Yu, Soricut, Schalkwyk, Dai, Hauth, Millican et~al.}]{team2023gemini}
{Gemini Team}, Rohan Anil, Sebastian Borgeaud, Jean-Baptiste Alayrac, Jiahui Yu, Radu Soricut, Johan Schalkwyk, Andrew~M Dai, Anja Hauth, Katie Millican, and 1 others. 2023.
\newblock Gemini: a family of highly capable multimodal models.
\newblock \emph{arXiv preprint arXiv:2312.11805}.

\bibitem[{{Gemma Team}(2025)}]{gemma_3n_2025}
{Gemma Team}. 2025.
\newblock \href {https://ai.google.dev/gemma/docs/gemma-3n} {Gemma 3n}.

\bibitem[{{Google DeepMind}(2024)}]{google2024gemini_live}
{Google DeepMind}. 2024.
\newblock Gemini {Live} {API} documentation.
\newblock \url{https://ai.google.dev/gemini-api/docs/live}.
\newblock Accessed: 2024-12-22.

\bibitem[{He et~al.(2024)He, Luo, Hu, Zhao, Zhou, Wu, Zhang, Han, Liu, and Sun}]{he2024ultraeval}
Chaoqun He, Renjie Luo, Shengding Hu, Yuanqian Zhao, Jie Zhou, Hanghao Wu, Jiajie Zhang, Xu~Han, Zhiyuan Liu, and Maosong Sun. 2024.
\newblock Ultraeval: A lightweight platform for flexible and comprehensive evaluation for llms.
\newblock \emph{arXiv preprint arXiv:2404.07584}.

\bibitem[{Held et~al.(2025)Held, Ryan, Shrivastava, Khan, Ziems, Li, Bartelds, Sun, Li, Gan, and Yang}]{cava2025}
Will Held, Michael~J. Ryan, Aditya Shrivastava, Ali~Sartaz Khan, Caleb Ziems, Ella Li, Martijn Bartelds, Michael Sun, Tan Li, Woody Gan, and Diyi Yang. 2025.
\newblock \href {https://talkarena.org/cava} {Cava: Comprehensive assessment of voice assistants}.
\newblock \url{https://github.com/SALT-NLP/CAVA}.
\newblock A benchmark for evaluating large audio models (LAMs) capabilities across six domains: turn taking, instruction following, function calling, tone awareness, safety, and latency.

\bibitem[{Huang et~al.(2024{\natexlab{a}})Huang, Chen, Yang, Liu, Li, Lin, Tseng, Diwan, Shih, Shi et~al.}]{huang2024dynamic2}
Chien-yu Huang, Wei-Chih Chen, Shu-wen Yang, Andy~T Liu, Chen-An Li, Yu-Xiang Lin, Wei-Cheng Tseng, Anuj Diwan, Yi-Jen Shih, Jiatong Shi, and 1 others. 2024{\natexlab{a}}.
\newblock Dynamic-superb phase-2: A collaboratively expanding benchmark for measuring the capabilities of spoken language models with 180 tasks.
\newblock \emph{arXiv preprint arXiv:2411.05361}.

\bibitem[{Huang et~al.(2024{\natexlab{b}})Huang, Lu, Wang, Hsiao, Kuan, Wu, Arora, Chang, Shi, Peng et~al.}]{huang2024dynamic}
Chien-yu Huang, Ke-Han Lu, Shih-Heng Wang, Chi-Yuan Hsiao, Chun-Yi Kuan, Haibin Wu, Siddhant Arora, Kai-Wei Chang, Jiatong Shi, Yifan Peng, and 1 others. 2024{\natexlab{b}}.
\newblock Dynamic-superb: Towards a dynamic, collaborative, and comprehensive instruction-tuning benchmark for speech.
\newblock In \emph{ICASSP 2024-2024 IEEE International Conference on Acoustics, Speech and Signal Processing (ICASSP)}, pages 12136--12140. IEEE.

\bibitem[{{ITU-T}(1996)}]{ITU-T-P800-1996}
{ITU-T}. 1996.
\newblock \href {https://www.itu.int/rec/T-REC-P.800-199608-I} {Methods for subjective determination of transmission quality}.
\newblock Recommendation P.800, International Telecommunication Union, Geneva, Switzerland.
\newblock Series P: Telephone Transmission Quality.

\bibitem[{Lalor et~al.(2016)Lalor, Wu, and Yu}]{lalor2016building}
John~P Lalor, Hao Wu, and Hong Yu. 2016.
\newblock Building an evaluation scale using item response theory.
\newblock In \emph{Proceedings of the 2016 Conference on Empirical Methods in Natural Language Processing}, pages 648--657.

\bibitem[{Lalor and Rodriguez(2023)}]{lalor2023py}
John~Patrick Lalor and Pedro Rodriguez. 2023.
\newblock py-irt: A scalable item response theory library for python.
\newblock \emph{INFORMS Journal on Computing}, 35(1):5--13.

\bibitem[{Li et~al.(2025)Li, Held, Ryan, Pipatanakul, Manakul, Zhu, and Yang}]{li2025mind}
Minzhi Li, William~Barr Held, Michael~J Ryan, Kunat Pipatanakul, Potsawee Manakul, Hao Zhu, and Diyi Yang. 2025.
\newblock Mind the gap! static and interactive evaluations of large audio models.
\newblock \emph{arXiv preprint arXiv:2502.15919}.

\bibitem[{Li et~al.(2024)Li, Mondal, Liang, Nghiem, and Boyd-Graber}]{li2024pedantscheapeffectiveinterpretable}
Zongxia Li, Ishani Mondal, Yijun Liang, Huy Nghiem, and Jordan~Lee Boyd-Graber. 2024.
\newblock \href {https://arxiv.org/abs/2402.11161} {Pedants: Cheap but effective and interpretable answer equivalence}.
\newblock \emph{Preprint}, arXiv:2402.11161.

\bibitem[{Liu et~al.(2025)Liu, Ehrenberg, Lo, Denoix, Barreau, Lample, Delignon, Chandu, von Platen, Muddireddy et~al.}]{liu2025voxtral}
Alexander~H Liu, Andy Ehrenberg, Andy Lo, Cl{\'e}ment Denoix, Corentin Barreau, Guillaume Lample, Jean-Malo Delignon, Khyathi~Raghavi Chandu, Patrick von Platen, Pavankumar~Reddy Muddireddy, and 1 others. 2025.
\newblock Voxtral.
\newblock \emph{arXiv preprint arXiv:2507.13264}.

\bibitem[{{LiveKit, Inc.}(2024)}]{livekit2024}
{LiveKit, Inc.} 2024.
\newblock Livekit agents 1.0.
\newblock \url{https://github.com/livekit/agents}.
\newblock Open-source WebRTC infrastructure and agent framework.

\bibitem[{Manakul et~al.(2025)Manakul, Gan, Ryan, Khan, Sirichotedumrong, Pipatanakul, Held, and Yang}]{manakul2025audiojudge}
Potsawee Manakul, Woody~Haosheng Gan, Michael~J Ryan, Ali~Sartaz Khan, Warit Sirichotedumrong, Kunat Pipatanakul, William Held, and Diyi Yang. 2025.
\newblock Audiojudge: Understanding what works in large audio model based speech evaluation.
\newblock \emph{arXiv preprint arXiv:2507.12705}.

\bibitem[{Mart{\'\i}nez-Plumed et~al.(2019)Mart{\'\i}nez-Plumed, Prud{\^e}ncio, Mart{\'\i}nez-Us{\'o}, and Hern{\'a}ndez-Orallo}]{martinez2019item}
Fernando Mart{\'\i}nez-Plumed, Ricardo~BC Prud{\^e}ncio, Adri{\'a} Mart{\'\i}nez-Us{\'o}, and Jos{\'e} Hern{\'a}ndez-Orallo. 2019.
\newblock Item response theory in ai: Analysing machine learning classifiers at the instance level.
\newblock \emph{Artificial intelligence}, 271:18--42.

\bibitem[{{Microsoft}(2025)}]{presidio}
{Microsoft}. 2025.
\newblock \href {https://github.com/microsoft/presidio} {Presidio - data protection and de-identification sdk}.

\bibitem[{Misir(2021)}]{misir2021benchmark}
Mustafa Misir. 2021.
\newblock Benchmark set reduction for cheap empirical algorithmic studies.
\newblock In \emph{2021 IEEE Congress on Evolutionary Computation (CEC)}, pages 1--8. IEEE.

\bibitem[{{OpenAI}(2024{\natexlab{a}})}]{openai2024gpt4o-audio}
{OpenAI}. 2024{\natexlab{a}}.
\newblock \href {https://platform.openai.com/docs/models/gpt-4o-audio-preview} {Gpt-4o audio preview}.
\newblock Accessed: October 18, 2025.

\bibitem[{{OpenAI}(2024{\natexlab{b}})}]{openai2024gpt4o-mini}
{OpenAI}. 2024{\natexlab{b}}.
\newblock {GPT-4o-mini}.
\newblock \url{https://platform.openai.com/docs/models/gpt-4o-mini}.
\newblock Accessed: 2025-10-18.

\bibitem[{{OpenAI}(2024{\natexlab{c}})}]{openai2024gpt4o-mini-audio}
{OpenAI}. 2024{\natexlab{c}}.
\newblock {GPT-4o-mini} audio preview.
\newblock \url{https://platform.openai.com/docs/models/gpt-4o-mini-audio-preview}.
\newblock Accessed: 2025-10-18.

\bibitem[{{OpenAI}(2024{\natexlab{d}})}]{openai2024embeddings}
{OpenAI}. 2024{\natexlab{d}}.
\newblock {text-embedding-3-large}.
\newblock \url{https://platform.openai.com/docs/models/text-embedding-3-large}.
\newblock Accessed: 2025-10-18.

\bibitem[{{OpenAI}(2025{\natexlab{a}})}]{openai2025gpt4o-mini-tts}
{OpenAI}. 2025{\natexlab{a}}.
\newblock {GPT-4o-mini} text-to-speech.
\newblock \url{https://platform.openai.com/docs/guides/text-to-speech}.
\newblock Accessed: 2025-10-18.

\bibitem[{{OpenAI}(2025{\natexlab{b}})}]{openai2025gpt4o-transcribe}
{OpenAI}. 2025{\natexlab{b}}.
\newblock {GPT-4o} speech to text.
\newblock \url{https://platform.openai.com/docs/guides/speech-to-text}.
\newblock Accessed: 2025-10-18.

\bibitem[{{OpenAI}(2025{\natexlab{c}})}]{openai2025gpt5}
{OpenAI}. 2025{\natexlab{c}}.
\newblock Introducing {GPT-5}.
\newblock \url{https://openai.com/index/introducing-gpt-5}.
\newblock Accessed: 2025-10-18.

\bibitem[{{OpenAI}(2025{\natexlab{d}})}]{openai2025gpt-realtime}
{OpenAI}. 2025{\natexlab{d}}.
\newblock Introducing {GPT} realtime.
\newblock \url{https://openai.com/index/introducing-gpt-realtime}.
\newblock Accessed: 2025-10-18.

\bibitem[{{OpenAI}(2025{\natexlab{e}})}]{openai2025gpt52}
{OpenAI}. 2025{\natexlab{e}}.
\newblock \href {https://openai.com/index/gpt-5-system-card-update-gpt-5-2/} {Update to {GPT-5} system card: {GPT-5.2}}.
\newblock Technical report, OpenAI.
\newblock Accessed: 2025-12-22.

\bibitem[{Patil et~al.(2025)Patil, Mao, Yan, Ji, Suresh, Stoica, and Gonzalez}]{patil2025bfcl}
Shishir~G Patil, Huanzhi Mao, Fanjia Yan, Charlie Cheng-Jie Ji, Vishnu Suresh, Ion Stoica, and Joseph~E. Gonzalez. 2025.
\newblock \href {https://openreview.net/forum?id=2GmDdhBdDk} {The berkeley function calling leaderboard ({BFCL}): From tool use to agentic evaluation of large language models}.
\newblock In \emph{Forty-second International Conference on Machine Learning}.

\bibitem[{Perlitz et~al.(2024)Perlitz, Bandel, Gera, Arviv, Ein-Dor, Shnarch, Slonim, Shmueli-Scheuer, and Choshen}]{perlitz2024efficient}
Yotam Perlitz, Elron Bandel, Ariel Gera, Ofir Arviv, Liat Ein-Dor, Eyal Shnarch, Noam Slonim, Michal Shmueli-Scheuer, and Leshem Choshen. 2024.
\newblock Efficient benchmarking (of language models).
\newblock In \emph{Proceedings of the 2024 Conference of the North American Chapter of the Association for Computational Linguistics: Human Language Technologies}, pages 5432--5446.

\bibitem[{Polo et~al.(2024)Polo, Weber, Choshen, Sun, Xu, and Yurochkin}]{polo2024tinybenchmarks}
Felipe~Maia Polo, Lucas Weber, Leshem Choshen, Yuekai Sun, Gongjun Xu, and Mikhail Yurochkin. 2024.
\newblock tinybenchmarks: evaluating llms with fewer examples.
\newblock \emph{arXiv preprint arXiv:2402.14992}.

\bibitem[{Radford et~al.(2023)Radford, Kim, Xu, Brockman, McLeavey, and Sutskever}]{radford2023robust}
Alec Radford, Jong~Wook Kim, Tao Xu, Greg Brockman, Christine McLeavey, and Ilya Sutskever. 2023.
\newblock Robust speech recognition via large-scale weak supervision.
\newblock In \emph{International conference on machine learning}, pages 28492--28518. PMLR.

\bibitem[{Ryan et~al.(2025)Ryan, Zhang, Salunkhe, Chu, Xu, and Yang}]{ryan2025autometrics}
Michael~J Ryan, Yanzhe Zhang, Amol Salunkhe, Yi~Chu, Di~Xu, and Diyi Yang. 2025.
\newblock Autometrics: Approximate human judgements with automatically generated evaluators.
\newblock \emph{arXiv preprint arXiv:2512.17267}.

\bibitem[{Saeki et~al.(2022)Saeki, Xin, Nakata, Koriyama, Takamichi, and Saruwatari}]{saeki2022utmos}
Takaaki Saeki, Detai Xin, Wataru Nakata, Tomoki Koriyama, Shinnosuke Takamichi, and Hiroshi Saruwatari. 2022.
\newblock Utmos: Utokyo-sarulab system for voicemos challenge 2022.
\newblock \emph{arXiv preprint arXiv:2204.02152}.

\bibitem[{Sakshi et~al.(2024)Sakshi, Tyagi, Kumar, Seth, Selvakumar, Nieto, Duraiswami, Ghosh, and Manocha}]{sakshi2024mmau}
S~Sakshi, Utkarsh Tyagi, Sonal Kumar, Ashish Seth, Ramaneswaran Selvakumar, Oriol Nieto, Ramani Duraiswami, Sreyan Ghosh, and Dinesh Manocha. 2024.
\newblock Mmau: A massive multi-task audio understanding and reasoning benchmark.
\newblock \emph{arXiv preprint arXiv:2410.19168}.

\bibitem[{Saon et~al.(2025)Saon, Dekel, Brooks, Nagano, Daniels, Satt, Mittal, Kingsbury, Haws, Morais et~al.}]{saon2025granite}
George Saon, Avihu Dekel, Alexander Brooks, Tohru Nagano, Abraham Daniels, Aharon Satt, Ashish Mittal, Brian Kingsbury, David Haws, Edmilson Morais, and 1 others. 2025.
\newblock Granite-speech: open-source speech-aware llms with strong english asr capabilities.
\newblock \emph{arXiv preprint arXiv:2505.08699}.

\bibitem[{Saranathan et~al.(2025)Saranathan, Xu, Alam, Kumar, Foltin, Wong, and Bhattacharya}]{saranathan2025sublime}
Gayathri Saranathan, Cong Xu, Mahammad~Parwez Alam, Tarun Kumar, Martin Foltin, Soon~Yee Wong, and Suparna Bhattacharya. 2025.
\newblock Sublime: Subset selection via rank correlation prediction for data-efficient llm evaluation.
\newblock In \emph{Proceedings of the 63rd Annual Meeting of the Association for Computational Linguistics (Volume 1: Long Papers)}, pages 30572--30593.

\bibitem[{Schaeffer et~al.(2025)Schaeffer, Koura, Tang, Subramanian, Singh, Mihaylov, Bhargava, Madaan, Chatterji, Goswami et~al.}]{schaeffer2025correlating}
Rylan Schaeffer, Punit~Singh Koura, Binh Tang, Ranjan Subramanian, Aaditya~K Singh, Todor Mihaylov, Prajjwal Bhargava, Lovish Madaan, Niladri~S Chatterji, Vedanuj Goswami, and 1 others. 2025.
\newblock Correlating and predicting human evaluations of language models from natural language processing benchmarks.
\newblock \emph{arXiv preprint arXiv:2502.18339}.

\bibitem[{Silero-Team(2024)}]{silero2024}
Silero-Team. 2024.
\newblock Silero vad: pre-trained enterprise-grade voice activity detector (vad), number detector and language classifier.
\newblock \url{https://github.com/snakers4/silero-vad}.

\bibitem[{Swayamdipta et~al.(2020)Swayamdipta, Schwartz, Lourie, Wang, Hajishirzi, Smith, and Choi}]{swayamdipta2020dataset}
Swabha Swayamdipta, Roy Schwartz, Nicholas Lourie, Yizhong Wang, Hannaneh Hajishirzi, Noah~A Smith, and Yejin Choi. 2020.
\newblock Dataset cartography: Mapping and diagnosing datasets with training dynamics.
\newblock In \emph{Proceedings of the 2020 Conference on Empirical Methods in Natural Language Processing (EMNLP)}, pages 9291--9303.

\bibitem[{Vivek et~al.(2023)Vivek, Ethayarajh, Yang, and Kiela}]{vivek2023anchor}
Rajan Vivek, Kawin Ethayarajh, Diyi Yang, and Douwe Kiela. 2023.
\newblock Anchor points: Benchmarking models with much fewer examples.
\newblock \emph{arXiv preprint arXiv:2309.08638}.

\bibitem[{Wang et~al.(2024)Wang, Zou, Lin, Sun, Liu, Zhang, Liu, Aw, and Chen}]{wang2024audiobench}
Bin Wang, Xunlong Zou, Geyu Lin, Shuo Sun, Zhuohan Liu, Wenyu Zhang, Zhengyuan Liu, AiTi Aw, and Nancy~F Chen. 2024.
\newblock Audiobench: A universal benchmark for audio large language models.
\newblock \emph{arXiv preprint arXiv:2406.16020}.

\bibitem[{Wang et~al.(2023)Wang, Chen, Wu, Zhang, Zhou, Liu, Chen, Liu, Wang, Li et~al.}]{wang2023neural}
Chengyi Wang, Sanyuan Chen, Yu~Wu, Ziqiang Zhang, Long Zhou, Shujie Liu, Zhuo Chen, Yanqing Liu, Huaming Wang, Jinyu Li, and 1 others. 2023.
\newblock Neural codec language models are zero-shot text to speech synthesizers.
\newblock \emph{arXiv preprint arXiv:2301.02111}.

\bibitem[{Xu et~al.(2025{\natexlab{a}})Xu, Guo, He, Hu, He, Bai, Chen, Wang, Fan, Dang et~al.}]{xu2025qwen2}
Jin Xu, Zhifang Guo, Jinzheng He, Hangrui Hu, Ting He, Shuai Bai, Keqin Chen, Jialin Wang, Yang Fan, Kai Dang, and 1 others. 2025{\natexlab{a}}.
\newblock Qwen2. 5-omni technical report.
\newblock \emph{arXiv preprint arXiv:2503.20215}.

\bibitem[{Xu et~al.(2025{\natexlab{b}})Xu, Guo, Hu, Chu, Wang, He, Wang, Shi, He, Zhu et~al.}]{xu2025qwen3}
Jin Xu, Zhifang Guo, Hangrui Hu, Yunfei Chu, Xiong Wang, Jinzheng He, Yuxuan Wang, Xian Shi, Ting He, Xinfa Zhu, and 1 others. 2025{\natexlab{b}}.
\newblock Qwen3-omni technical report.
\newblock \emph{arXiv preprint arXiv:2509.17765}.

\bibitem[{Yang et~al.(2024)Yang, Xu, Liu, Chu, Jiang, Zhou, Leng, Lv, Zhao, Zhou et~al.}]{yang2024air}
Qian Yang, Jin Xu, Wenrui Liu, Yunfei Chu, Ziyue Jiang, Xiaohuan Zhou, Yichong Leng, Yuanjun Lv, Zhou Zhao, Chang Zhou, and 1 others. 2024.
\newblock Air-bench: Benchmarking large audio-language models via generative comprehension.
\newblock \emph{arXiv preprint arXiv:2402.07729}.

\bibitem[{Yao et~al.(2024)Yao, Yu, Zhang, Wang, Cui, Zhu, Cai, Li, Zhao, He et~al.}]{yao2024minicpm}
Yuan Yao, Tianyu Yu, Ao~Zhang, Chongyi Wang, Junbo Cui, Hongji Zhu, Tianchi Cai, Haoyu Li, Weilin Zhao, Zhihui He, and 1 others. 2024.
\newblock Minicpm-v: A gpt-4v level mllm on your phone.
\newblock \emph{arXiv preprint arXiv:2408.01800}.

\bibitem[{Zeng et~al.(2024)Zeng, Du, Liu, Wang, Jiang, Zhao, Dong, and Tang}]{zeng2024glm}
Aohan Zeng, Zhengxiao Du, Mingdao Liu, Kedong Wang, Shengmin Jiang, Lei Zhao, Yuxiao Dong, and Jie Tang. 2024.
\newblock Glm-4-voice: Towards intelligent and human-like end-to-end spoken chatbot.
\newblock \emph{arXiv preprint arXiv:2412.02612}.

\bibitem[{Zhang et~al.(2025{\natexlab{a}})Zhang, Dorner, and Hardt}]{zhang2025benchmark}
Guanhua Zhang, Florian~E Dorner, and Moritz Hardt. 2025{\natexlab{a}}.
\newblock How benchmark prediction from fewer data misses the mark.
\newblock \emph{arXiv preprint arXiv:2506.07673}.

\bibitem[{Zhang et~al.(2025{\natexlab{b}})Zhang, Zhang, Lei, Wu, Jia, and Zhou}]{zhang2025wildspeech}
Jian Zhang, Linhao Zhang, Bokai Lei, Chuhan Wu, Wei Jia, and Xiao Zhou. 2025{\natexlab{b}}.
\newblock Wildspeech-bench: Benchmarking audio llms in natural speech conversation.
\newblock \emph{arXiv preprint arXiv:2506.21875}.

\bibitem[{Zhao et~al.(2024)Zhao, Ren, Hessel, Cardie, Choi, and Deng}]{zhao2024wildchat}
Wenting Zhao, Xiang Ren, Jack Hessel, Claire Cardie, Yejin Choi, and Yuntian Deng. 2024.
\newblock Wildchat: 1m chatgpt interaction logs in the wild.
\newblock \emph{arXiv preprint arXiv:2405.01470}.

\bibitem[{Zheng et~al.(2023)Zheng, Chiang, Sheng, Li, Zhuang, Wu, Zhuang, Li, Lin, Xing et~al.}]{zheng2023lmsys}
Lianmin Zheng, Wei-Lin Chiang, Ying Sheng, Tianle Li, Siyuan Zhuang, Zhanghao Wu, Yonghao Zhuang, Zhuohan Li, Zi~Lin, Eric~P Xing, and 1 others. 2023.
\newblock Lmsys-chat-1m: A large-scale real-world llm conversation dataset.
\newblock \emph{arXiv preprint arXiv:2309.11998}.

\bibitem[{Zhou et~al.(2023)Zhou, Lu, Mishra, Brahma, Basu, Luan, Zhou, and Hou}]{zhou2023instruction}
Jeffrey Zhou, Tianjian Lu, Swaroop Mishra, Siddhartha Brahma, Sujoy Basu, Yi~Luan, Denny Zhou, and Le~Hou. 2023.
\newblock Instruction-following evaluation for large language models.
\newblock \emph{arXiv preprint arXiv:2311.07911}.

\bibitem[{Zou et~al.(2023)Zou, Wang, Carlini, Nasr, Kolter, and Fredrikson}]{zou2023universal}
Andy Zou, Zifan Wang, Nicholas Carlini, Milad Nasr, J~Zico Kolter, and Matt Fredrikson. 2023.
\newblock Universal and transferable adversarial attacks on aligned language models.
\newblock \emph{arXiv preprint arXiv:2307.15043}.

\end{thebibliography}
\clearpage
\appendix
\section{Model Specifications}
\label{appendix:models}
\begin{table*}[t]
\centering
\small
\resizebox{\textwidth}{!}{
\begin{tabular}{lrclll}
\toprule
\textbf{Model} & \textbf{Params} & \textbf{Open} & \textbf{STT} & \textbf{TTS} & \textbf{Ref.} \\
\midrule
\multicolumn{6}{l}{\textit{End-to-End Omni-Modal Systems}} \\
GPT-4o-audio-preview & -- & \ding{55} & -- & -- & \citet{openai2024gpt4o-audio} \\
GPT-4o-mini-audio-preview & -- & \ding{55} & -- & -- & \citet{openai2024gpt4o-mini-audio} \\
GPT-realtime & -- & \ding{55} & -- & -- & \citet{openai2025gpt-realtime} \\
Qwen2.5-Omni-7B & 7B & \ding{51} & -- & -- & \citet{xu2025qwen2} \\
\midrule
\multicolumn{6}{l}{\textit{Speech-to-Text Models (with TTS for audio output)}} \\
Gemini-2.5-Pro+TTS & -- & \ding{55} & -- & GPT-4o-mini-tts & \citet{team2023gemini} \\
Gemini-2.5-Flash+TTS & -- & \ding{55} & -- & GPT-4o-mini-tts & \citet{team2023gemini} \\
Qwen3-Omni-30B-A3B-Instruct-thinker+TTS & 30B & \ding{51} & -- & GPT-4o-mini-tts & \citet{xu2025qwen3} \\
Ultravox-v0.4-ToolACE-8B+TTS & 8B & \ding{51} & -- & GPT-4o-mini-tts & \citet{fixie2024ultravox} \\
Ultravox-v0.5-llama-3.2-1B+TTS & 1B & \ding{51} & -- & GPT-4o-mini-tts & \citet{fixie2024ultravox} \\
Ultravox-v0.6-llama-3.1-8b+TTS & 8B & \ding{51} & -- & GPT-4o-mini-tts & \citet{fixie2024ultravox} \\
Granite-speech-3.3-8b+TTS & 8B & \ding{51} & -- & GPT-4o-mini-tts & \citet{saon2025granite} \\
Voxtral-Small-24B-2507+TTS & 24B & \ding{51} & -- & GPT-4o-mini-tts & \citet{liu2025voxtral} \\
Voxtral-Mini-3B-2507+TTS & 3B & \ding{51} & -- & GPT-4o-mini-tts & \citet{liu2025voxtral} \\
Gemma-3n-e4b+TTS & 4B & \ding{51} & -- & GPT-4o-mini-tts & \citet{gemma_3n_2025} \\
Gemma-3n-e2b+TTS & 2B & \ding{51} & -- & GPT-4o-mini-tts & \citet{gemma_3n_2025} \\
\midrule
\multicolumn{6}{l}{\textit{Pipeline Systems (STT + Text-LLM + TTS)}} \\
GPT-4o-mini+STT+TTS & -- & \ding{55} & GPT-4o-transcribe & GPT-4o-mini-tts & \citet{openai2024gpt4o-mini} \\
GPT-5+STT+TTS & -- & \ding{55} & GPT-4o-transcribe & GPT-4o-mini-tts & \citet{openai2025gpt5} \\
Llama-3.2-3B+STT+TTS & 3B & \ding{51} & GPT-4o-transcribe & GPT-4o-mini-tts & \citet{dubey2024llama} \\
\bottomrule
\end{tabular}
}
\caption{\textbf{Audio model specifications and processing configurations.} 
We evaluate 18 models across three architectural paradigms. End-to-end (E2E) 
systems natively process audio input and generate audio output. Speech-to-text 
(S2T) models encode audio for text-based reasoning and use GPT-4o-mini-tts for 
consistent audio output generation. Pipeline systems combine GPT-4o-transcribe 
for audio input, a text-based LLM for reasoning, and GPT-4o-mini-tts for audio 
output. Qwen3-Omni-30B-A3B-Instruct-thinker uses the thinker module configuration of Qwen3-Omni-30B-A3B-Instruct only. Parameter counts indicate the 
primary model size; dashes indicate proprietary models where parameters are 
not disclosed. \ding{51} indicates open-source models; \ding{55} indicates proprietary models.}
\label{tab:model_details}
\end{table*}
Table~\ref{tab:model_details} provides complete specifications for all 18 
models evaluated in our study. Models are categorized by architecture type, 
with key characteristics including parameter count, and 
public availability.

\subsection{Audio Processing Pipeline}

To ensure fair comparison across models with different native capabilities, 
we standardize audio input/output processing:

\paragraph{Audio Input Processing}
\begin{itemize}
    \item \textbf{End-to-end models}: Process audio directly using native encoders
    \item \textbf{Speech-to-text models}: Process audio directly using native encoders
    \item \textbf{Pipeline text models}: Use GPT-4o-transcribe API for speech-to-text 
    conversion, providing the text transcript to the language model
\end{itemize}

\paragraph{Audio Output Generation}
\begin{itemize}
    \item \textbf{End-to-end models}: Generate audio directly using native decoders
    \item \textbf{Speech-to-text models}: Generate text responses, then synthesize 
    speech using GPT-4o-mini-tts with default voice and settings
    \item \textbf{Pipeline text models}: Generate text responses, then synthesize 
    speech using GPT-4o-mini-tts with default voice and settings
\end{itemize}

\paragraph{Rationale}
Our evaluation adopts a deployer perspective: we standardize on GPT-4o-transcribe 
and GPT-4o-mini-tts—currently state-of-the-art STT/TTS systems—to simulate the 
best-case deployment scenario practitioners can achieve when using each model as 
a reasoning backbone. This isolates models' core intelligence from STT/TTS quality 
variations and reflects real-world practices where developers compose systems from 
best-available components rather than being constrained by a single model's native 
capabilities.
\vspace{1em}
\subsection{Model Selection Rationale}

Our model selection ensures diversity across multiple dimensions to validate 
that our benchmark subsets generalize broadly. We include representatives from 
all major architectural paradigms (4 end-to-end, 11 speech-to-text, 3 pipeline 
systems), spanning model scales from 1B parameters (Ultravox-v0.5-llama-3.2-1B) to large 
proprietary systems (GPT-5, Gemini-2.5-Pro). Our selection balances closed-source 
commercial APIs (7 models) and open-source alternatives (11 models), and covers models released from 
2023-2025 to ensure temporal robustness. This diversity, combined with standardized 
audio processing from the deployer perspective, ensures our benchmark subsets 
provide reliable model evaluations across the current and future LAM landscape.

\section{Benchmark Specifications}
\label{appendix:benchmarks}

We evaluate models on 40 tasks from 5 established audio benchmarks, totaling approximately 16,000 datapoints. Table~\ref{tab:all_tasks} provides complete specifications for all tasks used in our subset selection analysis. All tasks use instructions in English. 
\subsection{Benchmark-Specific Notes}

\paragraph{Multi-dimensional task splitting:} Some tasks measure performance using multiple metrics simultaneously. We treat each metric as a separate evaluation task:
\begin{itemize}
    \item \textbf{CAVA Jeopardy:} Original task measured by both correctness (PEDANT) and latency (response time), split into 2 evaluation tasks
    \item \textbf{WildSpeech-Bench:} All 5 task categories measured by both content quality (GPT-score) and speech quality (UTMOS), each split into 2 evaluation tasks (10 total)
\end{itemize}

\paragraph{Evaluation metrics:}
\begin{itemize}
    \item \textbf{Dynamic-SUPERB Phase 2:} 
    \begin{itemize}
        \item Acc. (LLM): GPT-4o judges whether model answer matches reference
        \item WER: Word Error Rate
        \item PER: Phoneme Error Rate
    \end{itemize}
    
    \item \textbf{CAVA:}
    \begin{itemize}
        \item PEDANT~\citep{li2024pedantscheapeffectiveinterpretable}: QA correctness metric
        \item Latency (s): Response time in seconds
        \item Exact match: String matching accuracy
        \item Function match: Correct function call execution
        \item Refusal rate: Keyword-based refusal detection~\citep{zou2023universal}
        \item IFEval~\citep{zhou2023instruction}: Instruction-following accuracy
        \item LAM-Judge: GPT-4o-audio judges whether response audio matches reference audio in pronunciation
        \item 1-JER: One minus Jaccard Error Rate for speaker diarization
    \end{itemize}
    
    \item \textbf{UltraEval-Audio:}
    \begin{itemize}
        \item ExistMatch: Whether the answer is contained in the response
        \item GPT-score: GPT-4o-mini rates transcribed content quality (1-10 scale)
    \end{itemize}
    
    \item \textbf{SpeakBench:}
    \begin{itemize}
        \item WinRate: Pairwise comparison win rate against GPT-4o-audio using gemini-2.5-flash as AudioJudge
    \end{itemize}
    
    \item \textbf{WildSpeech-Bench:}
    \begin{itemize}
        \item GPT-score: GPT-4o-mini rates transcribed content quality (1-10 scale)
        \item UTMOS~\citep{saeki2022utmos}: Objective speech quality predictor (1-5 scale)
    \end{itemize}
\end{itemize}

\subsection{Dataset Statistics}

Table~\ref{tab:benchmark_stats} summarizes the distribution across benchmarks.

\begin{table}[h]
\centering
\begin{tabular}{lrr}
\toprule
\textbf{Benchmark} & \textbf{Targets} & \textbf{Items} \\
\midrule
Dynamic-SUPERB Phase 2 & 14 & 3,863 \\
CAVA & 11 & 8,321 \\
UltraEval-Audio & 4 & 1,498 \\
SpeakBench & 1 & 82 \\
WildSpeech-Bench & 10 & 2,200 \\
\midrule
\textbf{Total} & 40 & 15,964 \\
\bottomrule
\end{tabular}
\caption{Distribution of tasks and datapoints across benchmarks.}
\label{tab:benchmark_stats}
\end{table}
\onecolumn
{\small
\begin{longtable}{p{3.5cm}p{4.5cm}p{0.7cm}p{0.7cm}p{1cm}p{2cm}p{0.7cm}}
\caption{\textbf{Complete task specifications across all benchmarks.} Normalization column shows how metrics are transformed to [0,1] where 1 represents best performance.}
\label{tab:all_tasks} \\
\toprule
\textbf{Task Name} & \textbf{Description} & \textbf{Input} & \textbf{Output} & \textbf{Metric} & \textbf{Norm.} & \textbf{Items} \\
\midrule
\endfirsthead
\multicolumn{7}{c}%
{{\tablename\ \thetable{} -- continued from previous page}} \\
\toprule
\textbf{Task Name} & \textbf{Description} & \textbf{Input} & \textbf{Output} & \textbf{Metric} & \textbf{Norm.} & \textbf{Items} \\
\midrule
\endhead
\midrule
\multicolumn{7}{r}{{Continued on next page}} \\
\endfoot
\bottomrule
\endlastfoot
\multicolumn{7}{l}{\textit{\textbf{Dynamic-SUPERB Phase-2}}} \\
\midrule
Accent Classification (AccentDB Extended) & 
Identifies regional English accent from speech & 
Audio & 
Text & 
Acc. (LLM) & 
Native [0,1] & 
200 \\
HEAR Language ID (VoxLingua107) & 
Recognizes spoken language from top 10 languages & 
Audio & 
Text & 
Acc. (LLM) & 
Native [0,1] & 
195 \\
Human Non-Speech Sound (Nonspeech7k) & 
Classifies non-speech human vocalizations & 
Audio & 
Text & 
Acc. (LLM) & 
Native [0,1] & 
140 \\
L2 English Accuracy Ranking (speechocean762) & 
Ranks pronunciation accuracy between two L2 speakers & 
Audio & 
Text & 
Acc. (LLM) & 
Native [0,1] & 
360 \\
L2 English Fluency Ranking (speechocean762) & 
Ranks speech fluency between two L2 speakers & 
Audio & 
Text & 
Acc. (LLM) & 
Native [0,1] & 
360 \\
L2 English Prosodic Ranking (speechocean762) & 
Ranks prosodic quality between two L2 speakers & 
Audio & 
Text & 
Acc. (LLM) & 
Native [0,1] & 
360 \\
PoS Estimation (LibriTTS) & 
Predicts part-of-speech tags from audio without transcription & 
Audio & 
Text & 
WER & 
$1-\min(e,1)$ & 
500 \\
SUPERB ASR (LibriSpeech-TestClean) & 
Automatic speech recognition on clean speech & 
Audio & 
Text & 
WER & 
$1-\min(e,1)$ & 
200 \\
SUPERB Emotion Recognition (RAVDESS) & 
Recognizes emotional state from speech & 
Audio & 
Text & 
Acc. (LLM) & 
Native [0,1] & 
240 \\
SUPERB Intent Classification (SLURP-Intent) & 
Identifies user intent from spoken commands & 
Audio & 
Text & 
Acc. (LLM) & 
Native [0,1] & 
200 \\
SUPERB Keyword Spotting (Speech Commands V1) & 
Detects specific keywords in short clips & 
Audio & 
Text & 
Acc. (LLM) & 
Native [0,1] & 
200 \\
SUPERB Phoneme Recognition (LibriSpeech-TestClean) & 
Recognizes phoneme sequences from speech & 
Audio & 
Text & 
PER & 
$1-\min(e,1)$ & 
200 \\
Target Speaker ASR (AMI) & 
Transcribes speech from specific target speaker in multi-speaker audio & 
Audio & 
Text & 
WER & 
$1-\min(e,1)$ & 
500 \\
Voice Disorder Classification (VOICED) & 
Classifies voice pathologies from sustained vowels & 
Audio & 
Text & 
Acc. (LLM) & 
Native [0,1] & 
208 \\

\midrule
\multicolumn{7}{l}{\textit{\textbf{CAVA}}} \\
\midrule
Jeopardy - Correctness (cava\_jeopardy) & 
Answers trivia questions in Jeopardy format & 
Audio & 
Audio & 
PEDANT & 
Native [0,1] & 
1000 \\
Jeopardy - Latency (cava\_jeopardy) & 
Measures response time to the whole answer & 
Audio & 
Audio & 
Latency (s) & 
$1-\min(e,5)/5$ & 
1000 \\
Emotion Recognition (emotion) & 
Identifies counterfactual emotion from speech prosody & 
Audio & 
Text & 
Exact match & 
Native [0,1] & 
1562 \\
Deception Detection (deception\_detection) & 
Identifies deceptive player (werewolf) from game dialogue & 
Audio & 
Text & 
Exact match & 
Native [0,1] & 
151 \\
Function Calling (function\_calling) & 
Executes appropriate function calls with audio input & 
Audio & 
Text & 
Function match & 
Native [0,1] & 
1000 \\
Jailbreak Base (jailbreak\_base) & 
Tests refusal to harmful requests in audio & 
Audio & 
Text & 
Refusal rate & 
Native [0,1] & 
520 \\
Jailbreak Persuasive (jailbreak) & 
Tests refusal to persuasive harmful requests in audio & 
Audio & 
Text & 
Refusal rate & 
Native [0,1] & 
520 \\
Multimodal Instruction Following & 
Follows text instructions when responding to audio request& 
Audio & 
Text & 
IFEval & 
Native [0,1] & 
1000 \\
Pronunciation OED (pronunciation\_oed) & 
Generates correct pronunciation from the oed of the word & 
Text & 
Audio & 
LAM-Judge & 
Native [0,1] & 
284 \\
Pronunciation Audio (pronunciation\_audio) & 
Generates pronunciation from reference audio & 
Audio & 
Audio & 
LAM-Judge & 
Native [0,1] & 
284 \\
Speaker Diarization (speaker\_diarization) & 
Identifies speakers of different sentences in conversation & 
Audio & 
Text & 
1-JER & 
Native [0,1] & 
1000 \\


\midrule
\multicolumn{7}{l}{\textit{\textbf{UltraEval-Audio}}} \\
\midrule
Speech Chatbot (speech-chatbot-alpaca-eval) & 
Speech-to-speech chatbot evaluation & 
Audio & 
Audio & 
GPT-score & 
$(s-1)/9$ & 
198 \\
LLaMA Questions (llama-questions) & 
Question answering from speech & 
Audio & 
Audio & 
ExistMatch & 
Native [0,1] & 
300 \\
Speech Web Questions (speech-web-questions) & 
Web-based question answering from speech & 
Audio & 
Audio & 
ExistMatch & 
Native [0,1] & 
500 \\
Speech TriviaQA (speech-triviaqa) & 
Trivia question answering from speech & 
Audio & 
Audio & 
ExistMatch & 
Native [0,1] & 
500 \\
\midrule


\midrule
\multicolumn{7}{l}{\textit{\textbf{SpeakBench}}} \\
\midrule
SpeakBench (speakbench) & 
Paralinguistic query answering& 
Audio & 
Audio & 
WinRate & 
Native [0,1] & 
82 \\
\midrule


\midrule
\multicolumn{7}{l}{\textit{\textbf{WildSpeech-Bench}}} \\
\midrule
Information Inquiry (Content Quality) & 
Search and obtain information from sources & 
Audio & 
Audio & 
GPT-score & 
$(s-1)/9$ & 
393 \\
Information Inquiry (Speech Quality) & 
Search and obtain information from sources & 
Audio & 
Audio & 
UTMOS & 
$(s-1)/4$ & 
393 \\
Solution Request (Content Quality) & 
Seek action plans for problems & 
Audio & 
Audio & 
GPT-score & 
$(s-1)/9$ & 
351 \\
Solution Request (Speech Quality) & 
Seek action plans for problems & 
Audio & 
Audio & 
UTMOS & 
$(s-1)/4$ & 
351 \\
Text Creation (Content Quality) & 
Create stories, poems, and text & 
Audio & 
Audio & 
GPT-score & 
$(s-1)/9$ & 
192 \\
Text Creation (Speech Quality) & 
Create stories, poems, and text & 
Audio & 
Audio & 
UTMOS & 
$(s-1)/4$ & 
192 \\
Opinion Queries (Content Quality) & 
Ask for opinions on subjective questions & 
Audio & 
Audio & 
GPT-score & 
$(s-1)/9$ & 
64 \\
Opinion Queries (Speech Quality) & 
Ask for opinions on subjective questions & 
Audio & 
Audio & 
UTMOS & 
$(s-1)/4$ & 
64 \\
Paralinguistic-Featured (Content Quality) & 
Handle pause, stress, tone, stuttering, homophones & 
Audio & 
Audio & 
GPT-score & 
$(s-1)/9$ & 
100 \\
Paralinguistic-Featured (Speech Quality) & 
Handle pause, stress, tone, stuttering, homophones & 
Audio & 
Audio & 
UTMOS & 
$(s-1)/4$ & 
100 \\
\midrule


\end{longtable}
}
\twocolumn

\section{Subset Selection Method Details}
\label{appendix:method_details}

\subsection{Random-Sampling-Learn: Complete Algorithm}
\label{appendix:random_sampling_learn}

\paragraph{Training procedure:}
\begin{enumerate}
    \item \textbf{Coreset sampling}: Randomly sample $n$ items from the full benchmark $D$ to form coreset $C \subset D$, using task-balanced probabilities: each item $i$ in task $t$ has probability $p_i = \frac{1}{T \cdot |T_t|}$ where $T$ is the number of tasks and $|T_t|$ is the number of items in task $t$.
    
    \item \textbf{Regression training}: Train a Ridge regression model $g$ on the $M = |\mathcal{M}|$ source models that minimize:
    \begin{equation}
    \frac{1}{M} \sum_{m \in \mathcal{M}} \left( \bar{s}(m, D) - g[s(m, C)] \right)^2 + \lambda \|g\|_2^2
    \end{equation}
    where:
    \begin{itemize}
        \item $\bar{s}(m, D) = \frac{1}{T} \sum_{t=1}^{T} \bar{s}_{m,t}$ is the task-averaged score of source model $m$ on the full benchmark
        \item $s(m, C) \in \mathbb{R}^n$ is the vector of model $m$'s scores on the $n$ coreset items
        \item $\lambda$ is the regularization parameter
    \end{itemize}
    
    \item \textbf{Hyperparameter selection}: The regularization parameter $\lambda$ is selected via 5-fold cross-validation over the set $\{0.001, 0.01, 0.1, 1.0, 10.0, 100.0\}$ using RidgeCV from scikit-learn.
    
    \item \textbf{Target prediction}: For each target model $f$, predict its full benchmark score as:
    \begin{equation}
    h_{\text{Random-Sampling-Learn}}(f) = g[s(f, C)]
    \end{equation}
\end{enumerate}

\subsection{Random-Search-Learn: Complete Algorithm}

\paragraph{Training procedure:}
\begin{enumerate}
    \item \textbf{Train-validation split}: Randomly split the $M$ source models $\mathcal{M}$ into training set $\mathcal{M}_{\text{train}}$ (75\%) and validation set $\mathcal{M}_{\text{val}}$ (25\%).
    
    \item \textbf{Coreset search}: For each iteration $i = 1, \ldots, N$ (where $N=1000$):
    \begin{enumerate}
        \item Sample candidate coreset $C_i \subset D$ with $|C_i| = n$ using task-balanced random sampling
        
        \item Train Ridge regression model $g_i$ on $\mathcal{M}_{\text{train}}$ to predict $\bar{s}(m, D)$ from $s(m, C_i)$, with regularization parameter $\lambda$ selected via cross-validation over $\{0.001, 0.01, 0.1, 1.0, 10.0, 100.0\}$
        
        \item Evaluate mean absolute error $\epsilon_i$ on validation set $\mathcal{M}_{\text{val}}$ by comparing predicted and true full benchmark scores
        
        \item Update best coreset: if $\epsilon_i < \epsilon_{\text{best}}$, set $C^* = C_i$
    \end{enumerate}
    
    \item \textbf{Final model training}: Retrain Ridge regression $g^*$ on all source models $\mathcal{M}$ using the selected coreset $C^*$, with $\lambda$ re-selected via cross-validation.
    
    \item \textbf{Target prediction}: For target model $f$, predict full benchmark score as $h(f) = g^*[s(f, C^*)]$.
\end{enumerate}

\subsection{Variance-Based Selection: Implementation Details}

For each item $i$ in the benchmark:
\begin{enumerate}
    \item Collect scores from all $K$ source models: $\{s_{i,1}, s_{i,2}, \ldots, s_{i,K}\}$
    \item Compute mean score: $\bar{s}_i = \frac{1}{K} \sum_{k=1}^{K} s_{i,k}$
    \item Compute variance: $\sigma_i^2 = \frac{1}{K} \sum_{k=1}^{K} (s_{i,k} - \bar{s}_i)^2$
\end{enumerate}

Sort all items by variance in descending order and select the top $n$ items globally (not per-task). This global selection strategy prioritizes the most discriminative items across the entire benchmark, which may result in unequal task representation compared to task-balanced methods.

\subsection{Difficulty-Based Selection: Implementation Details}

\paragraph{Difficulty computation:}
For each item $i$, difficulty is defined as:
\begin{equation}
D_i = 1 - \frac{1}{K}\sum_{k=1}^{K} s_{i,k}
\end{equation}
where $s_{i,k} \in [0,1]$ is the normalized score of model $k$ on item $i$. Items where most models fail have $D_i$ close to 1, while items where most models succeed have $D_i$ close to 0.

\paragraph{Two-phase stratified sampling:}
\begin{enumerate}
    \item \textbf{Phase 1 - Equal allocation}: 
    \begin{itemize}
        \item Partition all items into $B=10$ difficulty bins based on quantiles: bin $b$ contains items with $D_i \in [(b-1)/B, b/B)$
        \item From each bin $b$, sample $\lfloor n/B \rfloor$ items
        \item Within each bin, use task-balanced probabilities: item $i$ in task $t$ has probability $p_i \propto 1/|T_t|$, normalized to sum to 1 within the bin
    \end{itemize}
    
    \item \textbf{Phase 2 - Remainder allocation}:
    \begin{itemize}
        \item Calculate remainder: $r = n \bmod B$
        \item Re-bin all unsampled items into $\min(r, B)$ bins
        \item Sample one item from each of the first $r$ bins using task-balanced probabilities
    \end{itemize}
\end{enumerate}

This ensures: (1) exactly $n$ items are selected, (2) difficulty distribution is preserved across the full $[0,1]$ range, and (3) task balance is maintained throughout.

\subsection{IRT Implementation Details}
\label{appendix:irt_details}

\subsubsection{IRT Model Specification and Training}

We employ the 5-dimensional two-parameter logistic (M2PL) IRT model with hierarchical Bayesian priors:

\begin{align}
Y_{il} \mid \theta_l, \alpha_i, \beta_i &\sim \text{Bernoulli}(p_{il}) \\
p_{il} &= \sigma(\alpha_i^\top \theta_l - \beta_i) \\
\theta_l &\sim \mathcal{N}(\mu_\theta \mathbf{1}_5, u_\theta^{-1} I_5) \\
\alpha_i &\sim \mathcal{N}(\mu_\alpha \mathbf{1}_5, u_\alpha^{-1} I_5) \\
\beta_i &\sim \mathcal{N}(\mu_\beta, u_\beta^{-1})
\end{align}

with hyperpriors: $\mu_\theta, \mu_\alpha, \mu_\beta \sim \mathcal{N}(0, 10)$ and $u_\theta, u_\alpha, u_\beta \sim \text{Gamma}(1, 1)$.

\paragraph{Training Procedure}
We fit the IRT model using variational inference via the \texttt{py-irt} library~\citep{lalor2023py} on all source models:

\begin{enumerate}
    \item \textbf{Data preparation}: Extract binary responses $Y_{il} \in \{0,1\}$ for all source models and items. For tasks with continuous scores in $[0,1]$, we binarize by finding threshold $c$ such that $\sum_{i,l} Y_{il} \approx \sum_{i,l} \mathbb{1}[Y_{il} \geq c]$ to preserve the overall mean score.
    
    \item \textbf{Model training}: Train the 5-dimensional IRT model with learning rate 0.1 for 500 epochs using the Adam optimizer with fixed random seed for reproducibility.
\end{enumerate}

The resulting model provides point estimates $\hat{\alpha}_i \in \mathbb{R}^5$ and $\hat{\beta}_i \in \mathbb{R}$ for each item, and $\hat{\theta}_l \in \mathbb{R}^5$ for each source model.

\subsubsection{IRT-Based Item Embeddings}

Following~\citet{polo2024tinybenchmarks}, we construct item embeddings by concatenating the IRT parameters:

\begin{equation}
E_i = [\hat{\alpha}_i; \hat{\beta}_i] \in \mathbb{R}^{6}
\end{equation}

where $\hat{\alpha}_i \in \mathbb{R}^5$ is the discrimination parameter vector and $\hat{\beta}_i \in \mathbb{R}$ is the scalar difficulty parameter. This creates a 6-dimensional representation for each item that encodes: (1) which latent abilities are required to answer the item correctly (via $\alpha_i$), and (2) the overall difficulty of the item (via $\beta_i$).

These embeddings have two key advantages over raw correctness vectors:
\begin{itemize}
    \item \textbf{Dimensionality}: The embedding dimension is 6 rather than $K$ (number of source models, often hundreds), reducing the curse of dimensionality in clustering.
    \item \textbf{Stability}: IRT parameters represent latent item properties learned from the entire source model population, making them more stable under distribution shift than individual model responses.
\end{itemize}

\subsubsection{Performance Prediction via p-IRT}

Given a target model $m$ evaluated on the selected anchor points $\mathcal{A} = \{a_1, \ldots, a_n\}$ with responses $\{Y_{a_1,m}, \ldots, Y_{a_n,m}\}$, we estimate its ability parameters $\hat{\theta}_m$ by finding $\theta$ that maximizes the log-likelihood:

\begin{equation}
\sum_{i \in \mathcal{A}} \left[ Y_{im} \log p_{im}(\theta) + (1-Y_{im}) \log(1-p_{im}(\theta)) \right]
\end{equation}

where $p_{im}(\theta) = \sigma(\hat{\alpha}_i^\top \theta - \hat{\beta}_i)$ uses the pre-trained item parameters. We solve this optimization using BFGS initialized at $\theta_0 = \mathbf{0}$.

With $\hat{\theta}_m$ estimated, we compute the p-IRT performance estimate using task-averaged scores where observed items use their actual responses and unseen items use IRT predictions:

\begin{equation}
\text{Score}(m) = \frac{1}{T} \sum_{t=1}^T \frac{\sum_{i \in I_t} b_i \cdot s_{im}}{\sum_{i \in I_t} b_i}
\end{equation}

where $b_i$ is the task balance weight and:

\begin{equation}
s_{im} = \begin{cases}
Y_{im} & \text{if } i \in \mathcal{A} \\
\hat{p}_{im} & \text{if } i \notin \mathcal{A}
\end{cases}
\end{equation}

with $\hat{p}_{im} = \sigma(\hat{\alpha}_i^\top \hat{\theta}_m - \hat{\beta}_i)$ being the IRT-predicted probability of correctness for unseen item $i$.

This formulation is theoretically principled: if the IRT model perfectly captures the data-generating process, then $\mathbb{E}[\hat{p}_{im}] = \mathbb{E}[Y_{im}]$, making our estimate an unbiased estimator of the true full benchmark score. The estimator directly replaces missing observations with their conditional expectations given the observed data, which is the optimal prediction under mean squared error. This approach ensures that each task contributes equally to the final score through the balance weights $b_i$, and leverages cross-task information through $\hat{\theta}_m$ to improve predictions even with sparse per-task observations.
\subsection{Anchor-Based Selection Details}
\label{appendix:anchor_details}

\subsubsection{Task-Aware Weighted K-Means Clustering}

Our implementation adapts the anchor points framework from \citet{vivek2023anchor} to handle multi-task audio benchmarks. The framework uses weighted K-Means clustering on item embeddings, which can be source model score vectors (original anchor points) or alternative representations (acoustic, semantic, or combined embeddings).

\paragraph{Task-Aware Weighting}
Each item $i$ in task $t$ receives balance weight:
\begin{equation}
b_i = \frac{1}{T \cdot |T_t|}
\end{equation}
where $T$ is the number of tasks and $|T_t|$ is the number of items in task $t$. These weights are normalized to sum to 1 and ensure each task contributes equally to the clustering and anchor selection regardless of its size.

\paragraph{Anchor Selection via Weighted K-Means}
We perform weighted K-Means clustering on item embeddings with $k=n$ clusters:
\begin{equation}
\min_{\{C_1,...,C_n\}} \sum_{i=1}^{n} \sum_{x_j \in C_i} b_j \|x_j - \mu_i\|^2
\end{equation}
where $\mu_i$ is the weighted centroid of cluster $C_i$ and $x_j \in \mathbb{R}^D$ is item $j$'s embedding vector (dimensionality $D$ depends on the embedding choice). Each centroid is mapped to its nearest real datapoint using Euclidean distance to select the $n$ anchor points:
\begin{equation}
a_i = \argmin_{j: x_j \in C_i} \|x_j - \mu_i\|^2
\end{equation}

\paragraph{Cluster Weights}
For anchor point $i$ representing cluster $C_i$, the weight is the sum of balance weights in that cluster:
\begin{equation}
w_i = \sum_{j \in C_i} b_j
\end{equation}
Since balance weights sum to 1 across all items, cluster weights automatically sum to 1: $\sum_{i=1}^{n} w_i = 1$. This maintains task balance in the final APW score—clusters containing more items or items from underrepresented tasks receive proportionally higher weights.

\paragraph{Differences from Original Anchor Points}
Our method differs from \citet{vivek2023anchor} in three key ways:

\begin{enumerate}
    \item \textbf{Distance metric}: We use Euclidean distance on normalized embeddings instead of correlation-based distances. Since all audio metrics are pre-normalized to $[0,1]$, Euclidean distance effectively captures performance similarity without requiring correlation computation or logit transforms.
    
    \item \textbf{Clustering algorithm}: We use weighted K-Means instead of K-Medoids (PAM). K-Means provides native sample weight support in scikit-learn, enabling efficient task-aware clustering with $O(n \cdot D \cdot K \cdot I)$ complexity where $I < 100$ iterations. We map centroids to nearest datapoints post-hoc rather than constraining medoids during optimization.
    
    \item \textbf{Task awareness}: We introduce task-based balance weights for multi-task benchmarks, ensuring equal task contribution regardless of dataset size. The original method assumed single-task datasets where uniform weighting suffices.
\end{enumerate}

\section{Complete Subset Selection Results}
\label{appendix:complete_results}

\subsection{Correlation Curves for All Methods}
\label{appendix:all_method_curves}

\begin{figure}[h]
\centering
\includegraphics[width=\columnwidth]{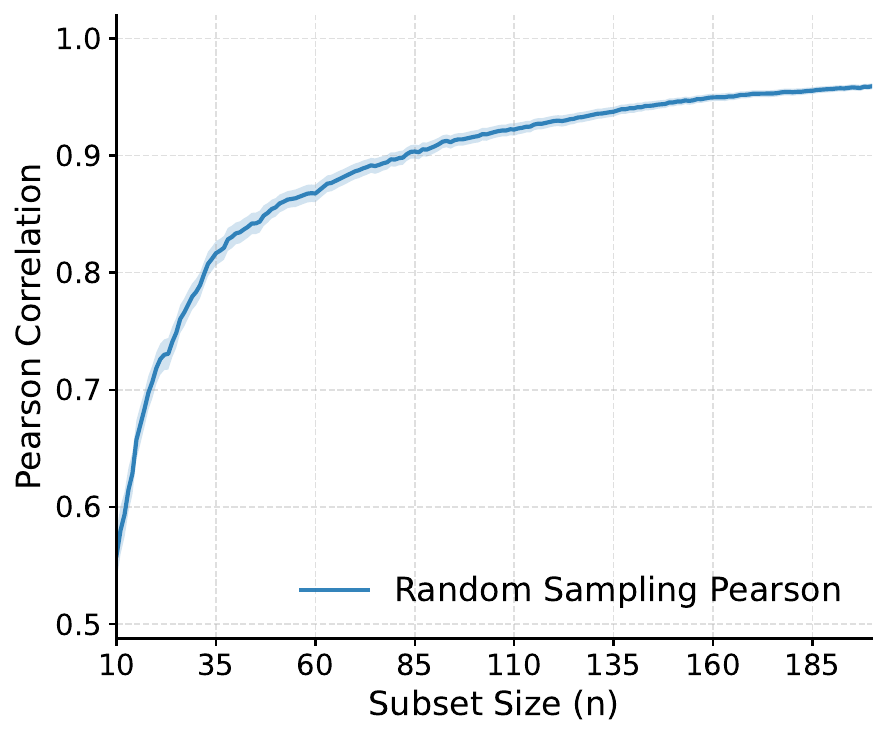}
\caption{\textbf{Random Sampling (Pearson).} AUCC=0.891, $N_{90}=83$, $N_{95}=164$.}
\label{fig:correlation_random}
\end{figure}

\begin{figure}[h]
\centering
\includegraphics[width=\columnwidth]{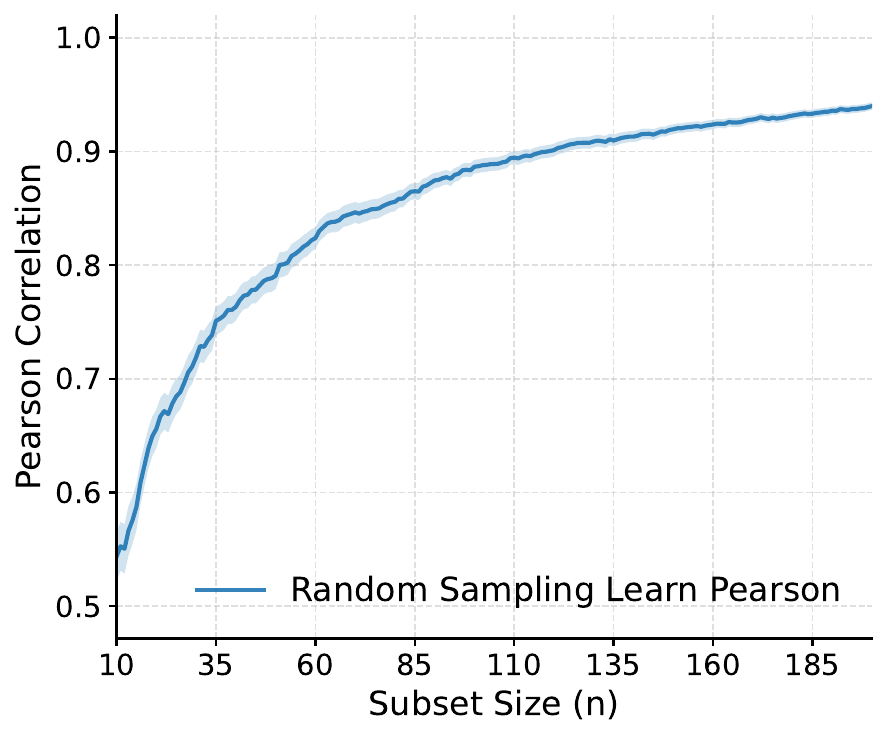}
\caption{\textbf{Random-Sampling-Learn (Pearson).} AUCC=0.854, $N_{90}=119$, $N_{95}=300$.}
\label{fig:correlation_random_learn}
\end{figure}

\begin{figure}[h]
\centering
\includegraphics[width=\columnwidth]{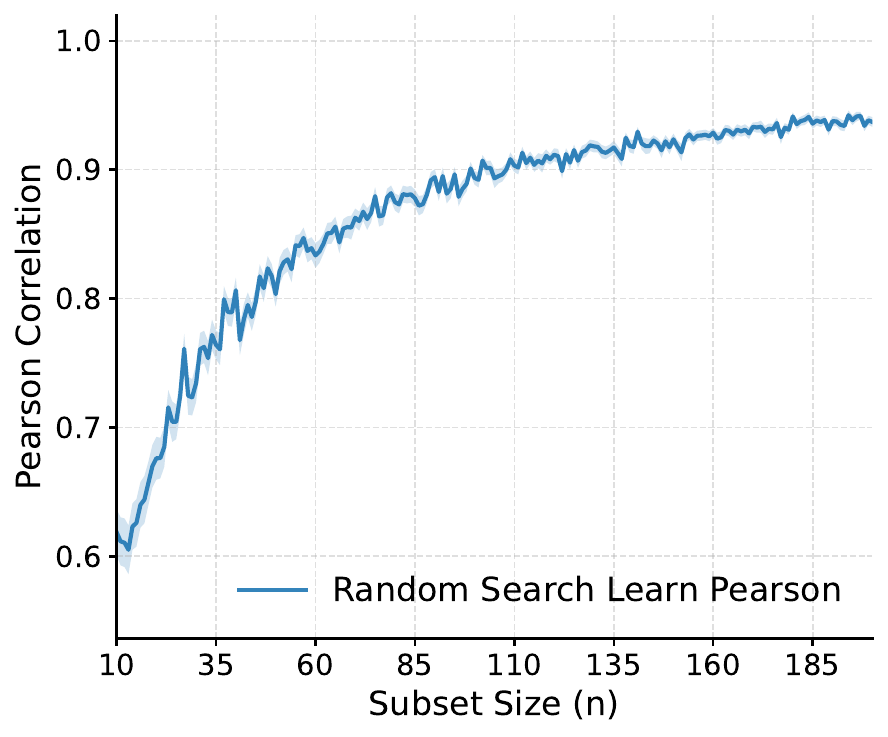}
\caption{\textbf{Random-Search-Learn (Pearson).} AUCC=0.866, $N_{90}=99$, $N_{95}=300$.}
\label{fig:correlation_random_search}
\end{figure}

\begin{figure}[h]
\centering
\includegraphics[width=\columnwidth]{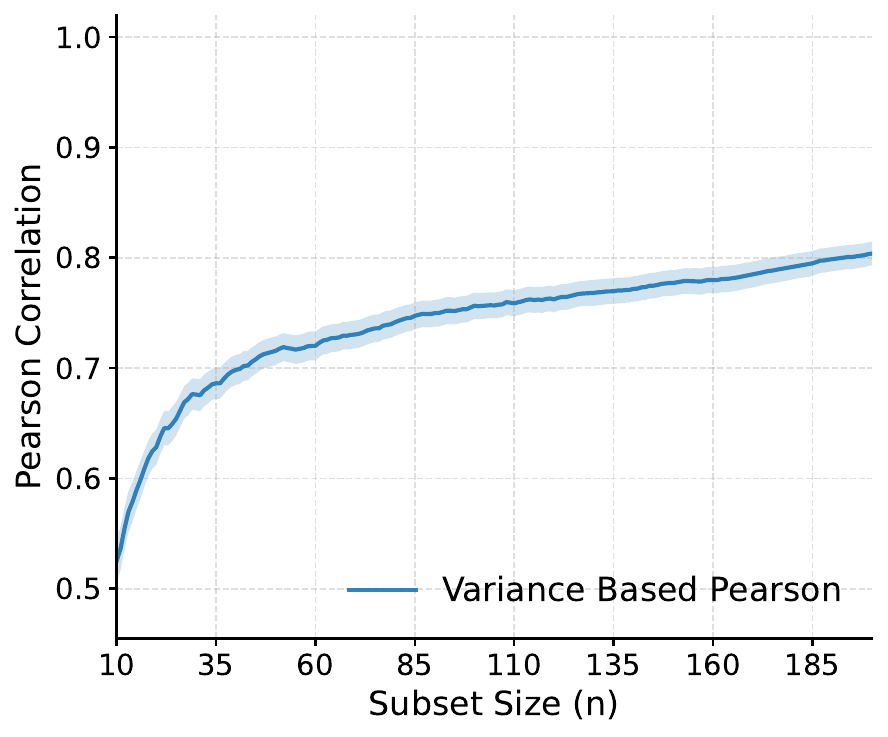}
\caption{\textbf{Variance-based Selection (Pearson).} AUCC=0.742, $N_{90}$=--, $N_{95}$=--.}
\label{fig:correlation_variance}
\end{figure}

\begin{figure}[h]
\centering
\includegraphics[width=\columnwidth]{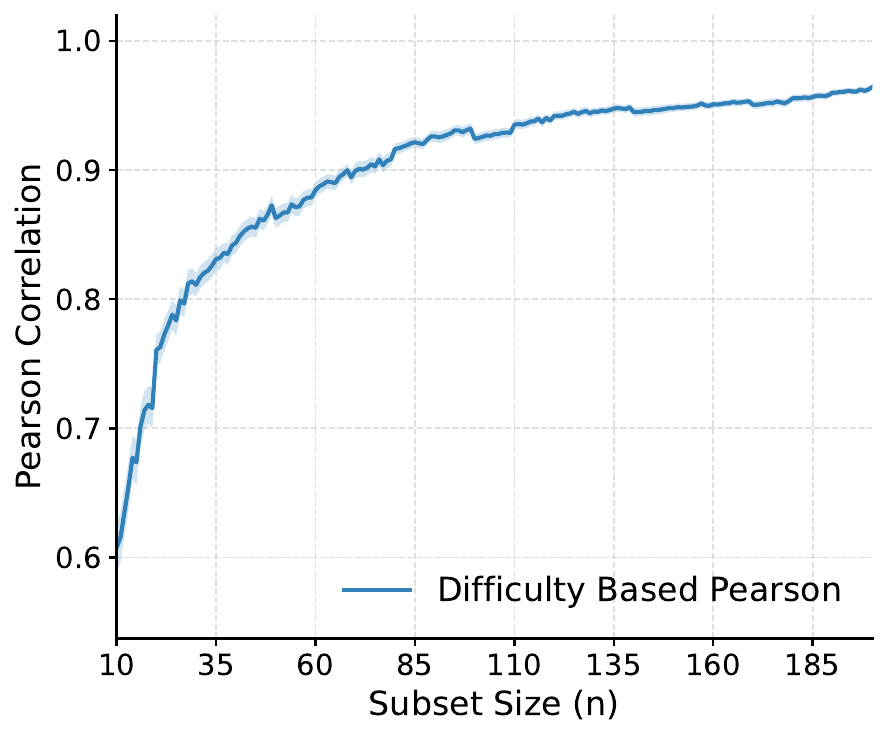}
\caption{\textbf{Difficulty-based Selection (Pearson).} AUCC=0.902, $N_{90}=71$, $N_{95}=157$.}
\label{fig:correlation_oracle}
\end{figure}

\begin{figure}[h]
\centering
\includegraphics[width=\columnwidth]{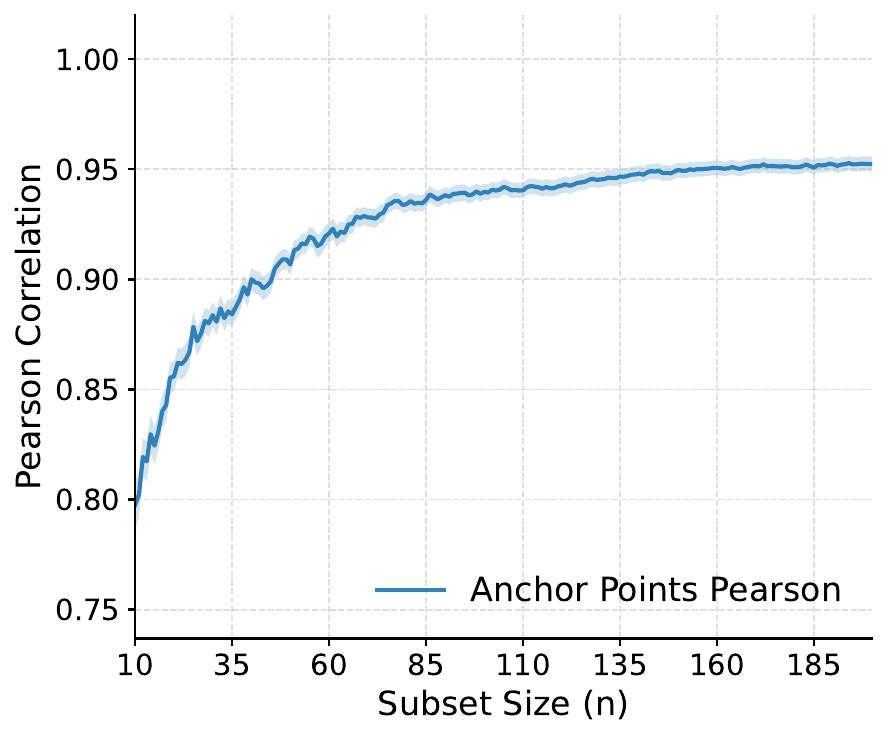}
\caption{\textbf{Anchor Points (Pearson).} AUCC=0.927, $N_{90}=40$, $N_{95}=155$.}
\label{fig:correlation_anchor}
\end{figure}

\begin{figure}[h]
\centering
\includegraphics[width=\columnwidth]{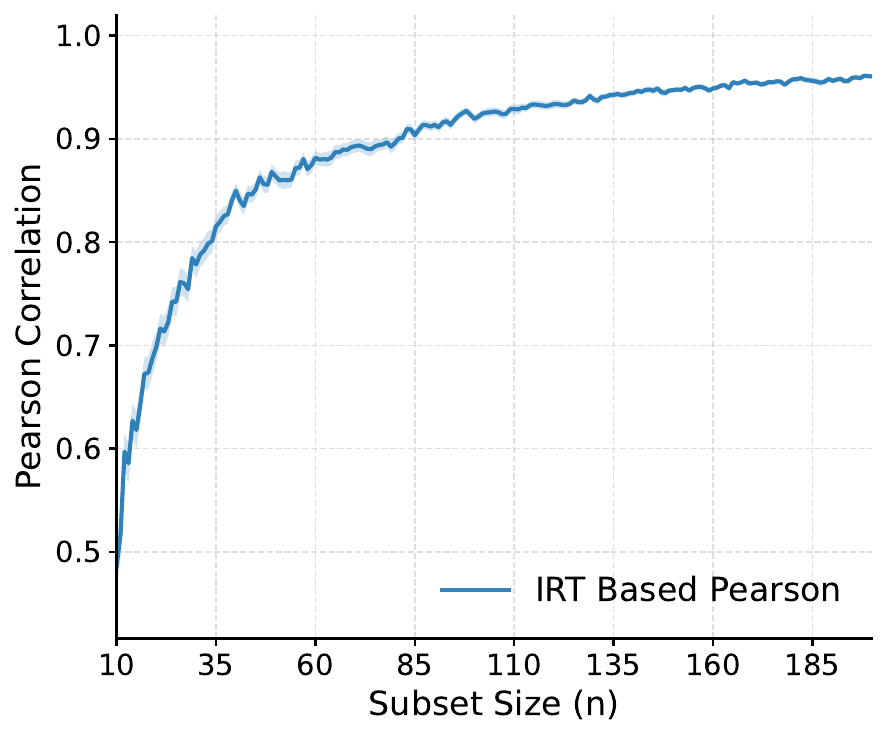}
\caption{\textbf{IRT + Anchor Points (Pearson).} AUCC=0.892, $N_{90}=81$, $N_{95}=156$.}
\label{fig:correlation_irt}
\end{figure}

\begin{figure}[h]
\centering
\includegraphics[width=\columnwidth]{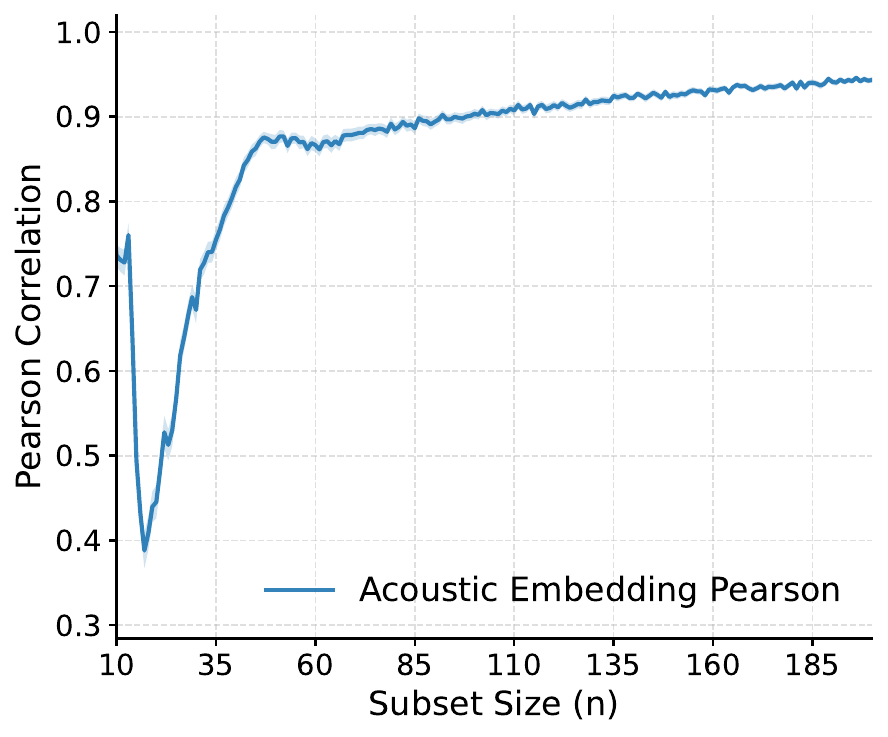}
\caption{\textbf{Acoustic Embedding (Pearson).} AUCC=0.850, $N_{90}=92$, $N_{95}=250$.}
\label{fig:correlation_acoustic}
\end{figure}

\begin{figure}[h]
\centering
\includegraphics[width=\columnwidth]{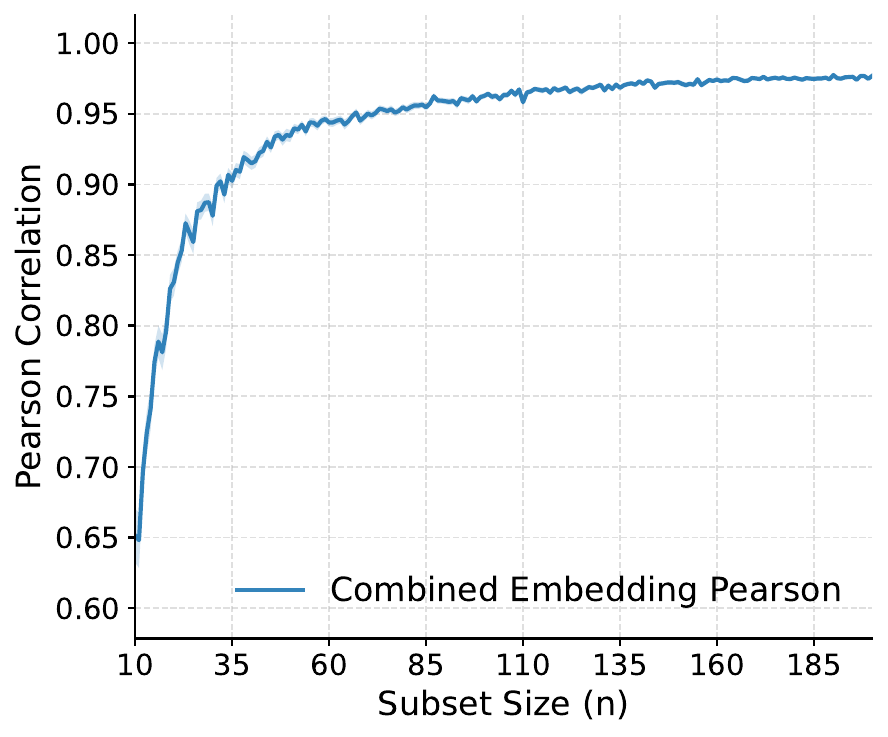}
\caption{\textbf{Combined Embedding (Pearson).} AUCC=0.943, $N_{90}=32$, $N_{95}=67$.}
\label{fig:correlation_combined}
\end{figure}

Figures~\ref{fig:correlation_random}--\ref{fig:correlation_combined} show detailed correlation curves with confidence intervals for all subset selection methods evaluated in this work.

\subsection{Alternative Correlation Metrics}
\label{appendix:alternative_correlations}

Table~\ref{tab:spearman_correlation} and Table~\ref{tab:kendall_correlation} report Spearman and Kendall correlations respectively, complementing the Pearson results in the main text. All three metrics show consistent trends, with Combined Embedding achieving the best overall performance and Anchor Points excelling at small subset sizes.

\begin{table*}[t]
\centering
\small
\resizebox{\textwidth}{!}{%
\begin{tabular}{lcccccccc}
\toprule
\textbf{Method} & \multicolumn{6}{c}{\textbf{Spearman Correlation by Subset Size}} & \textbf{AUCC} & $N_{90}$ / $N_{95}$ \\
\cmidrule(lr){2-7} \cmidrule(lr){8-8} \cmidrule(lr){9-9}
& $n=10$ & $n=20$ & $n=30$ & $n=50$ & $n=100$ & $n=200$ & $[10,200]$ & \\
\midrule
Random Sampling & 0.518$_{\pm\text{0.023}}$ & 0.661$_{\pm\text{0.016}}$ & 0.735$_{\pm\text{0.014}}$ & 0.798$_{\pm\text{0.011}}$ & 0.870$_{\pm\text{0.007}}$ & 0.918$_{\pm\text{0.004}}$ & 0.844 & 145 / 300 \\
Random-Sampling-Learn & 0.519$_{\pm\text{0.023}}$ & 0.636$_{\pm\text{0.018}}$ & 0.682$_{\pm\text{0.016}}$ & 0.747$_{\pm\text{0.014}}$ & 0.846$_{\pm\text{0.009}}$ & 0.903$_{\pm\text{0.006}}$ & 0.814 & 200 / 500 \\
Random-Search-Learn & 0.573$_{\pm\text{0.019}}$ & 0.635$_{\pm\text{0.018}}$ & 0.694$_{\pm\text{0.017}}$ & 0.756$_{\pm\text{0.013}}$ & 0.846$_{\pm\text{0.008}}$ & 0.899$_{\pm\text{0.006}}$ & 0.823 & 200 / 500 \\
Variance-based & 0.497$_{\pm\text{0.019}}$ & 0.591$_{\pm\text{0.017}}$ & 0.638$_{\pm\text{0.015}}$ & 0.678$_{\pm\text{0.014}}$ & 0.718$_{\pm\text{0.013}}$ & 0.766$_{\pm\text{0.012}}$ & 0.707 & -- / -- \\
Difficulty-based & 0.566$_{\pm\text{0.021}}$ & 0.710$_{\pm\text{0.015}}$ & 0.759$_{\pm\text{0.012}}$ & 0.807$_{\pm\text{0.010}}$ & 0.873$_{\pm\text{0.006}}$ & 0.916$_{\pm\text{0.005}}$ & 0.857 & 110 / 450 \\
IRT-based & 0.462$_{\pm\text{0.020}}$ & 0.655$_{\pm\text{0.016}}$ & 0.722$_{\pm\text{0.015}}$ & 0.804$_{\pm\text{0.011}}$ & 0.866$_{\pm\text{0.007}}$ & 0.903$_{\pm\text{0.007}}$ & 0.834 & 176 / 800 \\
Anchor Points & \textbf{0.769$_{\pm\text{0.013}}$} & \textbf{0.831$_{\pm\text{0.009}}$} & \textbf{0.858$_{\pm\text{0.008}}$} & \textbf{0.879$_{\pm\text{0.007}}$} & \textbf{0.897$_{\pm\text{0.005}}$} & \underline{0.926$_{\pm\text{0.005}}$} & \textbf{0.896} & \textbf{52} / \textbf{--} \\
Semantic Embedding & 0.455$_{\pm\text{0.022}}$ & 0.567$_{\pm\text{0.018}}$ & 0.733$_{\pm\text{0.014}}$ & 0.829$_{\pm\text{0.010}}$ & 0.886$_{\pm\text{0.006}}$ & 0.892$_{\pm\text{0.006}}$ & 0.817 & 200 / -- \\
Acoustic Embedding$^\dagger$ & \underline{0.666$_{\pm\text{0.015}}$} & \underline{0.426$_{\pm\text{0.020}}$} & \underline{0.634$_{\pm\text{0.018}}$} & \underline{0.827$_{\pm\text{0.009}}$} & \underline{0.858$_{\pm\text{0.008}}$} & 0.888$_{\pm\text{0.006}}$ & \underline{0.807} & 180 / -- \\
Combined Embedding$^\dagger$ & 0.615$_{\pm\text{0.019}}$ & 0.787$_{\pm\text{0.011}}$ & 0.824$_{\pm\text{0.011}}$ & 0.889$_{\pm\text{0.007}}$ & 0.918$_{\pm\text{0.005}}$ & \textbf{0.939$_{\pm\text{0.004}}$} & 0.901 & \underline{55} / \underline{300} \\
\bottomrule
\end{tabular}
}
\caption{\textbf{Spearman correlation between subset and full benchmark rankings.}}
\label{tab:spearman_correlation}
\end{table*}

\begin{table*}[t]
\centering
\small
\resizebox{\textwidth}{!}{%
\begin{tabular}{lcccccccc}
\toprule
\textbf{Method} & \multicolumn{6}{c}{\textbf{Kendall Correlation by Subset Size}} & \textbf{AUCC} & $N_{80}$ / $N_{90}$ \\
\cmidrule(lr){2-7} \cmidrule(lr){8-8} \cmidrule(lr){9-9}
& $n=10$ & $n=20$ & $n=30$ & $n=50$ & $n=100$ & $n=200$ & $[10,200]$ & \\
\midrule
Random Sampling & 0.427$_{\pm\text{0.019}}$ & 0.544$_{\pm\text{0.015}}$ & 0.623$_{\pm\text{0.014}}$ & 0.692$_{\pm\text{0.012}}$ & 0.771$_{\pm\text{0.010}}$ & 0.848$_{\pm\text{0.008}}$ & 0.751 & 127 / \underline{500} \\
Random-Sampling-Learn & 0.428$_{\pm\text{0.020}}$ & 0.528$_{\pm\text{0.016}}$ & 0.576$_{\pm\text{0.015}}$ & 0.637$_{\pm\text{0.014}}$ & 0.750$_{\pm\text{0.011}}$ & 0.829$_{\pm\text{0.009}}$ & 0.720 & 158 / 900 \\
Random-Search-Learn & 0.480$_{\pm\text{0.017}}$ & 0.540$_{\pm\text{0.017}}$ & 0.582$_{\pm\text{0.015}}$ & 0.651$_{\pm\text{0.013}}$ & 0.745$_{\pm\text{0.010}}$ & 0.819$_{\pm\text{0.009}}$ & 0.728 & 141 / -- \\
Variance-based & 0.415$_{\pm\text{0.019}}$ & 0.487$_{\pm\text{0.016}}$ & 0.521$_{\pm\text{0.015}}$ & 0.555$_{\pm\text{0.014}}$ & 0.601$_{\pm\text{0.014}}$ & 0.648$_{\pm\text{0.013}}$ & 0.591 & -- / -- \\
Difficulty-based & 0.465$_{\pm\text{0.018}}$ & 0.595$_{\pm\text{0.015}}$ & 0.643$_{\pm\text{0.013}}$ & 0.704$_{\pm\text{0.012}}$ & 0.775$_{\pm\text{0.009}}$ & 0.841$_{\pm\text{0.008}}$ & 0.766 & 93 / \underline{500} \\
IRT-based & 0.364$_{\pm\text{0.017}}$ & 0.544$_{\pm\text{0.016}}$ & 0.611$_{\pm\text{0.015}}$ & 0.696$_{\pm\text{0.012}}$ & 0.768$_{\pm\text{0.010}}$ & 0.821$_{\pm\text{0.009}}$ & 0.736 & 142 / 800 \\
Anchor Points & \textbf{0.659$_{\pm\text{0.014}}$} & \textbf{0.731$_{\pm\text{0.011}}$} & \textbf{0.762$_{\pm\text{0.010}}$} & \underline{0.790$_{\pm\text{0.010}}$} & \underline{0.818$_{\pm\text{0.009}}$} & \underline{0.861$_{\pm\text{0.008}}$} & \underline{0.816} & \underline{55} / -- \\
Semantic Embedding & 0.371$_{\pm\text{0.019}}$ & 0.460$_{\pm\text{0.016}}$ & 0.619$_{\pm\text{0.014}}$ & 0.719$_{\pm\text{0.012}}$ & 0.800$_{\pm\text{0.009}}$ & 0.802$_{\pm\text{0.009}}$ & 0.723 & 99 / -- \\
Acoustic Embedding & \underline{0.557$_{\pm\text{0.015}}$} & 0.330$_{\pm\text{0.017}}$ & 0.528$_{\pm\text{0.016}}$ & 0.719$_{\pm\text{0.011}}$ & 0.758$_{\pm\text{0.010}}$ & 0.797$_{\pm\text{0.009}}$ & 0.709 & 166 / -- \\
Combined Embedding & 0.523$_{\pm\text{0.017}}$ & \underline{0.680$_{\pm\text{0.013}}$} & \underline{0.722$_{\pm\text{0.012}}$} & \textbf{0.802$_{\pm\text{0.009}}$} & \textbf{0.844$_{\pm\text{0.008}}$} & \textbf{0.879$_{\pm\text{0.007}}$} & \textbf{0.826} & \textbf{49} / \textbf{350} \\
\bottomrule
\end{tabular}
}
\caption{\textbf{Kendall correlation between subset and full benchmark rankings.} We report $N_{80}$ and $N_{90}$ thresholds (instead of $N_{90}$ and $N_{95}$) as Kendall's $\tau$ is inherently more conservative than Pearson's $r$ and Spearman's $\rho$, making higher thresholds difficult to achieve.}
\label{tab:kendall_correlation}
\end{table*}
\clearpage
\section{Task Distribution Analysis}
\label{appendix:task_distribution}
\begin{figure*}[h]
\centering
\includegraphics[width=\textwidth]{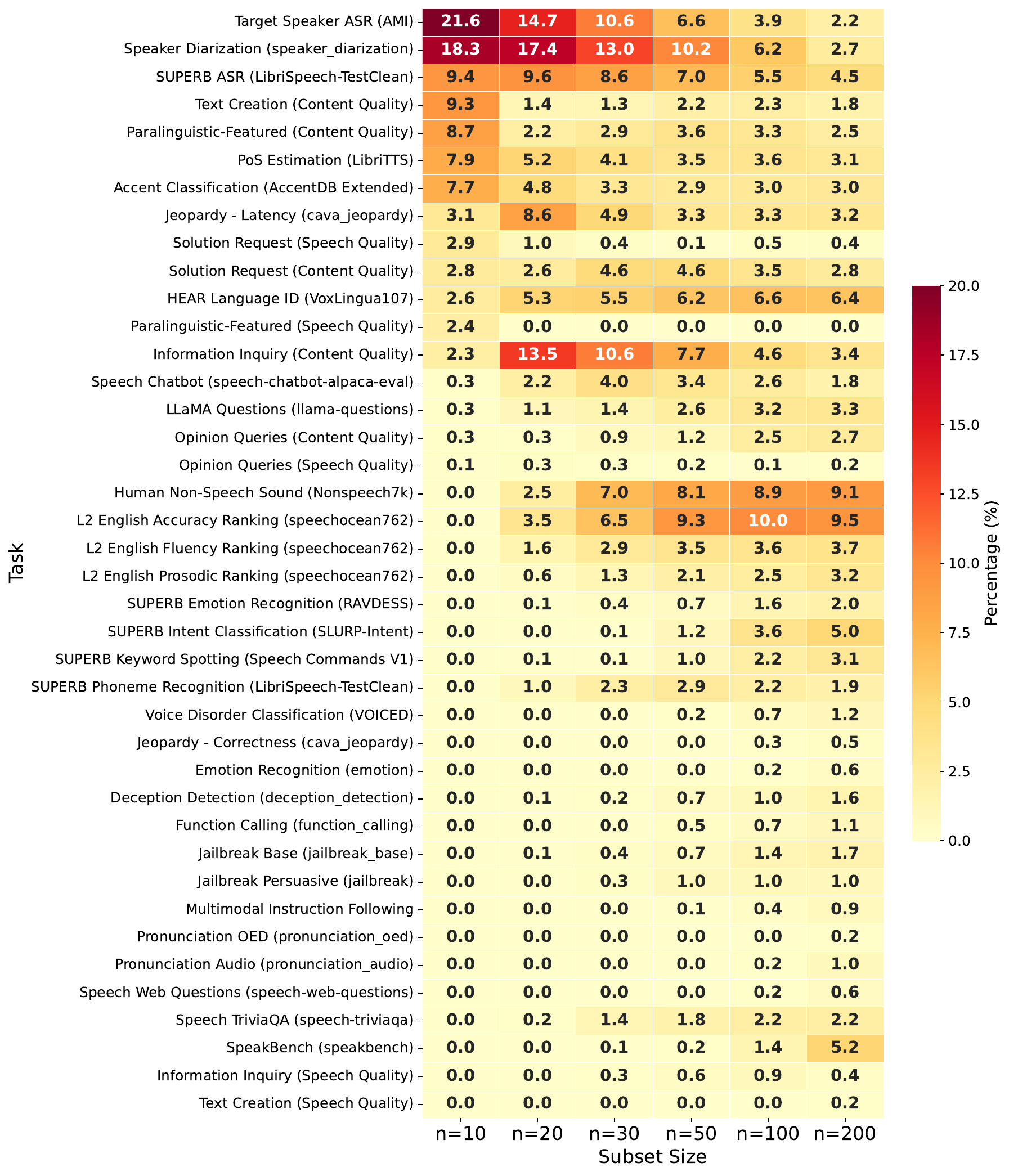}
\caption{\textbf{Task distribution for Anchor Points method.} 
Heatmap shows the percentage of items from each task in subsets of 
varying sizes (n=10 to n=200), averaged across 100 random seeds. 
Darker red indicates higher representation. Tasks are ordered by 
their representation at n=10.}
\label{fig:anchor_point_task_heatmap}
\end{figure*}

\begin{figure*}[h]
\centering
\includegraphics[width=\textwidth]{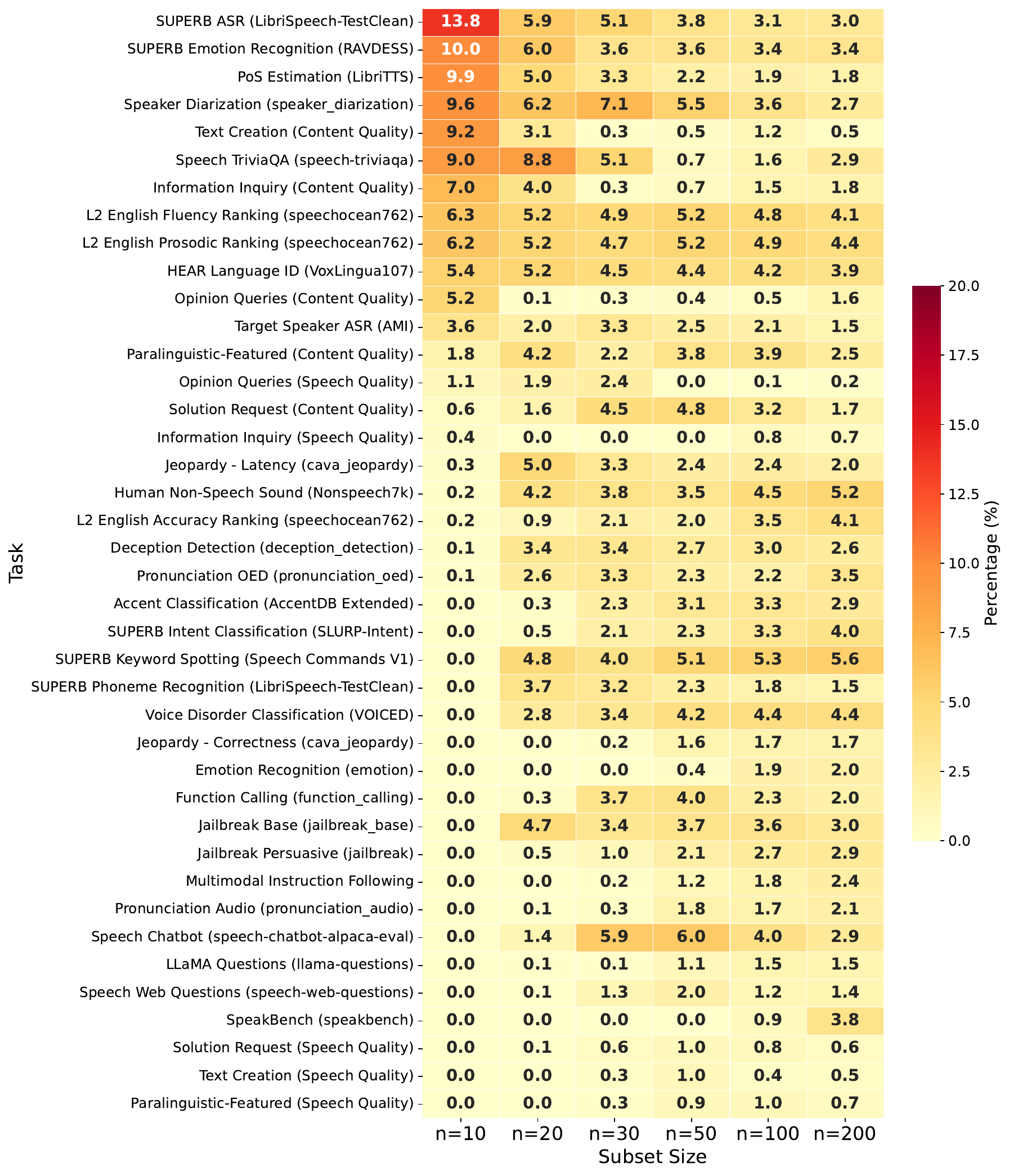}
\caption{\textbf{Task distribution for Combined Embedding method.} 
Heatmap shows the percentage of items from each task in subsets of 
varying sizes (n=10 to n=200), averaged across 100 random seeds. 
Darker red indicates higher representation. Tasks are ordered by 
their representation at n=10.}
\label{fig:combine_task_heatmap}
\end{figure*}

To understand what our selection methods prioritize, we analyze the 
task composition of selected subsets for both Anchor Points (used for 
n$\le$30) and Combined Embedding (used for n$\ge$50). Figures~\ref{fig:anchor_point_task_heatmap} 
and~\ref{fig:combine_task_heatmap} show the percentage distribution of 
tasks across subset sizes for both methods.

\subsection{Selection Patterns Across Subset Sizes}

Both methods exhibit a clear hierarchy in how they construct evaluation coverage, progressing from foundational capabilities to refined, multifaceted assessment:

\paragraph{Small Subsets ($n \leq 30$): Foundational Capabilities}
At minimal subset sizes, both methods heavily prioritize fundamental speech understanding capabilities that form the basis of model performance. In Anchor Points at $n=10$, basic tasks dominate: Target Speaker ASR (21.6\%), Speaker Diarization (18.3\%), Text Creation (9.3\%), and PoS Estimation (7.9\%). Combined Embedding shows similar patterns: SUPERB ASR (13.8\%), SUPERB Emotion Recognition (10.0\%), PoS Estimation (9.9\%), Speaker Diarization (9.6\%), Text Creation (9.2\%), and Speech TriviaQA (9.0\%). 

These tasks represent the core building blocks of audio model capabilities---the ability to accurately recognize speech, identify speakers, understand basic linguistic structure, and generate coherent content. A model's performance on these foundational dimensions establishes its baseline competence and determines whether it possesses the prerequisite skills for more sophisticated audio understanding. When evaluation budget is minimal, capturing these fundamental capabilities provides the most essential characterization of what a model can and cannot do.

\paragraph{Large Subsets ($n \geq 100$): Refined and Multifaceted Capabilities}
As subset size increases, selection shifts toward refined evaluation of paralinguistic and specialized capabilities that reflect more diverse and nuanced aspects of audio understanding. Tasks virtually absent at $n=10$ gain substantial representation by $n=100$-$200$: Human Non-Speech Sound increases from 0.0\% to 8.9--9.1\% (Anchor Points) and 0.2\% to 5.2\% (Combined), SpeakBench emerges from 0.0\% to 5.2\% (Combined), L2 English Accuracy/Fluency Ranking grow from 0.0\% to 9.3--10.0\% (Anchor Points), and Pronunciation tasks increase modestly.

This shift reflects an expansion in the dimensions along which model capabilities are evaluated. Paralinguistic tasks---understanding prosody, accent, fluency, and non-speech audio---capture sophisticated aspects of audio perception that go beyond literal content understanding. These refined capabilities constitute a model's full competence profile: beyond basic speech recognition and content generation, can it perceive subtle acoustic cues, handle diverse speaker characteristics, and understand audio in its full contextual richness? When evaluation budget allows, incorporating these dimensions provides a more complete picture of model capabilities, revealing strengths and weaknesses across the full spectrum of audio understanding rather than just foundational skills.

\subsection{Comparison of Selection Methods}

\paragraph{Concentrated vs. Distributed Capability Coverage}
Combined Embedding (Figure~\ref{fig:combine_task_heatmap}) achieves more uniform task distribution than Anchor Points (Figure~\ref{fig:anchor_point_task_heatmap}), particularly at small-to-medium sizes. At $n=10$, Anchor Points concentrates on just $\sim7$ tasks, while Combined Embedding distributes across 10+ tasks with $>5\%$ representation. At $n=200$, Combined Embedding shows relatively balanced 2--6\% across most tasks, while Anchor Points maintains sharper peaks (Human Non Speech Sound 9.1\%, L2 English Accuracy Ranking 9.5\%).
This partly explains their performance differences (Figure~\ref{fig:correlation_curves})---at $n \leq 30$, concentrated coverage of foundational capabilities suffices to characterize model quality for Anchor Points, while at $n \geq 50$, broader capability coverage better approximates the multifaceted nature of comprehensive evaluation for Combined Embedding.
\clearpage

\section{Conversational Agent Framework Implementation Details}
\label{appendix:agent_framework}

\subsection{System Architecture}

Our conversational agent framework is built on LiveKit Agents 1.0~\citep{livekit2024}, an open-source WebRTC infrastructure and agent framework that enables real-time audio streaming with low latency and provides high-level abstractions for pipeline voice assistant development. We extend this framework to support three distinct model architectures:

\begin{itemize}
    \item \textbf{End-to-end omni-modal}: Audio-in, audio-out models (e.g., GPT-4o-audio, Qwen2.5-Omni) that natively process and generate speech
    \item \textbf{Speech-to-text}: Audio-in, text-out models that understand audio but generate only text, requiring external TTS
    \item \textbf{Pipeline systems}: STT + text-only model + TTS, augmenting text models with separate speech components
\end{itemize}
\subsection{Audio Processing and Turn Detection}

We employ Silero VAD~\citep{silero2024} for voice activity detection with the following configuration optimized for conversational interactions:

\begin{itemize}
    \item \texttt{min\_speech\_duration}: 0.1s (captures short utterances)
    \item \texttt{min\_silence\_duration}: 2.0s (allows natural pauses)
    \item \texttt{prefix\_padding\_duration}: 0.2s (pre-speaking buffer)
    \item \texttt{activation\_threshold}: 0.4 (balanced sensitivity)
    \item \texttt{sample\_rate}: 16000 Hz
\end{itemize}

All audio is processed at 16kHz sampling rate. The 200ms pre-speaking buffer ensures we capture the beginning of utterances by maintaining a rolling window of audio frames before speech detection triggers. When the user begins speaking, these buffered frames are prepended to the captured audio, preventing cutoff of initial phonemes.

\subsection{System Prompts and Conversation Management}

All models receive consistent system prompts constructed from scenario-specific instructions plus standardized guidelines:

\begin{quote}
\textit{\{scenario\_prompt\}}

You are a voice assistant. \textit{[Architecture-specific instructions]} Respond naturally and conversationally.

\textbf{You should never reveal to the user which model you are.} If asked, say you are a voice assistant.
\end{quote}

Architecture-specific instructions vary by model type:
\begin{itemize}
    \item Audio-in/audio-out: "You receive audio input and respond with audio. Speak naturally in English."
    \item Audio-in/text-out: "You receive audio input and respond with text that will be converted to speech."
    \item Text-in/text-out: "You receive text transcribed from audio and respond with text that will be converted to speech."
\end{itemize}

Conversation context for the agent tracks:
\begin{itemize}
    \item Message roles (user/assistant/system/tool)
    \item User input audio
    \item Assistant text output or audio transcription
    \item Tool calls made by the assistant and their results
\end{itemize}

\subsection{Function Calling and Tool Integration}

All models have access to a consistent set of tools regardless of architecture:
\begin{itemize}
    \item \textbf{Web search}: DuckDuckGo API~\citep{duckduckgo} integration for retrieving information URLs
    \item \textbf{URL fetching}: Content extraction from web pages
    \item \textbf{Scenario-specific APIs}: Domain-specific tools (travel, shopping, calendar, social media, smart home, filesystem, messaging, job search) dynamically loaded based on evaluation scenario
\end{itemize}

Tool execution is tracked by a verifier component that monitors function calls and validates scenario goal completion, providing real-time feedback to participants on model progress.

\subsection{Real-time State Broadcasting}

To enable frontend visualization and interaction monitoring, the system broadcasts state updates via LiveKit's data channel (see Figure~\ref{fig:conversation_interface} in the conversation interface):

\begin{itemize}
    \item \textbf{Agent state changes}: "listening", "thinking", "speaking"
    \item \textbf{User state changes}: "listening", "speaking", "away"
    \item \textbf{Function call execution}: Function name, arguments, results, success/failure
    \item \textbf{Verification status}: Scenario goal completion, function call correctness
\end{itemize}

\section{Human Evaluation Protocol and Scenario Design}
\label{appendix:human_subjects}

\subsection{IRB Approval and Ethical Oversight}
This research was approved by the Institutional Review Board (IRB) at the authors' institution prior to data collection. All procedures followed institutional guidelines for research involving human subjects.

\subsection{Participant Recruitment}
We recruited native English speakers from the United States via Prolific, a crowdsourcing platform for research participants. Recruitment was limited to participants who were native English speakers, located in the United States, and 18 years of age or older. We balanced recruitment to achieve approximately equal gender representation (50\% male, 50\% female). A total of 776 participants completed the study.

\subsection{Compensation}
Participants were compensated \$0.25 base payment and \$0.25 per minute of conversation (\$2.50 for a full 10-minute conversation), with additional bonuses: \$1 for successfully completing goal-oriented tasks and \$0.25 for providing feedback. This resulted in compensation ranging from \$2.75 to \$4 for 10-13 minutes of total participation time (including consent, conversation, and rating submission), yielding an hourly rate of at least \$15/hour minimum wage rate.

\subsection{Informed Consent and Data Usage}
\begin{figure*}[t]
    \centering
    \includegraphics[width=0.95\textwidth]{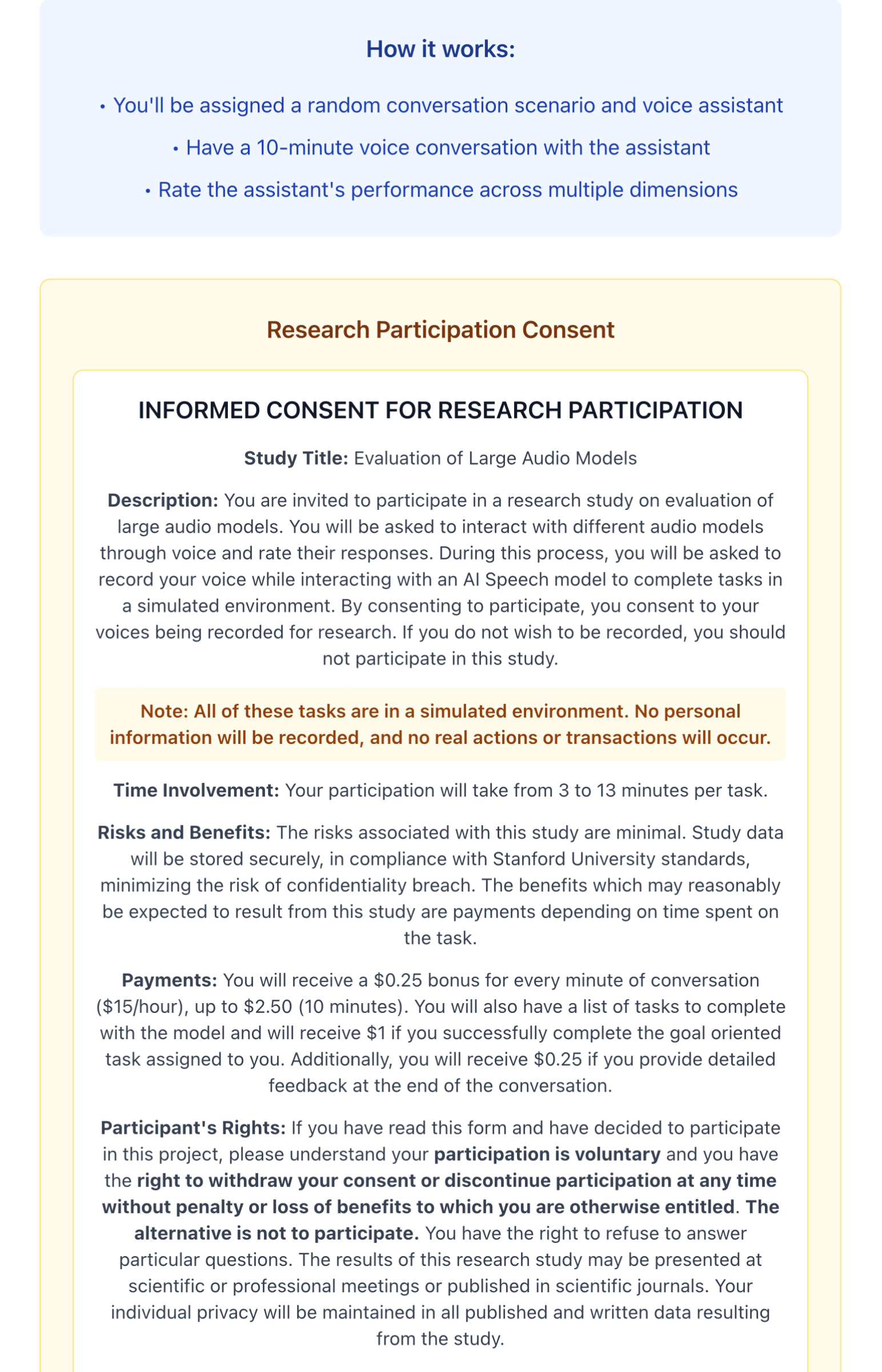}
    \caption{\textbf{Study overview and informed consent form.} The page begins with "How it works" explaining the study workflow, followed by the detailed informed consent section covering purpose, procedures, risks, benefits, compensation, and participant rights.}
    \label{fig:consent_form}
\end{figure*}

Before beginning the study, all participants were presented with a consent form (Figure~\ref{fig:consent_form}) that first explained the study workflow: assignment to a random conversation scenario and voice assistant, a 10-minute voice conversation, and rating the assistant's performance across multiple dimensions. Following this overview, participants reviewed detailed information about the research purpose, study procedures, voice recording, data usage for research purposes, privacy protections, and their right to withdraw at any time.

The consent form specifically informed participants that:
\begin{itemize}[noitemsep]
    \item Their voices would be recorded during interaction with AI speech models
    \item All tasks occur in a simulated environment with no real actions or transactions
    \item No personal information would be recorded
    \item Study data would be stored securely in compliance with institutional standards
    \item Risks associated with participation are minimal
    \item Participation is voluntary and they have the right to withdraw consent at any time without penalty
    \item Results may be presented at scientific or professional meetings or published in scientific journals, with individual privacy maintained
\end{itemize}

\subsection{Instructions Given to Participants}

\begin{figure*}[t]
    \centering
    \includegraphics[width=\textwidth]{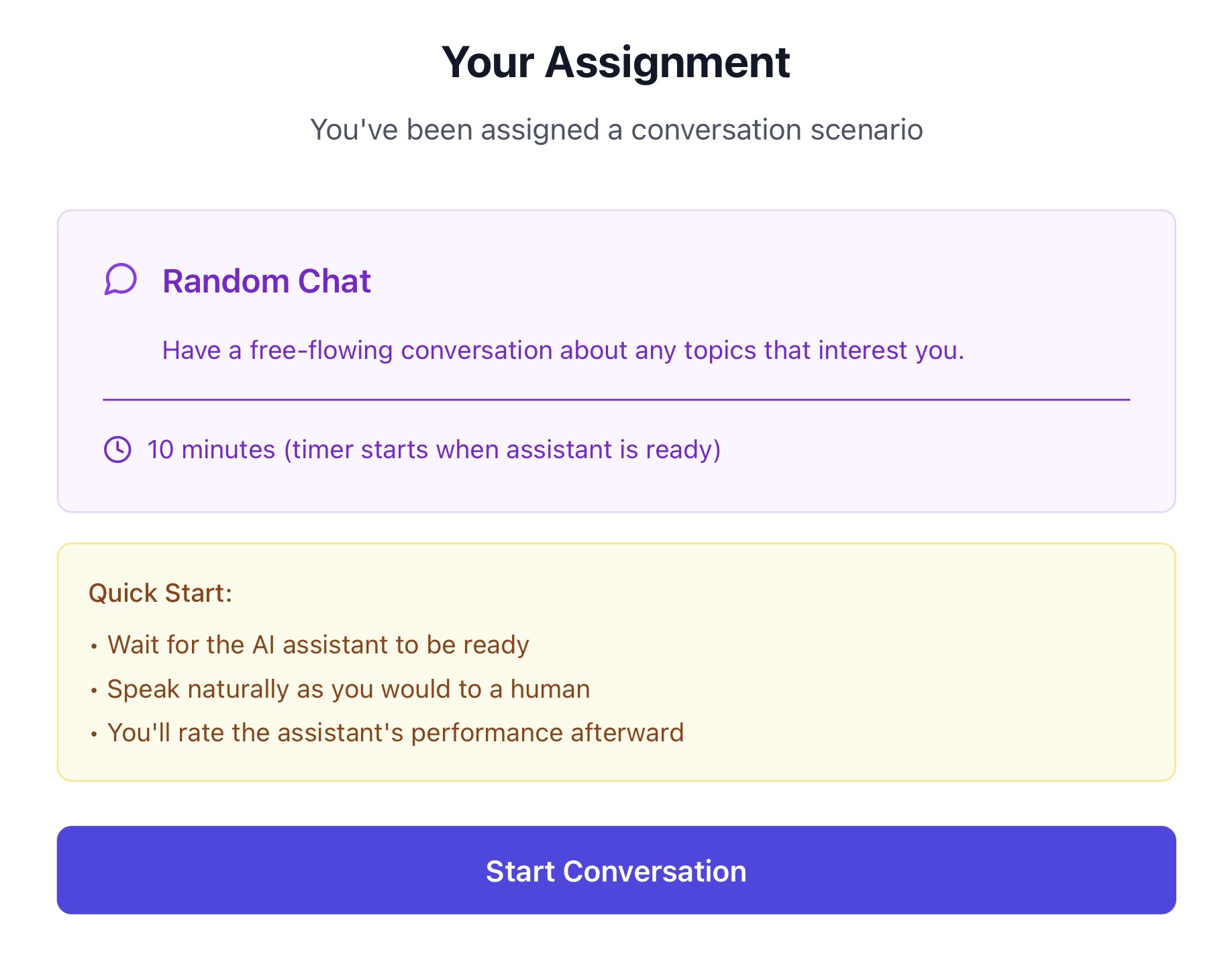}
    \caption{\textbf{Open chat scenario assignment.} Example of scenario instructions for free-form conversations.}
    \label{fig:open_chat_scenario}
\end{figure*}

Upon consenting to participate, Participants then received scenario-specific instructions based on their randomly assigned conversation type:

\paragraph{Open Chat (20\% of conversations):} Participants were shown a simple instruction screen indicating they would engage in free-form conversation with the AI assistant on any topic of interest for 10 minutes, with no specific goals or constraints (Figure~\ref{fig:open_chat_scenario}).

\begin{figure*}[t]
    \centering
    \includegraphics[width=\textwidth]{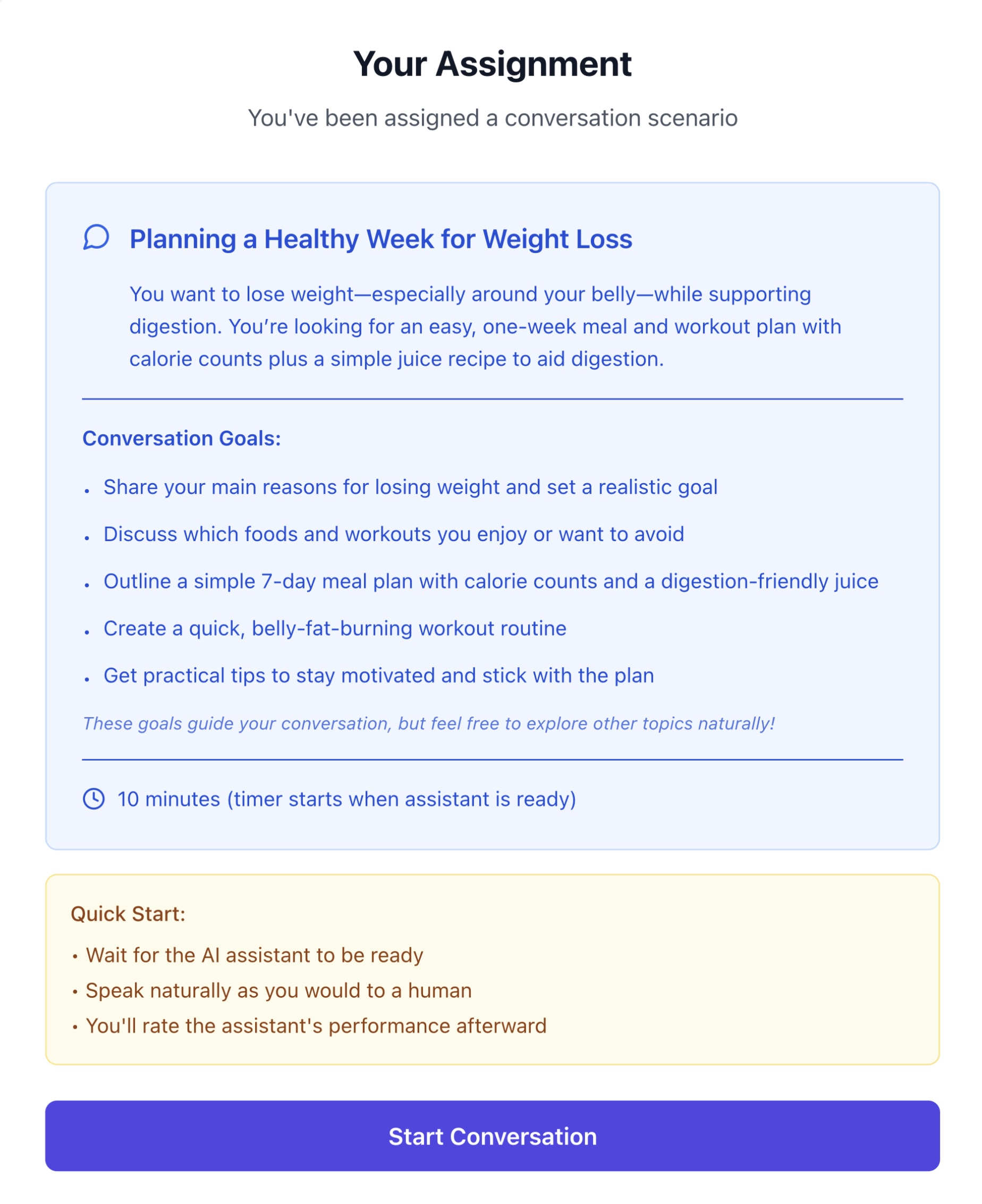}
    \caption{\textbf{Goal-oriented scenario assignment.} Example showing the "Weight Loss" scenario with structured context and goals.}
    \label{fig:goal_oriented_scenario}
\end{figure*}

\paragraph{Goal-Oriented Dialogue (40\% of conversations):} Participants received a scenario card (Figure~\ref{fig:goal_oriented_scenario}) containing: a brief scenario title, your goal, situation description, numbered task steps, and clarifying notes.

\begin{figure*}[t]
    \centering
    \includegraphics[width=\textwidth]{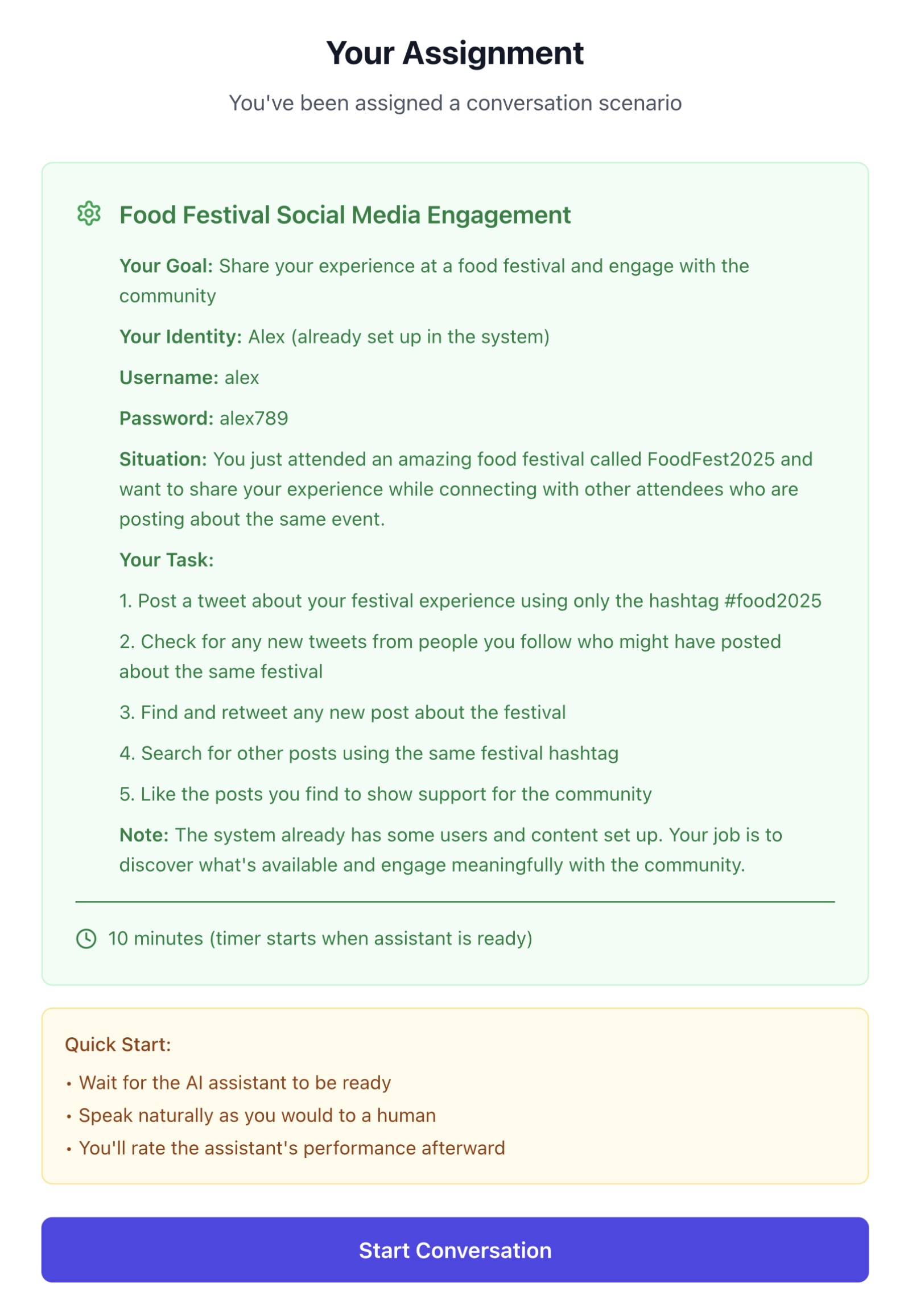}
    \caption{\textbf{Function calling scenario assignment.} Example showing the "Social Media Engagement" scenario with detailed task requirements.}
    \label{fig:function_calling_scenario}
\end{figure*}

\paragraph{Function Calling Tasks (40\% of conversations):} Participants received structured task scenarios (Figure~\ref{fig:function_calling_scenario}) including scenario title, goal, identity if needed, situation, key details, and step-by-step tasks.

\subsection{Conversation Interface}

\begin{figure*}[t]
    \centering
    \includegraphics[width=\textwidth]{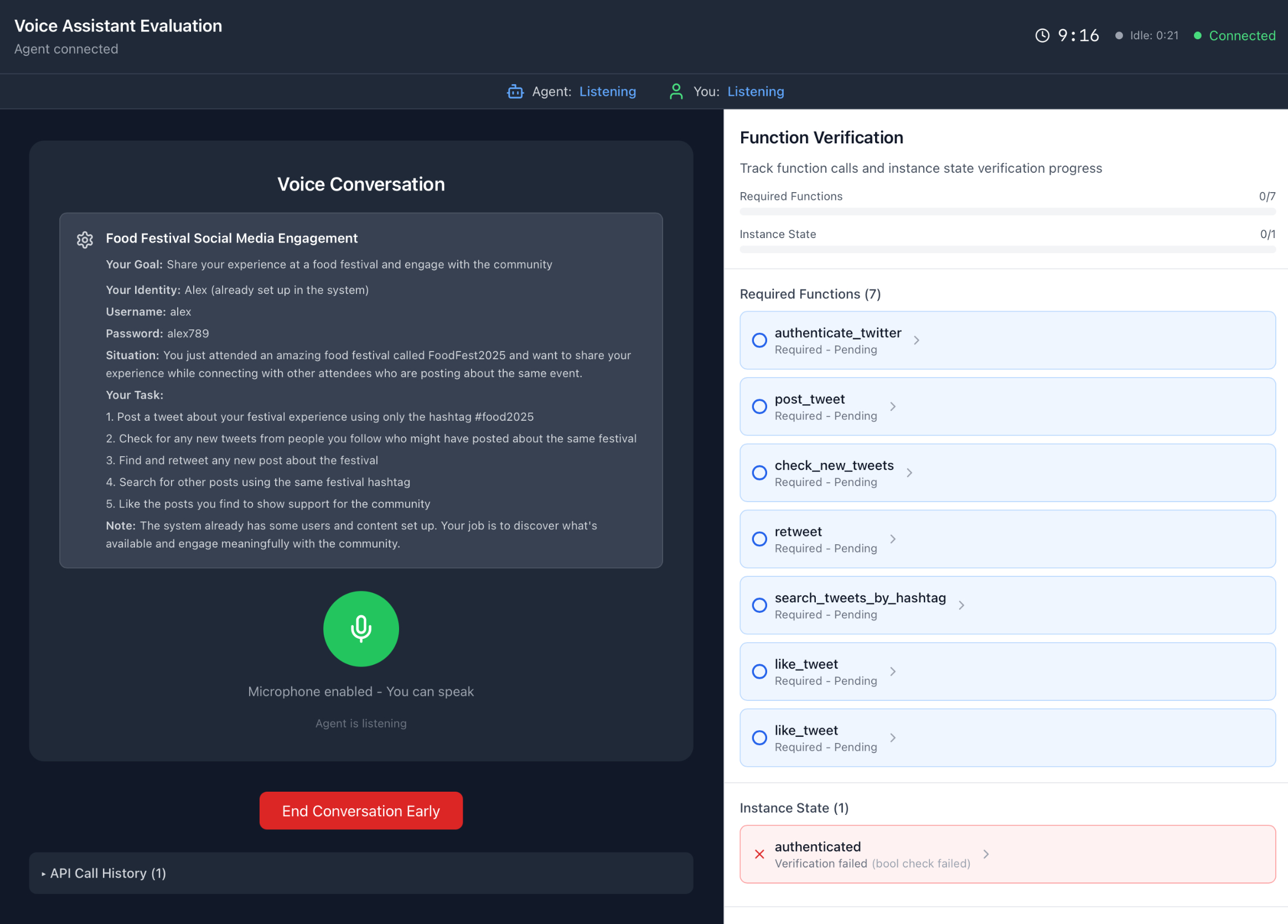}
    \caption{\textbf{Real-time conversation interface.} Shows the voice interaction screen with scenario information (left), conversation status indicators, and function verification panel (right) for tracking task completion in function calling scenarios.}
    \label{fig:conversation_interface}
\end{figure*}

During the conversation, participants interacted through a real-time voice interface (Figure~\ref{fig:conversation_interface}). 

The interface displayed:
\begin{itemize}[noitemsep]
    \item The assigned scenario information in the left panel
    \item Real-time conversation status indicators showing the status of the agent and user (e.g. listening or speaking)
    \item A microphone button to control voice input
    \item Connection status and elapsed time
    \item For function calling tasks: a right panel showing "Function Verification" with real-time tracking of required function calls and instance state verification progress
    \item An "End Conversation Early" button for participants who wished to terminate before the 10-minute timer
\end{itemize}

\subsection{Post-Conversation Evaluation}

\begin{figure*}[t]
    \centering
    \includegraphics[width=\textwidth]{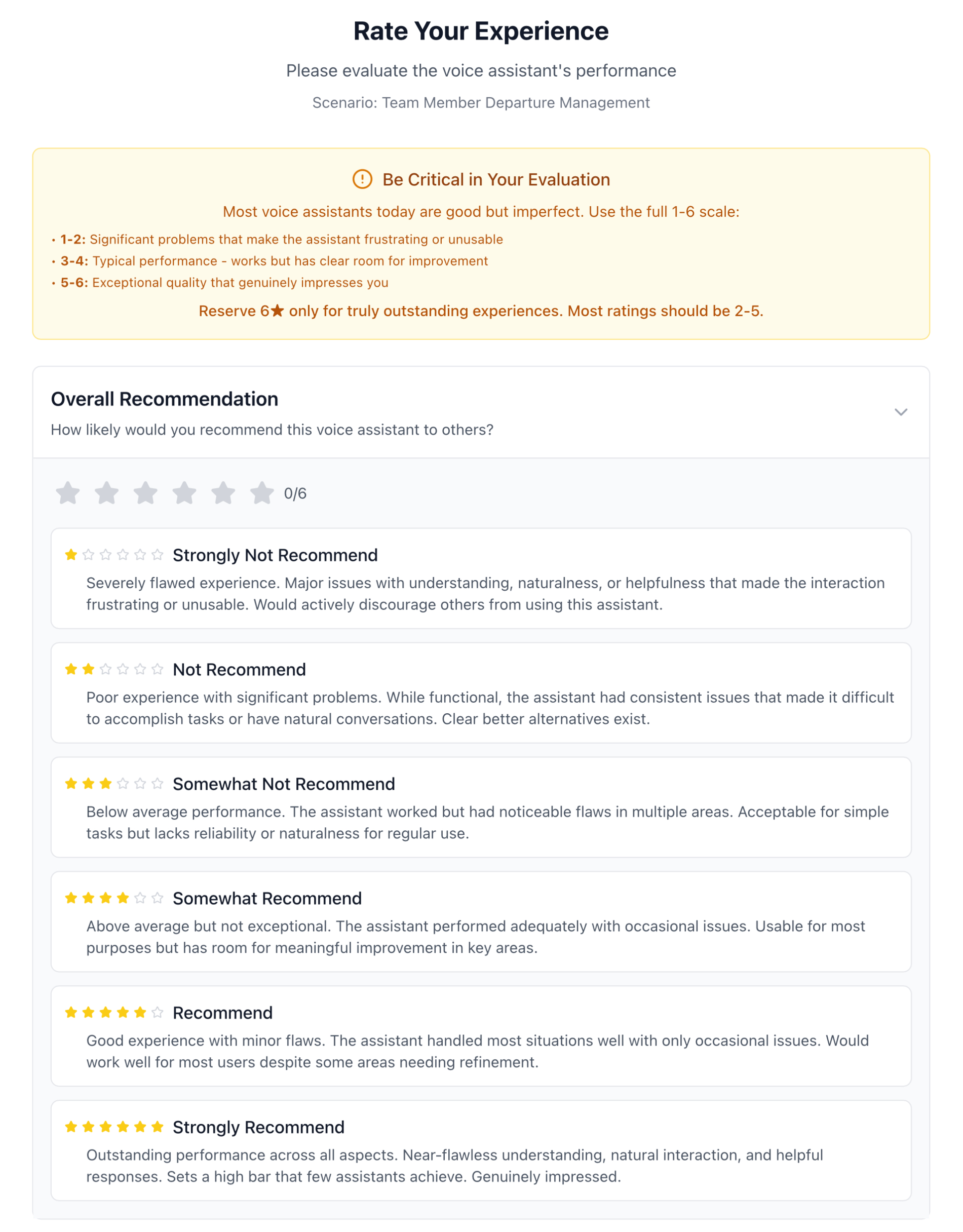}
    \caption{\textbf{Overall recommendation rating interface.} Expandable 6-point scale with detailed descriptions for each rating level.}
    \label{fig:evaluation_overall}
\end{figure*}

\begin{figure*}[t]
    \centering
    \includegraphics[width=\textwidth]{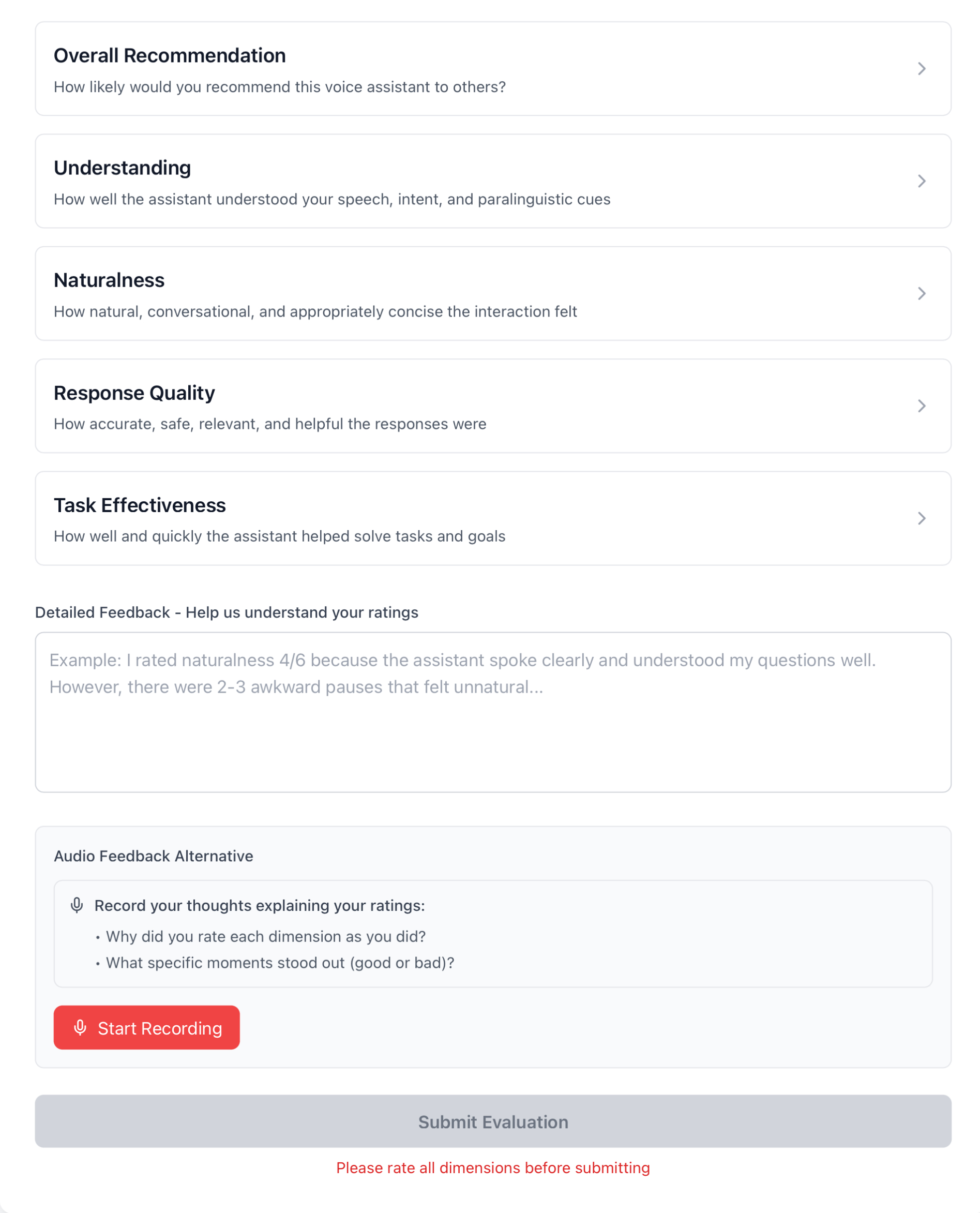}
    \caption{\textbf{Multi-dimensional rating interface.} All five evaluation dimensions with expandable scales, text feedback area, and optional audio feedback recording.}
    \label{fig:evaluation_dimensions}
\end{figure*}

After completing the conversation, participants provided ratings on an evaluation page (Figures~\ref{fig:evaluation_overall} and~\ref{fig:evaluation_dimensions}). Participants were instructed to use the full 1-6 scale, with guidance that 1-2 indicates significant problems, 3-4 represents typical performance with room for improvement, and 5-6 is reserved for exceptional quality.

Participants rated five dimensions using 6-point Likert scales: (1) Overall Recommendation, (2) Understanding (speech, intent, and paralinguistic cues), (3) Naturalness (conversational flow and conciseness), (4) Response Quality (accuracy, safety, relevance, helpfulness), and (5) Task Effectiveness (efficiency in achieving goals).

Participants then provided required written feedback explaining their ratings and could optionally record audio feedback.

\subsection{Scenario Design and Generation}
\label{appendix:scenario_design}
\subsubsection{Goal-Oriented Scenario Generation}
\label{appendix:scenario_generation}
We created realistic conversation scenarios by adapting real user-chatbot interactions from two complementary text-based datasets: LMSYS-Chat-1M~\citep{zheng2023lmsys}, containing diverse Chatbot Arena conversations, and WildChat~\citep{zhao2024wildchat}, documenting authentic ChatGPT usage patterns.

Using GPT-4.1 with few-shot prompting and reject sampling, we transformed appropriate conversations into structured scenarios containing: (1) a brief title, (2) 2-3 sentence description providing user context, and (3) 3-5 conversation goals representing logical discussion milestones. Goals serve as conversation helpers to encourage sustained engagement rather than strict requirements. We rejected conversations requiring specialized knowledge, visual aids, external tools, or containing sensitive content. This process yielded 500 candidate scenarios from each dataset (1,000 total).

A second filtering stage using o4-mini validated and improved scenarios for conversational suitability, ensuring appropriateness for general participants in voice-only interactions.

\subsubsection{Function Calling Task Scenarios}
\label{appendix:function_calling_scenarios}

We adapted the Berkeley Function-Calling Leaderboard (BFCL) v3~\citep{patil2025bfcl} multi-turn function calling framework to create 40 realistic voice assistant evaluation scenarios spanning 8 domains: calendar management (5 tasks), shopping (5 tasks), travel booking (8 tasks), social media (5 tasks), smart home control (1 task), filesystem operations (7 tasks), messaging (5 tasks), and job search (4 tasks). Each scenario includes: (1) a user goal and situational context, (2) initial system state with pre-populated data, (3) available function definitions, (4) required verifiable actions for successful completion, (5) forbidden operations that constitute violations, and (6) expected final state for verification. 

\clearpage
\section{Qualitative Analysis of Human Evaluations}
\label{appendix:qualitative}
\subsection{Failure Modes and User Feedback}
To understand model limitations from users' perspective, we analyzed open-ended feedback from all 776 conversations. For participants who provided optional audio feedback, we first transcribed recordings to text using Whisper-large-v3~\cite{radford2023robust} and concatenated with written feedback. To protect participant privacy, all feedback was processed through Microsoft Presidio~\cite{presidio} to automatically detect and mask personally identifiable information (PII) including names, addresses, phone numbers, email addresses, and other identifying details. Detected entities were replaced with placeholders before any subsequent analysis or storage.

We then applied a three-stage automated pipeline powered by GPT-5.2~\cite{openai2025gpt52} to the privacy-protected feedback. Among the 741 participants who left feedback, 621 expressed some form of dissatisfaction or suggested improvements.

\textbf{Stage 1: Dissatisfaction Detection and Summarization.} We fed each piece of raw feedback to GPT-5.2 to determine whether it expressed dissatisfaction and, if so, to generate a concise 1--2 sentence summary of the specific failure mode.

\textbf{Stage 2: Category Generation.} We provided all 621 dissatisfaction summaries from Stage 1 to GPT-5.2 in a single prompt, instructing it to inductively generate comprehensive failure mode categories that cover the range of problems mentioned. The model generated 25 distinct categories.

\textbf{Stage 3: Feedback Categorization.} Using the 25 categories generated in Stage 2, we classified each dissatisfaction summary by asking GPT-5.2 to assign it to one or more applicable categories, allowing for multi-label classification. We then coded the 621 feedback responses into the resulting categories.

\subsubsection{Overall Failure Mode Distribution}
\label{sec:overall_failure}
Table~\ref{tab:failure_modes} presents the complete distribution of failure modes. The results reveal clear patterns in user dissatisfaction that are not adequately captured by existing benchmarks.

\begin{table}[h]
\centering
\resizebox{\columnwidth}{!}{%
\small
\begin{tabular}{lc}
\toprule
\textbf{Failure Mode} & \textbf{Count (\%)} \\
\midrule
Robotic/Unnatural Speaking Style & 266 (42.8\%) \\
Task Execution Failure & 147 (23.7\%) \\
Slow Response Latency & 143 (23.0\%) \\
Stilted Conversation Flow & 117 (18.8\%) \\
Overly Verbose Responses & 107 (17.2\%) \\
Unhelpful Response Strategy & 95 (15.3\%) \\
Misunderstood User Intent & 82 (13.2\%) \\
Looping/Stalling/Non-Responsiveness & 70 (11.3\%) \\
Audio Output Glitches & 68 (11.0\%) \\
Insufficient Proactivity & 57 (9.2\%) \\
Poor Speech Recognition & 54 (8.7\%) \\
Poor UX/Interface Support & 44 (7.1\%) \\
Missing Confirmation/Progress Transparency & 43 (6.9\%) \\
Inconsistent Voice or Language Output & 42 (6.8\%) \\
Incomplete Answers & 39 (6.3\%) \\
Tool/API/Integration Errors & 37 (6.0\%) \\
Incorrect or Unreliable Information & 29 (4.7\%) \\
Outdated or Not-Current Information & 19 (3.1\%) \\
Tone/Interpersonal Issues & 16 (2.6\%) \\
Poor Instruction Following & 15 (2.4\%) \\
Over-Agreeable/Lack of Critical Pushback & 14 (2.3\%) \\
Context/Memory/State Tracking Issues & 12 (1.9\%) \\
Privacy/Security/Auth Friction & 11 (1.8\%) \\
Bias/Defensiveness/Trust Issues & 5 (0.8\%) \\
Unintended or Premature Actions & 4 (0.6\%) \\
\bottomrule
\end{tabular}
}
\caption{\textbf{Distribution of failure modes from human feedback.} Count and percentage of 621 dissatisfaction cases across 25 categories. Percentages sum to >100\% due to multi-label classification where feedback could be assigned to multiple categories.}
\label{tab:failure_modes}
\end{table}

\paragraph{Conversational Quality Dominates User Dissatisfaction}
The most striking finding is that conversational quality issues far outweigh technical capability failures. Three related categories dominate user complaints:

\textbf{Robotic/Unnatural Speaking Style (42.8\%)} emerged as the single most common issue, with users describing outputs as monotone, overly formal, or having unnatural prosody. Representative feedback includes:
\begin{itemize}[noitemsep]
    \item \textit{``The assistant understood the queries and provided sufficient, reasonable answers, but its speech sounded slightly robotic and overly formal—at times like it was reading from a book—reducing naturalness.''}
    \item \textit{``I rated naturalness 4/6 because it felt a little monotone and robotic with little inflection or changes of pace that are typical in normal human conversation.''}
    \item \textit{``I would have liked maybe a little side chatter.''}
\end{itemize}

\textbf{Stilted Conversation Flow (18.8\%)} reflects issues with dialogue structure, including excessive numbered lists, scripted delivery, and poor turn-taking:
\begin{itemize}[noitemsep]
    \item \textit{``It was more focused on lists, so I did not get a chance to really have a normal conversation.''}
    \item \textit{``It felt less like a flowing conversation and more like a series of individual responses.''}
    \item \textit{``Overall I recommended her as five the only reason that I gave her a little bit lower score of four on the task effectiveness is because she was very cautious and kept repeating herself over and over again.''}
\end{itemize}

\textbf{Overly Verbose Responses (17.2\%)} captures users' frustration with excessive detail when concise answers were expected:
\begin{itemize}[noitemsep]
    \item \textit{``I think the lengthy response can let people lose focus and forget the task at hand and become distracted by other parts that the AI is talking about.''}
    \item \textit{`Even for simple questions, replies contained too much detail, causing the user to lose attention/tune out despite the information generally being useful.''}
    \item \textit{`I do feel like some of the information could be overwhelming with how much she gives.''}
\end{itemize}

Collectively, these three conversational quality categories appear in 490 mentions. Since feedback can match multiple categories, we find that 352 unique feedback instances (56.7\%) mention at least one of these three conversational quality issues, making them the dominant source of user dissatisfaction. While some benchmarks include speech quality metrics (e.g., UTMOS scores in WildSpeech-Bench for speech quality), these measures capture only isolated utterance quality rather than conversational dynamics like verbosity, formality appropriateness, or turn-taking. This gap between static quality assessment and interactive conversational experience helps explain why our full benchmark achieves 0.851 correlation with human overall satisfaction---benchmarks primarily optimize for correctness and isolated speech quality, while users evaluate holistic conversational experience including response length, style appropriateness, and dialogue flow.

\paragraph{Task Execution and Technical Reliability}
Beyond conversational quality, task execution failure (23.7\%) and technical issues significantly impact user experience. Task execution complaints typically involved incomplete function calls or failed tool integrations:
\begin{itemize}[noitemsep]
    \item \textit{``It kept saying something about there was errors on the calendar that I had to manually fix. It wouldn't try to solve them. And overall, it wouldn't book the meeting at the end.''}
    \item \textit{``I felt that the AI assistant was really helpful in what I was trying to do. Unfortunately, she wasn't able to help me find a cheap flight on the day that I wanted to take my trip. She asked if I wanted to change the date or add more money to my card, which I chose not to do.''}
\end{itemize}

\paragraph{Latency and Infrastructure Limitations}
Slow response latency (23.0\%) and audio output glitches (11.0\%) represent infrastructure limitations rather than core model capabilities. Notably, feedback in these categories typically expressed mild frustration rather than severe dissatisfaction:
\begin{itemize}[noitemsep]
    \item \textit{`There were a few moments that the assistant took longer than expected to reply to my responses. Other than that, the experience was quite pleasant.''}
    \item \textit{``There was some stuttering at times, but other than that, it was very good.''}
\end{itemize}
These issues highlight the need for real-time processing capabilities, such as streaming audio input processing that begins before users finish speaking. While proprietary APIs like GPT-Realtime API~\cite{openai2025gpt-realtime} and Gemini Live API~\cite{google2024gemini_live} offer such capabilities, they employ model-specific optimizations that conflate infrastructure improvements with core model capabilities, which would not reflect what our benchmarking is trying to optimize for. We did not build a more real-time system that processes audio while speaking either, as this requires engineering changes to the model's internal structure and would not support closed-source models in our evaluation framework. Our standardized processing approach prioritizes isolating model performance from deployment optimizations, though this comes at the cost of increased latency. Importantly, we include latency measurements in our 40 benchmark tasks, so latency impact can be quantified in our regression models.

\paragraph{Speech Recognition Less Problematic Than Expected}
Notably, poor speech recognition accounts for only 8.7\% of complaints. This suggests contemporary LAMs have largely solved basic ASR for native English speakers in controlled conditions. The remaining challenges primarily involve edge cases (accents, background noise, domain-specific terminology) or downstream intent interpretation (13.2\%) rather than raw transcription accuracy. This finding aligns with our observation in Table~\ref{tab:human_ratings} that Understanding scores consistently exceed Overall Satisfaction across all models---speech comprehension is generally not the limiting factor in user experience.

\paragraph{Implications for Benchmark Design}
Our qualitative analysis reveals three critical gaps in current LAM evaluation:

\begin{enumerate}[leftmargin=*,noitemsep]
    \item \textbf{Missing conversational quality metrics:} Naturalness, conciseness, and appropriate formality drive 78.9\% of user dissatisfaction yet are not systematically evaluated in existing benchmarks.
    
    \item \textbf{Static evaluation misses interactive failures:} Latency, turn-taking, error recovery, and real-time audio quality only manifest in live conversation, not in offline benchmark tasks.
    
    \item \textbf{Accuracy-usability tradeoff unaddressed:} Benchmarks prioritize correctness (ASR word error rate, task completion) while users weight naturalness and efficiency equally or higher in determining overall satisfaction.
\end{enumerate}

These findings justify our human preference validation approach: benchmark subset selection must be validated against user experience to ensure selected items capture not just the capabilities measured by benchmarks, but the dimensions users actually care about. Our regression models (Section~\ref{sec:preference_prediction}) partially address this gap by learning to weight benchmark items according to their correlation with human satisfaction, effectively discovering which benchmark tasks serve as proxies for conversational quality that benchmarks do not directly measure.

\subsubsection{Model-Specific Failure Patterns}

\begin{figure*}[t]
\centering
\includegraphics[width=0.95\textwidth]{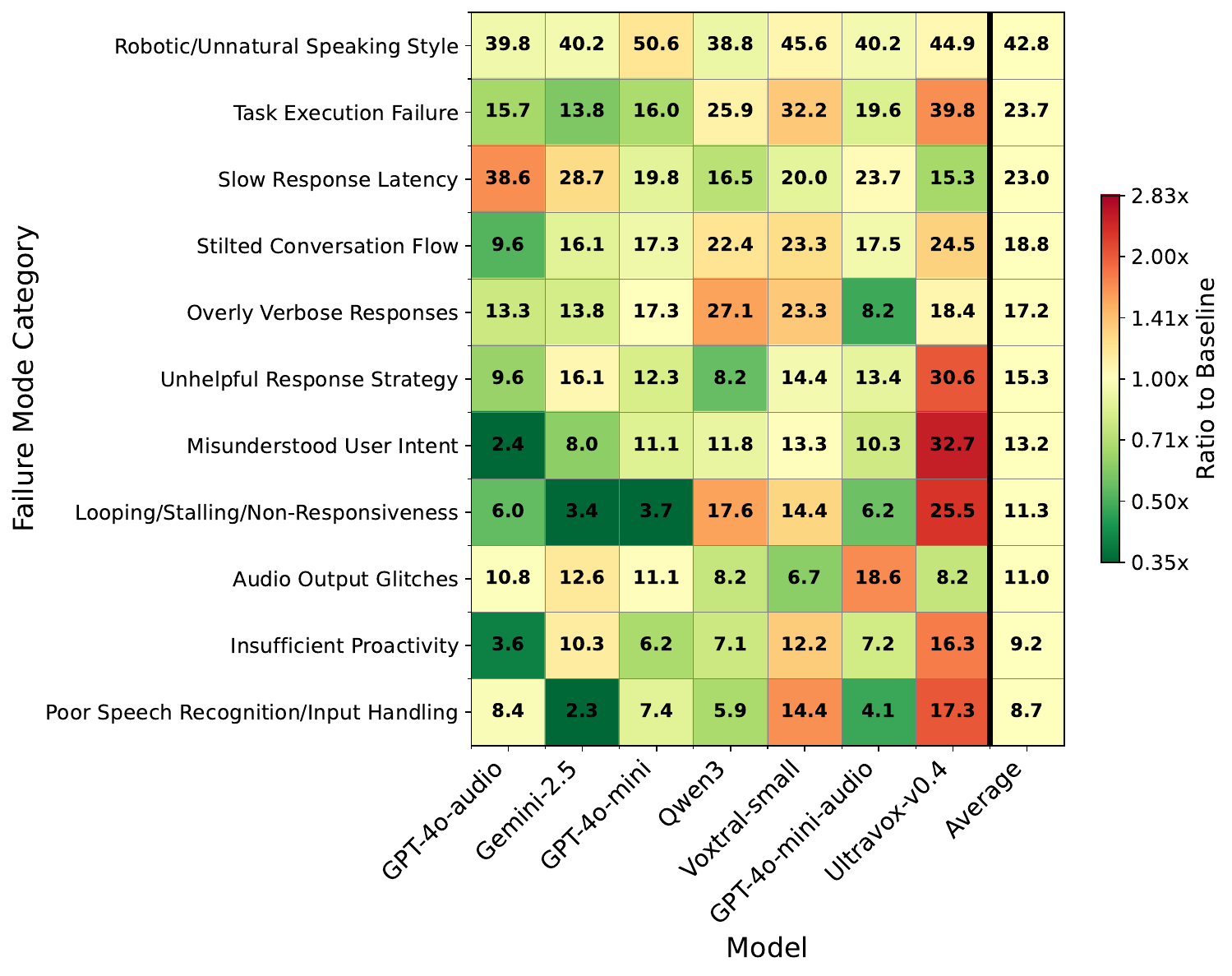}
\caption{\textbf{Model-specific failure mode distributions.} Heatmap shows percentage of dissatisfaction cases mentioning each failure category for each model. Cell color intensity represents the ratio to baseline (average across all models), with darker red indicating higher-than-average rates and darker green indicating lower-than-average rates. Models are ordered by overall human satisfaction (left to right: highest to lowest). Only failure modes with >8\% overall prevalence are shown.}
\label{fig:failure_heatmap}
\end{figure*}

While Section~\ref{sec:overall_failure} identified overall trends, individual models exhibit distinct failure profiles reflecting architectural choices and optimization priorities. Figure~\ref{fig:failure_heatmap} visualizes failure mode distributions across all 7 evaluated models for the most prevalent categories (>8\% overall rate).

\paragraph{TTS Quality Degrades Naturalness in Pipeline Systems}

Models using external TTS (GPT-4o-mini+STT+TTS, Voxtral+TTS, Ultravox+TTS) show elevated robotic style complaints. GPT-4o-mini with full pipeline exhibits the highest rate at 50.6\% (1.18× baseline), while Voxtral (45.6\%, 1.07×) and Ultravox (44.9\%, 1.05×) also exceed the 42.8\% average. In contrast, GPT-4o-audio (39.8\%, 0.93×) and GPT-4o-mini-audio (40.2\%, 0.94×)—which use native generation or higher-quality TTS—perform closer to baseline. This pattern confirms that pipeline architectures incorporating separate TTS components degrade conversational naturalness, with the full STT→LLM→TTS pipeline showing the most severe impact.

\paragraph{Closed-Source Large Models Trade Latency for Quality}

Proprietary large models exhibit disproportionately high latency issues: GPT-4o-audio (38.6\%, 1.68× baseline), Gemini-2.5-Flash (28.7\%, 1.25×), and GPT-4o-mini-audio (23.7\%, 1.03×) all exceed the 23.0\% average, while open-source models like Ultravox (15.3\%, 0.67×) and Qwen3-Omni (16.5\%, 0.72×) show lower rates. This suggests closed-source providers prioritize response quality over speed, which introduce user-perceptible delays. Interestingly, users appear willing to tolerate this tradeoff—GPT-4o-audio achieves the highest satisfaction (4.98) despite the highest latency complaints, indicating quality outweighs responsiveness for overall experience.

\paragraph{Large Open-Source Models Struggle with Conciseness}

Qwen3-Omni (27.1\%, 1.58× baseline) and Voxtral (23.3\%, 1.36×) show the highest rates of overly verbose responses, substantially exceeding the 17.2\% average. These models also exhibit elevated stilted conversation flow issues (Qwen3: 22.4\%, 1.19×; Voxtral: 23.3\%, 1.24×), suggesting systematic problems with conversational brevity and natural dialogue structure. In contrast, GPT-4o-mini-audio achieves remarkably low verbosity (8.2\%, 0.48×), indicating successful optimization for concise interaction. This verbosity gap likely stems from instruction-tuning differences—open-source models may be trained to provide comprehensive explanations while proprietary voice assistants are optimized for minimal, conversational responses.

\paragraph{Ultravox Audio Understanding Limitations}

Ultravox-v0.4-ToolACE-8B still exhibits severe audio comprehension issues: poor speech recognition (17.3\%, 1.99× baseline) and misunderstood user intent (32.7\%, 2.48×) are both approximately double the average rates. These understanding failures compound with task execution problems (39.8\%, 1.68×) and unhelpful response strategies (30.6\%, 2.00×), creating a cascading failure pattern that explains the model's lowest overall satisfaction (3.34). Notably, Ultravox's benchmark scores (0.384) correctly predict poor performance, but the failure mode analysis reveals the specific bottleneck: audio input processing rather than output capabilities.

\paragraph{Native Audio Output Does Not Guarantee Quality}

GPT-4o-mini-audio shows 18.6\% audio output glitches (1.69× baseline) and 30.9\% inconsistent voice output (4.54× the 6.8\% baseline, not shown in figure)—dramatically higher than models using external TTS like GPT-4o-mini+TTS (8.2\% glitches, 6.2\% inconsistency). This counterintuitive finding suggests that \textbf{small-scale native audio generation does not necessarily provide advantages over high-quality TTS synthesis}. While end-to-end architectures theoretically enable better prosody control and voice consistency, gpt-4o-mini-audio appear to lack the capacity for stable audio generation, producing artifacts, dropouts, and voice switching issues. This explains why GPT-4o-mini-audio (3.69 satisfaction) substantially underperforms the pipeline-based GPT-4o-mini+STT+TTS (4.51), despite the latter's higher robotic style complaints—reliability trumps naturalness when audio output actively fails.

These findings demonstrate that failure modes are not uniformly distributed—architectural choices, model scale, and optimization priorities create distinct profiles that benchmarks alone cannot reveal. Understanding these patterns enables practitioners to select models aligned with deployment constraints: prioritize large end-to-end models for conversational applications, accept pipeline systems when leveraging existing text-based capabilities, and recognize that small-scale native audio generation currently offers limited advantages over high-quality TTS synthesis.
\begin{table*}[h]
\centering
\small
\begin{tabular}{lccccc}
\toprule
& \textbf{Overall} & \textbf{Understanding} & \textbf{Naturalness} & \textbf{Quality} & \textbf{Effectiveness} \\
\midrule
\textbf{Overall} & --- & 0.669 & 0.626 & 0.773 & 0.781 \\
\textbf{Understanding} & 0.949 & --- & 0.523 & 0.679 & 0.658 \\
\textbf{Naturalness} & 0.970 & 0.957 & --- & 0.586 & 0.528 \\
\textbf{Quality} & 0.978 & 0.990 & 0.969 & --- & 0.793 \\
\textbf{Effectiveness} & 0.957 & 0.992 & 0.946 & 0.986 & --- \\
\bottomrule
\end{tabular}
\caption{\textbf{Pairwise correlations between rating dimensions.} Upper triangle: sample-level correlations across 776 conversations. Lower triangle: model-level correlations across 7 models.}
\label{tab:dimension_correlations}
\end{table*}
\subsection{Rating Dimension Correlation Analysis}
\label{appendix:dimension_correlations}

To understand the relationships between different aspects of user satisfaction, we analyzed pairwise correlations between rating dimensions at both the conversation level (N=776 individual conversations) and model level (N=7 models). Table~\ref{tab:dimension_correlations} presents the complete correlation matrix. Model-level correlations are uniformly high (all r $>$ 0.94) due to the limited sample size (N=7 models) and the fact that aggregation removes individual conversation variance. Consequently, we focus our analysis on sample-level correlations, which provide more nuanced insights into how different aspects of model performance relate to user satisfaction across 776 individual interactions.

\paragraph{Response Quality and Task Effectiveness Drive Overall Satisfaction:}
At the sample level, Response Quality (r=0.773) and Task Effectiveness (r=0.781) exhibit the strongest correlations with Overall Satisfaction. These two dimensions are also highly interdependent (r=0.793), indicating they capture related aspects of model utility: when a model provides high-quality responses, it tends to complete tasks effectively, and vice versa. Together, these functional capabilities are the primary drivers of user satisfaction in voice assistant interactions and are therefore predictive for the overall rating.

\paragraph{Naturalness Shows Weak Interdependence and Limited Impact:}
In contrast, Naturalness demonstrates notably weaker correlations with other dimensions. Its correlation with Task Effectiveness (r=0.528) and Understanding (r=0.523) are the lowest pairwise correlations in the matrix, suggesting Naturalness represents a relatively independent aspect of conversational quality. Moreover, Naturalness exhibits the weakest correlation with Overall Satisfaction (r=0.626) among all four specific dimensions, indicating that conversational flow and naturalness, while measurable, contribute less to overall user satisfaction than functional capabilities. This pattern suggests that users prioritize task completion and response quality over conversational naturalness—a model can feel somewhat robotic yet still achieve high satisfaction if it delivers accurate, effective results.

\paragraph{Understanding Shows High Variance but Limited Discriminative Power:}
Understanding demonstrates reasonable sample-level correlation with Overall Satisfaction (r=0.669), but notably shows the lowest model-level correlation among all dimensions (r=0.949 vs. r=0.970 for Naturalness). This discrepancy likely stems from ceiling effects: as shown in Table~\ref{tab:human_ratings}, Understanding scores are consistently high across all models (range: 4.036-5.368), with even the lowest-performing model exceeding 4.0 on the 6-point scale. This restricted range at the model level reduces discriminative power, making Understanding less useful for distinguishing between models despite its reasonable within-model variance across individual conversations.

\subsection{Verifiable Task Completion and User Satisfaction}
For the 40\% of conversations involving function calling tasks, we analyzed objectively verifiable task progress in terms of steps alongside subjective user satisfaction metrics. Table~\ref{tab:function_calling} presents the results across all evaluated models.
\begin{table*}[h]
\centering
\small
\setlength{\tabcolsep}{8pt}
\begin{tabular}{lcccc}
\toprule
\textbf{Model} & \textbf{Objective Task Progress} & \textbf{Task Effectiveness} & \textbf{Overall Satisfaction} & \textbf{N} \\
\midrule
GPT-4o-audio-preview & $67.7\% \pm {\scriptstyle 5.6\%}$ & $4.77 \pm {\scriptstyle 0.20}$ & $4.80 \pm {\scriptstyle 0.16}$ & 44 \\
GPT-4o-mini+STT+TTS & $64.1\% \pm {\scriptstyle 5.6\%}$ & $4.64 \pm {\scriptstyle 0.21}$ & $4.27 \pm {\scriptstyle 0.21}$ & 45 \\
Gemini-2.5-Flash+TTS & $61.2\% \pm {\scriptstyle 6.3\%}$ & $4.34 \pm {\scriptstyle 0.22}$ & $4.43 \pm {\scriptstyle 0.22}$ & 35 \\
GPT-4o-mini-audio-preview & $55.7\% \pm {\scriptstyle 5.6\%}$ & $3.80 \pm {\scriptstyle 0.28}$ & $3.52 \pm {\scriptstyle 0.24}$ & 44 \\
Voxtral-Small-24B+TTS & $44.0\% \pm {\scriptstyle 6.8\%}$ & $2.91 \pm {\scriptstyle 0.24}$ & $3.03 \pm {\scriptstyle 0.25}$ & 35 \\
Qwen3-Omni-30B+TTS & $36.6\% \pm {\scriptstyle 5.9\%}$ & $3.93 \pm {\scriptstyle 0.26}$ & $3.73 \pm {\scriptstyle 0.25}$ & 45 \\
Ultravox-v0.4-ToolACE-8B+TTS & $30.1\% \pm {\scriptstyle 5.9\%}$ & $2.58 \pm {\scriptstyle 0.29}$ & $2.50 \pm {\scriptstyle 0.28}$ & 36 \\
\bottomrule
\end{tabular}
\caption{\textbf{Function calling performance.} Objective Task Progress shows percentage of verifiable task steps completed $\pm$ standard error; Task Effectiveness and Overall Satisfaction show mean user ratings $\pm$ standard error on function-calling scenarios only (6-point Likert scale, higher is better). N indicates the number of function-calling evaluations per model.}
\label{tab:function_calling}
\end{table*}

On function-calling tasks, GPT-4o-audio-preview achieved the highest objective task progress at 67.7\%, followed by GPT-4o-mini+STT+TTS at 64.1\% and Gemini-2.5-Flash+TTS at 61.2\%. Smaller open-source models showed substantially lower performance, with Ultravox-v0.4-ToolACE-8B+TTS completing only 30.1\% of verifiable task steps. The ranking on function-calling scenarios largely mirrors the overall results in Table~\ref{tab:human_ratings}, with GPT-4o-audio-preview maintaining its leading position and the performance gap between commercial and open-source models remaining substantial.

However, objective task progress correlates only moderately with overall satisfaction ($r=0.87$) and task effectiveness ($r=0.85$), with notable divergences revealing the complexity of user satisfaction. Most strikingly, GPT-4o-mini-audio-preview achieves 55.7\% task progress compared to Qwen3-Omni-30B+TTS's 36.6\%—a substantial difference—yet receives lower overall satisfaction (3.52 vs. 3.73). Similarly, GPT-4o-mini+STT+TTS completes 64.1\% of task steps but receives lower overall satisfaction (4.27) than Gemini-2.5-Flash+TTS at 61.2\% progress and 4.43 satisfaction. These discrepancies reveal that user satisfaction depends not only on task completion but also on interaction quality—conversational naturalness, responsiveness, and perceived effort—which can vary substantially across models even at similar completion rates. This underscores the limitation of benchmarks that measure task success in isolation without capturing holistic user experience.

\section{Fair Comparison: Pairwise Ranking Accuracy}
\label{appendix:fair_comparison}

While the LOMO Pearson correlation in Section~\ref{sec:preference_prediction} provides fine-grained comparison across subset selection methods, it includes in-sample predictions for training models, which may overestimate generalization performance. To provide a fair comparison between regression-based predictions and original subset scores, we evaluate pairwise ranking accuracy using a 5-2 train-test split.

\subsection{Evaluation Protocol}

We perform exhaustive 5-2 cross-validation across all possible splits of the 7 human-evaluated models:

\begin{enumerate}[leftmargin=*,noitemsep]
    \item For each of the ${7 \choose 2} = 21$ possible held-out pairs $(m_i, m_j)$:
    \begin{itemize}[noitemsep]
        \item Train Ridge regression on the remaining 5 models' subset scores and human ratings
        \item Select regularization strength $\alpha \in \{10^{-4}, 10^{-3}, \ldots, 10^4\}$ via nested leave-one-out CV on the 5 training models
        \item Retrain on all 5 models with the selected $\alpha$ and predict for the 2 held-out models: $\hat{y}_{m_i}, \hat{y}_{m_j}$
        \item Check if the pairwise ranking is correct: $(\hat{y}_{m_i} > \hat{y}_{m_j}) \iff (y_{m_i} > y_{m_j})$
    \end{itemize}
    \item Compute \textbf{pairwise ranking accuracy}: proportion of correctly ranked pairs across all 21 splits
\end{enumerate}

For comparison, we compute pairwise ranking accuracy using original subset scores on the same 21 held-out pairs without any regression training. This provides a fair evaluation where both approaches make predictions on truly unseen models.
\begin{table*}[h]
\centering
\small
\begin{tabular}{lcccc}
\toprule
\textbf{n} & \multicolumn{2}{c}{\textbf{"Best" Subset}} & \multicolumn{2}{c}{\textbf{Random Baseline}} \\
\cmidrule(lr){2-3} \cmidrule(lr){4-5}
 & Regression & Original & Regression & Original \\
\midrule
10 & $0.797 \pm {\scriptstyle 0.008}$ & $0.809 \pm {\scriptstyle 0.004}$ & $0.625 \pm {\scriptstyle 0.017}$ & $0.705 \pm {\scriptstyle 0.012}$ \\
20 & $0.817 \pm {\scriptstyle 0.007}$ & $0.807 \pm {\scriptstyle 0.004}$ & $0.666 \pm {\scriptstyle 0.015}$ & $0.751 \pm {\scriptstyle 0.011}$ \\
30 & $0.833 \pm {\scriptstyle 0.007}$ & $0.816 \pm {\scriptstyle 0.003}$ & $0.676 \pm {\scriptstyle 0.014}$ & $0.774 \pm {\scriptstyle 0.009}$ \\
50 & $0.882 \pm {\scriptstyle 0.004}$ & $0.825 \pm {\scriptstyle 0.004}$ & $0.715 \pm {\scriptstyle 0.013}$ & $0.802 \pm {\scriptstyle 0.008}$ \\
100 & $0.883 \pm {\scriptstyle 0.003}$ & $0.828 \pm {\scriptstyle 0.004}$ & $0.757 \pm {\scriptstyle 0.011}$ & $0.830 \pm {\scriptstyle 0.006}$ \\
200 & $0.878 \pm {\scriptstyle 0.003}$ & $0.850 \pm {\scriptstyle 0.004}$ & $0.800 \pm {\scriptstyle 0.008}$ & $0.846 \pm {\scriptstyle 0.005}$ \\
\midrule
Full & $0.857$ & $0.810$ & -- & -- \\
\bottomrule
\end{tabular}
\caption{\textbf{Fair pairwise ranking accuracy comparison.} Proportion of correctly ranked model pairs using 5-2 cross-validation (mean $\pm$ standard error over 100 random seeds). Both regression predictions and original subset scores are evaluated on the same 21 held-out pairs. "Best" Subset uses Anchor Points for $n \leq 30$ and Combined Embedding for $n \geq 50$.}
\label{tab:fair_pairwise_comparison}
\end{table*}

\subsection{Results}

Table~\ref{tab:fair_pairwise_comparison} presents pairwise ranking accuracy under fair 5-2 cross-validation. Key findings:

\begin{itemize}[leftmargin=*,noitemsep]
    \item \textbf{Regression improves generalization for larger subsets}: For $n \geq 30$, Ridge regression consistently outperforms original scores for best subsets. The advantage is most pronounced at $n=50$ and $n=100$ ($\Delta \approx 0.055$). This suggests that regression effectively predicts human preferences by learning how to weight different benchmark dimensions, capturing the compositional nature of user satisfaction better.
    
    \item \textbf{Small subsets show comparable performance}: At $n \leq 20$, regression and original scores achieve similar accuracy, likely because small subsets lack coverage of all important dimensions needed for regression to learn robust mappings. Additionally, the 5-2 split constraint limits training to only 5 source models. Our final released benchmarks use all 7 models for training, which should yield better human preference prediction.
    
    \item \textbf{Quality items are essential for effective regression}: Random sampling shows a striking pattern—regression \textit{underperforms} original scores across all sizes (e.g., $\Delta = -0.080$ at $n=10$, $\Delta = -0.046$ at $n=200$). This demonstrates that regression amplifies the signal from high-quality items but also amplifies noise from uninformative ones. Without principled selection, adding regression to random items degrades performance by overfitting to spurious patterns.
    
    \item \textbf{Quality over quantity}: Consistent with findings in Section~\ref{sec:preference_prediction}, regression performance peaks at $n=100$ ($0.883$) for best subsets before dropping to $0.878$ at $n=200$ and $0.857$ for the full benchmark. This confirms that adding more items introduces lower-informative examples that perturb regression weights.
\end{itemize}

These results validate our main findings: Ridge regression generalizes better to unseen models when applied to high-quality, diverse subsets, but requires principled item selection to avoid amplifying noise from uninformative examples.

\clearpage
\section{Licenses}
\label{app:licenses}

We list the licenses for artifacts involved in this work as follows:

\textbf{Models:}
\begin{itemize}
    \item \textbf{GPT-4o-audio-preview}, \textbf{GPT-4o-mini-audio-preview}, \textbf{GPT-4o-mini}, \textbf{GPT-5}, \textbf{GPT-realtime}, \textbf{GPT-4o-transcribe}, \textbf{GPT-4o-mini-tts}, \textbf{GPT-4.1}, \textbf{GPT-5.2}: Proprietary models. Usage governed by \textit{OpenAI Terms of Service}. \url{https://openai.com/policies/}
    \item \textbf{Gemini-2.5-Pro}, \textbf{Gemini-2.5-Flash}: Proprietary models. Usage governed by \textit{Google Generative AI Terms of Service}. \url{https://ai.google.dev/gemini-api/terms}
    \item \textbf{Qwen2.5-Omni-7B}: \textit{Apache License 2.0}.
    \item \textbf{Qwen3-Omni-30B-A3B-Instruct}: \textit{Apache License 2.0}
    \item \textbf{Ultravox-v0.4-ToolACE-8B}, \textbf{Ultravox-v0.5-llama-3.2-1B}, \textbf{Ultravox-v0.6-llama-3.1-8b}: \textit{MIT License}. 
    \item \textbf{Llama-3.2-3B}: \textit{Llama 3.2 Community License Agreement}. \url{https://www.llama.com/llama3_2/license/}
    \item \textbf{Voxtral-Small-24B-2507}, \textbf{Voxtral-Mini-3B-2507}: \textit{Apache License 2.0}
    \item \textbf{Granite-speech-3.3-8b}: \textit{Apache License 2.0}. 
    \item \textbf{Gemma-3n-e4b}, \textbf{Gemma-3n-e2b}: \textit{Gemma Terms of Use}. \url{https://ai.google.dev/gemma/terms}
\end{itemize}

\textbf{Benchmarks:}
\begin{itemize}
    \item \textbf{Dynamic-SUPERB Phase-2}: Individual datasets within the benchmark have varying licenses. Complete license information available at \url{https://github.com/dynamic-superb/dynamic-superb/blob/main/docs/dataset_license.md}
    \item \textbf{CAVA}: \textit{CC BY-SA 4.0} © 2024 Talk Arena. \url{https://github.com/SALT-NLP/CAVA}
    \item \textbf{UltraEval-Audio}: \textit{Apache License 2.0}.
    \item \textbf{SpeakBench} (AudioJudge): \textit{MIT License}.
    \item \textbf{WildSpeech-Bench}: \textit{Creative Commons Attribution 4.0 International (CC BY 4.0)}. 
\end{itemize}

\section{Intended Use and Compliance}
\label{sec:intended_use}

All artifacts used in this work are employed consistent with their intended purposes as specified by their creators. Existing benchmarks (Dynamic-SUPERB, CAVA, UltraEval-Audio, SpeakBench, WildSpeech-Bench), datasets (LMSYS-Chat-1M, WildChat, BFCL v3), and pre-trained models are used within their documented scope for research evaluation of audio model capabilities. Speech processing components (GPT-4o-transcribe, GPT-4o-mini-tts, Silero VAD, WavLM-Large, Whisper-large-v3) are employed for their intended audio processing purposes.

Our released HUMANS benchmark and human preference dataset are intended solely for research purposes in audio model evaluation and meta-analysis. The human preference dataset is derived from research participants who consented to academic research use, and any derivative use must comply with these original access conditions and privacy protections outlined in Section~\ref{sec:ethics}.

\section{Package Details}
\label{app:implementation}

We implement our experiments using Python 3.12 with the following key packages:
\begin{itemize}[noitemsep]
    \item PyTorch 2.8.0 with torchaudio 2.8.0 for model inference
    \item Transformers 4.51.3 (Hugging Face) for model loading and inference
    \item NumPy 2.2.6 for numerical computations
    \item Scikit-learn 1.7.0 for Ridge regression, PCA, and K-Means clustering
    \item SciPy 1.16.0 for statistical computations
    \item Datasets 3.6.0 (Hugging Face) for dataset loading and processing
    \item LiveKit 1.0.11 and livekit-agents 1.1.4 for real-time conversational agent infrastructure
    \item vLLM 0.10.2 for model deployment
\end{itemize}

All experiments were conducted on NVIDIA A6000 GPUs (48GB VRAM) for open-source model deployments.

\end{document}